\documentclass{article}

\usepackage{microtype}
\usepackage{graphicx}

\usepackage{hyperref}

\usepackage[accepted]{icml2021}

\usepackage{subcaption}
\usepackage{amsmath,amssymb,amsfonts}
\usepackage{changepage}
\usepackage{tabularx,booktabs}
\usepackage{multirow}
\usepackage{float}
\usepackage{wrapfig}
\usepackage{lipsum}

\usepackage[normalem]{ulem}

\newtheorem{proposition}{Proposition}
\newtheorem{remark}{Remark}

\def\b0{{\boldsymbol{0}}}
\newcommand{\mE}{{\mathbb E}}
\newcommand{\mR}{{\mathbb R}}
\newcommand{\mV}{{\mathcal{V}}}
\newcommand{\mI}{{\mathcal{I}}}

\newcommand{\jiaojiao}[1]{\textcolor{black}{#1}}

\icmltitlerunning{Scalable Computations of Wasserstein Barycenter via Input Convex Neural Networks}
\begin{document}

\twocolumn[
    \icmltitle{Scalable Computations of Wasserstein Barycenter via \\ Input Convex Neural Networks
    }



    \icmlsetsymbol{equal}{*}

    \begin{icmlauthorlist}
        \icmlauthor{Jiaojiao Fan}{gatech}
        \icmlauthor{Amirhossein Taghvaei}{uci}
        \icmlauthor{Yongxin Chen}{gatech}
    \end{icmlauthorlist}

    \icmlaffiliation{uci}{University of California, Irvine}
    \icmlaffiliation{gatech}{Georgia Institute of Technology}

    \icmlcorrespondingauthor{Yongxin Chen}{yongchen@gatech.edu}

    \icmlkeywords{Machine Learning, ICML}

    \vskip 0.3in
]



\printAffiliationsAndNotice{}  


\begin{abstract}
    Wasserstein Barycenter is a principled approach to represent the weighted mean of a given set of probability distributions, utilizing the geometry induced by optimal transport. In this work, we present a novel scalable algorithm to approximate the Wasserstein Barycenters
    aiming at high-dimensional applications in machine learning. Our proposed algorithm is based on the Kantorovich dual formulation of the Wasserstein-2 distance as well as a recent neural network architecture, input convex neural network, that is known to parametrize convex functions. The distinguishing features of our method are: i) it only requires samples from the marginal distributions; ii) unlike the existing approaches, it represents the Barycenter with a generative model and can thus generate infinite samples from the barycenter without querying the marginal distributions; iii) it works similar to Generative Adversarial Model in one marginal case.
    We demonstrate the efficacy of our algorithm by comparing it with the state-of-art methods in multiple experiments.
    \footnote{\jiaojiao{Download Code at \url{https://github.com/sbyebss/Scalable-Wasserstein-Barycenter}}}
\end{abstract}

\newcommand{\red}{\color{red}}
\section{Introduction}
The Wasserstein barycenter is concerned with the (weighted) average of multiple given probability distributions. It is based on the natural geometry over the space of distributions induced by optimal transport \cite{Vil03} theory and serves as a counterpart of arithmetic mean/average for data of distribution-type. Compared to other methods, Wasserstein barycenter provides a principled approach to average probability distributions, fully utilizing the underlying geometric structure of the data \cite{AguCar11}. During the past few years, it has found applications in several machine learning problems. For instance, in sensor fusion, Wasserstein barycenter is used to merge/average datasets collected from multiple sensors to generate a single collective result~\cite{elvander2018tracking}. The advantage of Wasserstein barycenter is its ability to preserves the modality of the different datasets, a highly desirable property in practice \cite{JiaNinGeo12}. Wasserstein Barycenter has also been observed to be effective in removing batch effects of the sensor measurements \cite{YanTab19}.  It has also found application in large scale Bayesian inference for averaging the results from Markov chain Monte Carlo (MCMC) Bayesian inference carried out over subsets of the observations \cite{srivastava2015wasp,staib2017parallel,SriLiDun18}. It has also been useful in image processing for texture mixing~\cite{rabin2011wasserstein} and shape interpolation~\cite{SolDeGui15}.


The bottleneck of utilizing Wasserstein barycenter in machine learning applications remains to be computational.
Indeed, when the data is discrete, namely, the given probability distributions are over discrete space (e.g., grid), the Wasserstein barycenter problem can be solved using linear programming~\cite{anderes2016discrete}.
This has been greatly accelerated by introducing an entropy term \citep[Algorithm 1]{CutDou14} \cite{SolDeGui15} as in Sinkhorn algorithm \cite{Cut13}. However, these methods are not suitable for many machine learning applications involving distributions over continuous space.
First of all, it requires discretization of the distribution to implement these methods and thus doesn't scale to high dimensional settings.
Secondly, in some applications such as MCMC Bayesian inference \cite{AndDeDouJor03,SriLiDun18} the explicit formulas of the distributions are not accessible, which precludes these discretization-based algorithms.
\jiaojiao{When the support atoms are free to move, there are algorithms that interchangeably optimize the support weights and locations \citep[Algorithm 2]{CutDou14} \citep{sinkhorn_barycenter,barycenter-polynomial} which can also be formulated in stochastic optimization framework \cite{claici2018stochastic}.} But these free-support methods become computationally highly expensive when the number of support points  is large.

    {\bf Contribution:}
We propose a computationally efficient and scalable algorithm for estimating the Wasserstein barycenter of probability distributions over continuous spaces. Our method is based on a Kantorovich-type dual characterization of the Wasserstein barycenter, which involves optimization over convex functions, and  the recently introduced input convex neural networks (ICNN) \cite{AmoXuKol17,CheShiZha18}, that provides powerful representation of convex functions. Remarkably, in our framework, the (weighted) barycenter is represented by a generator network  \cite{GooPouBen14,pmlr-v70-arjovsky17a}, that allows characterization of continuous distributions and fast and unlimited sampling from barycenter.  We prove the consistency of our formulation in Proposition~\ref{prop:solution}, and  demonstrate its performance and its scaling properties in truly high-dimensional setting through extensive evaluations over various benchmark experiments including synthetic and real data-set and provide comparisons with several state-of-art algorithms: \citet{korotin2021continuous}, \citet{li2020continuous}, \citet{CutDou14}. Our experiments reveal significant improvement in estimating the barycenter in high-dimensional setting compared to most existing algorithms. We also showcase the the ability of our method to perform as a generative adversarial network (GAN) in the one marginal case and propose a heuristic extension to learn barycenter of arbitrary weights through a single training process.


    {\bf Related work:}
Our proposed algorithm is closely related to continuous Wasserstein barycenter proposed by \citet{li2020continuous} and \citet{korotin2021continuous}. Similar to our approach, both of them are based on dual formulation of the Wasserstein barycenter problem and representing the potential functions with neural networks. However, the approach in \citet{li2020continuous} does not restrict the potential functions to be convex. Instead, an entropic or $l_2$ regularization term is added to ensure that the optimal potential functions are approximately convex. The addition of the regularization term introduces undesirable bias error which becomes severe in high-dimensional problems as shown in Figure~\ref{fig:Gaussian 3 marginal highD results}-\ref{fig:0-1} and also reported in \citep[Table1-4]{korotin2021continuous} \citep[Section 5]{li2020continuous}. The approach in \citet{korotin2021continuous} restricts the potentials to be convex using ICNN, however, it includes  a cycle regularizer term to ensure the potential functions are dual conjugate. Their formulation also involves a congruence regularizer to guarantee that the optimal potential functions are consistent with the true barycenter. The congruence regularizer requires selection of a priory probability distribution that is bounded below by the true barycenter, which is a non-trivial task.
Moreover, the addition of the regularization terms distorts the nice optimization landscape of the original problem. The problem may become non-convex even for the simple setting of restriction to quadratic functions (see Sec.~\ref{sec:landscape} in supplementary material). In contrast, our formulation does not involve additional regularization terms and retains the optimization landscape of the original problem.
Moreover, a distinct feature of our algorithm is representing the barycenter using a generative model which allows a low-dimensional representation of the barycenter and access to infinitely many samples, while both of these methods represent the barycenter using the Monge maps from the marginals, and limits the number of samples to the number available from marginal distributions.

Earlier stochastic Wasserstein barycenter method \cite{claici2018stochastic} also aims at calculating barycenters for continuous distributions using samples. However, they adopt a semi-discrete approach that models the barycenter with a finite set of points. That is, even though the marginals are continuous, the barycenter is discrete.  Several other sample-based algorithms \cite{staib2017parallel,KuaTab19,MiYuBenWan20} are also of semi-discrete-type.
Most other Wasserstein barycenter algorithms are for discrete distributions and require discretization if applied to continuous distributions. An incomplete list includes \cite{CutDou14,BenCarCutPey15,SolDeGui15}.

The subject of this work is also related to the vast amount of literature on estimating the optimal transport map and Wasserstein distance (see~\cite{peyre2019computational} for a complete list). Closely related to this paper are the recent works that aim to extend the optimal transport map estimation to large-scale machine learning settings~\cite{genevay2016stochastic,seguy2017large,liu2018two,CheGeoTan18,leygonie2019adversarial,xie2019scalable}. In particular, our algorithm is inspired by the recent advances in estimation of optimal transport map and Wasserstein-2 distance using ICNNs~\cite{taghvaei20192,MakTagOhLee19,korotin2021wasserstein}.

\section{Background}
\subsection{Optimal transport and Wasserstein distance}
Given two probability distributions $\nu, \mu$ over Euclidean space $\mR^n$ with finite second moments, the optimal transport \cite{Vil03} (OT) problem with quadratic unit cost seeks an optimal joint distribution of $\nu,\mu$ that minimizes the total transport cost. More specifically, it is formulated as $W_{2}^{2}\left(\nu, \mu\right):=\min_{\pi\in \Pi(\nu,\mu)} \int_{\mR^n\times\mR^n} \|x-y\|^{2} d \pi(x, y)$,
where $\Pi(\nu,\mu)$ denotes the set of all joint distributions of $\nu$ and $\mu$. The square-root of the minimum transport cost defines the celebrated Wasserstein-2 distance $W_2$, which is known to enjoy many nice geometrical properties compared to other distances for distributions \cite{ambrosio2008gradient}. 

The Kantorovich dual of the OT problem reads
\begin{equation}\label{eq:sup}
  \frac{1}{2} W_{2}^{2}(\nu, \mu)=\sup _{(\phi, \psi) \in \Phi} \mathbb{E}_{\nu}[\phi(X)]+\mathbb{E}_{\mu}[\psi(Y)],
\end{equation}
where  $\Phi:= \{(\phi, \psi) \in L^{1}(\nu) \times L^{1}(\mu)\,;\, \phi(x)+\psi(y) \leq \frac{1}{2}\|x-y\|^{2},~~\forall x, y\}$.
Let $f(x)=\|x\|^{2}/2-\phi(x)$, then \eqref{eq:sup} can be rewritten as
\begin{equation}\label{eq:extract two momentum}
  \frac{1}{2} W_{2}^{2}(\nu, \mu)\!=\!C_{\nu, \mu}\!\!-\!\!\underset{f \in \textbf{CVX}}{\inf}\{ \mathbb{E}_{\nu}[f(X)]\!+\!\mathbb{E}_{\mu}[f^*(Y)]\}
\end{equation}
where \textbf{CVX} stands for the set of convex functions, $C_{\nu, \mu}:=(1 / 2) \{\mathbb{E}_{\nu}[\|X\|^{2}]+
  \mathbb{E}_{\mu}[\|Y\|^{2}]\}$, and the $f^{*}$ is the convex conjugate \cite{Roc70} function of $f$. The formulation \eqref{eq:extract two momentum} is known as the semi-dual formulation of OT. \jiaojiao{The \textbf{CVX} condition restricts the search space for $f$ which becomes handy for design of optimization algorithms.}
\begin{remark}
  \label{remark:map}
  When both of the marginal distributions have densities, \textit{Brenier' Theorem} gives that $\nabla f^*$ is the optimal transport map from $\mu$ to $\nu$ \cite{Vil03} and $\nabla f$ is the optimal map from $\nu$ to $\mu$.
\end{remark}

\subsection{Wasserstein Barycenter}\label{sec:barycenter}
Wasserstein barycenter is OT-based average of probability distributions. Given a set of probability distributions $\mu_i,~ i=1,2,\ldots,N$ and a weight vector $a\in \mR^N$ ($a_i\ge 0,~i=1,2,\ldots,N$ and $\sum_{i=1}^N a_i = 1$), the associated Wasserstein barycenter is defined as the minimizer of
\begin{equation}\label{eq:initial target}
  \min_{\nu} ~ \sum_{i=1}^{N} a_i W_{2}^{2}\left(\nu, \mu_{i}\right).
\end{equation}
The barycenter problem \eqref{eq:initial target} can be reformulated as a linear programming \cite{AguCar11}. However, the linear programming-base algorithms don't scale well for high dimensional problems.
A special case that can be solved efficiently is when the marginal distributions $\{\mu_i\}$ are Gaussian. Denote the mean and covariance of $\mu_i$ as $m_{i}$ and $\Sigma_{i}$ respectively, then their Wasserstein barycenter is a Gaussian distribution with mean being $m=\sum_{i=1}^{N} a_{i} m_{i}$ and covariance $\Sigma$ being the unique solution to the fixed-point equation $\Sigma=\sum_{i=1}^{N} a_{i}(\Sigma^{1 / 2} \Sigma_{i} \Sigma^{1 / 2})^{1 / 2}$.
In \citet{alvarez2016fixed}, a simple however efficient algorithm was proposed to solve for $\Sigma$.

\subsection{Input Convex Neural Network}
Input Convex Neural Network (ICNN) is a type of deep neural networks architecture that characterize convex functions \cite{AmoXuKol17}. A fully ICNN (FICNN) leads to a function that is convex with respect to all inputs.

\begin{figure}[ht]
  \begin{center}
    \centerline{\includegraphics[width=0.9\columnwidth]{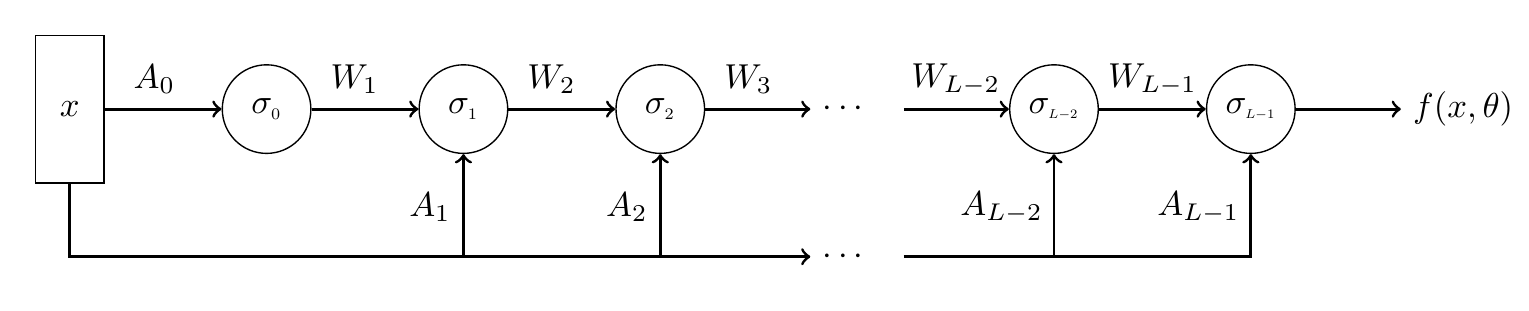}  }
    \caption{
    {Fully input convex neural network (FICNN) }}
    \label{fig:FICNN}
  \end{center}
  \vskip -0.2in
\end{figure}

The FICNN architecture is shown in Fig. \ref{fig:FICNN}. It is a $L$-layer feedforward neural network propagating following, for $l=0,1, \ldots, L-1$
\begin{equation}\label{eq:FICNN}
  z_{l+1}=\sigma_{l}\left(W_{l} z_{l}+A_{l} x+b_{l}\right),
\end{equation}
where $\left\{W_{l}\right\},\left\{A_{l}\right\}$ are weight matrices (with the convention that $W_{0}=0$), $\left\{b_{l}\right\}$ are the bias terms, and $\sigma_{l}$ denotes the entry-wise activation function at the layer $l$. Denote the total set of parameters by $\theta=\left(\left\{W_{l}\right\},\left\{A_{l}\right\},\left\{b_{l}\right\}\right)$, then this network defines a map from input $x$ to $f(x ; \theta)=z_{L}$. This map $f(x ; \theta)$ is convex in $x$ provided 1) $W_{1:L-1}$ are non-negative; 2) $\sigma_{0:L-1}$ are convex; 3) $\sigma_{1:L-1}$ are non-decreasing \citep{MakTagOhLee19}. We remark that FICNN has the ability to approximate any convex function over a compact domain with a desired accuracy \citep{CheShiZha18}, which makes FICNN an ideal candidate for modeling convex functions.


\section{Methods and algorithms}
We study the Wasserstein barycenter problem \eqref{eq:initial target} for a given set of marginal distributions $\{\mu_i;~i=1,\ldots, N\}$. We consider the setting where the analytic forms of the marginals are not available. Instead, we only have access to independent samples from them. It can be either the cases where a fix set of samples is provided as in supervised learning, or the cases where one can keep sampling from the marginals like in the MCMC Bayesian \citep{SriLiDun18}. Our goal is to recover the true continuous Barycenter $\nu$.
\subsection{Deriving the dual problem over convex functions} \label{sec:derive the dual}
For a fixed $\nu$, the objective function of \eqref{eq:initial target} is simply a (scaled) summation of the Wasserstein cost between $\nu$ and $\mu_i$. Thus, we utilize the semi-dual formulation \eqref{eq:extract two momentum} of OT to evaluate the objective function of \eqref{eq:initial target}. 
\jiaojiao{However, convex conjugate function $f^*$ is not available explicitly in most of applications, thus we characterize it as }
\begin{equation} \label{eq:conjugate f}
	f^{*}(y) = \sup_{g \in  \textbf{CVX}} \langle y, \nabla g(y)\rangle-f(\nabla g(y))
\end{equation}
with the maximum being achieved at $g=f^*$, the semi-dual formulation \eqref{eq:extract two momentum} can be rewritten as
\begin{equation}\label{eq:Amir's function}
	\frac{1}{2} W_{2}^{2}(\nu, \mu) =  \sup_{f \in \textbf{CVX}}\,\underset{{g \in \textbf{CVX}} }{\inf} \mathcal{V}_{\nu,\mu}(f, g) +
	C_{\nu, \mu},
\end{equation}
where $\mathcal{V}_{\nu,\mu}(f, g)$ is a functional of $f$ and $g$ defined as
\begin{equation} \label{eq:capital V definition}
	\mathcal{V}_{\nu, \mu}(f,\! g)\!=\!-\mathbb{E}_{\nu}[f(X\!)]\!-\!\mathbb{E}_{\mu}[\langle Y,\! \nabla g(Y)\rangle\!\!-\!\!f(\nabla g(Y\!))].
\end{equation}
This formulation \eqref{eq:Amir's function} has been utilized in conjugation with FICNN to solve OT problem in \citep{MakTagOhLee19} and proved to be advantageous.


Plugging \eqref{eq:Amir's function} into the Wasserstein barycenter problem \eqref{eq:initial target}, we obtain the following reformulation
\begin{equation}\label{eq:precise final target}
	\underset{\nu}{\min}
	\sum_{i=1}^{N} a_i \left\{
	\underset{f_i \in \textbf{CVX}} {\sup}\,\underset{{g_i \in \textbf{CVX}} }{\inf}
	\mathcal{V}_{\nu, \mu_i}(f_i, g_i)
	+ C_{\nu, \mu_i}\right\}.
\end{equation}
Note that we have used different functions $(f_i, g_i)$ to estimate $W_2^2 (\nu, \mu_i)$. The first minimization is over all the possible probability distributions to search for the Wasserstein barycenter. This min-max-min formulation enjoys the following property, whose proof is in the supplementary material.
\begin{proposition}\label{prop:solution}
	When all the marginal distributions $\mu_i$ are absolutely continuous with respect to the Lebesgue measure, the unique Wasserstein barycenter $\nu^\star$ of them solves \eqref{eq:precise final target}. Moreover, the corresponding optimal $f_i^\star$ is the optimal potential in \eqref{eq:extract two momentum} associated with marginals $\nu^\star$ and $\mu_i$.
\end{proposition}

\begin{remark}
	Obtaining convergence rate for first-order optimization algorithms solving~\eqref{eq:precise final target} is challenging even in the ideal setting that the optimization is carried out in the  space of probability distributions. The difficulty arises because of the optimization over $\nu$.
	While the inner optimization over  $f_i$ and $g_i$ are concave and convex respectively, the optimization over $\nu$ is not convex. Precisely, it is not geodesically convex on the space of probability distributions equipped with Wasserstein-2 metric~\citep{ambrosio2008gradient}. However, it is possible to obtain guarantees in a restricted setting by establishing a Polyak- Lojasiewicz type inequality. In particular, assuming all $\mu_i$ are Gaussian with positive-definite covariance matrices, the gradient-descent algorithm admits a linear convergence rate~\citep{chewi2020gradient}.
\end{remark}

\subsection{Solving the barycenter problem}
Consider the Wasserstein barycenter problem for a fixed weight vector $a$. Following \citep{MakTagOhLee19} we use FICNN architecture to represent convex  functions $f_i$ and $g_i$. We now use a generator $h$ to model the distribution $\nu$, by transforming samples from a simple distribution $\eta$ (e.g., Gaussian, uniform) to a complicated distribution, thereby we recover a continuous Barycenter distribution. Thus, using this network parametrization and discarding constant terms, we arrive at the following optimization problem
\begin{equation}\label{eq:FICNN final object}
	\underset{h}{\min} \underset{f_{i} \in \textbf{FICNN}}{\sup}\,
	\underset{{g_{i} \in \textbf{FICNN}} }{\inf}
	\frac{1}{2}\mE_\eta [\|h(Z)\|^2]+
	\sum_{i=1}^{N} a_i \overline{\mV}_{\eta, \mu_i}(f_i, g_i),
\end{equation}
where $\overline{\mV}_{\eta, \mu_i}(f, g)$ is defined as
\begin{align*} \label{eq:overline V definition}
	-\mathbb{E}_{\eta}[f_i(h(Z))]-\mathbb{E}_{\mu_i}[\langle Y^i, \nabla g_i(Y^i)\rangle-f_i(\nabla g_i(Y^i))].
\end{align*}
\begin{figure}[h!]
	\begin{center}
		\centerline{\includegraphics[width=\columnwidth]{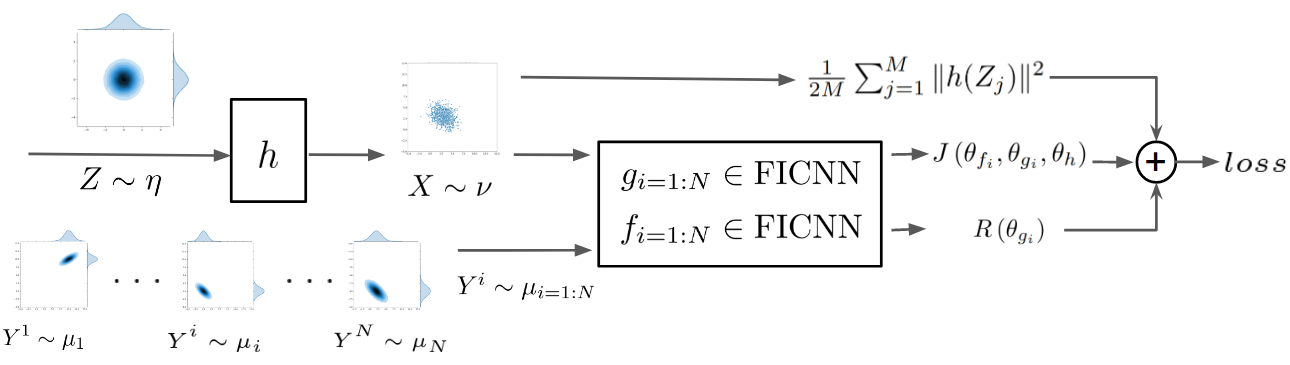}  }
		\caption{Block diagram for our neural Wasserstein barycenter (NWB) algorithm}
		\label{diagram1}
	\end{center}
	\vskip -0.2in
\end{figure}
We propose Neural Wasserstein Barycenter (NWB) algorithm (Algorithm \ref{al:three loops average weight}) to solve this three-loop min-max-min problem by alternatively updating $h$, $f_i$ and $g_i$ using stochastic optimization algorithms. This pipeline is illustrated by the block diagram (Figure \ref{diagram1}). We remark that the objective function in \eqref{eq:FICNN final object} can be estimated using samples from $\mu_i, \eta$. Thus, we just need access to the samples generated by the marginal distributions $\mu_i$ instead of their analytic form to compute their Wasserstein barycenter.
In practice, we found it more effective to replace the convexity constraints for $g_i$ with a convexity penalty, that is, the negative values of the weight matrices $W_l$ in FICNN \eqref{eq:FICNN}.

Denoting the parameters of $h, f_i, g_i$ by $\theta_h, \theta_{f_i}, \theta_{g_i}$ respectively and the batch size by $M$, we arrive at the batch estimation of the objective
\begin{equation}\label{eq:three loops empirical FICNN counterpart}
	\sum_{i=1}^{N}a_i\left[
		J(\theta_{f_i}, \theta_{g_i},\theta_{h})
		+R\left(\theta_{g_i}\right)\right]
	+ \frac{1}{2M}\sum_{j=1}^{M} ||h(Z_j)||^2,
\end{equation}
where $J\left(\theta_{f_i}, \theta_{g_i},\theta_{h}\right)$ represents
\begin{align*}
	\frac{1}{M} \sum_{j=1}^{M}
	f_i\left(\nabla g_i\left(Y_{j}^i\right)\right)-\left\langle Y_{j}^i, \nabla g_i\left(Y_{j}^i\right)\right\rangle
	-f_i\left(h(Z_j)\right),
\end{align*}

$R\left(\theta_{g_i}\right)=\lambda \sum_{W_{l} \in \theta_{g_i}}\left\|\max \left(-W_{l}, 0\right)\right\|_{F}^{2}$,
$Y_{j}^i$ represents the $j^{th}$ sample generated by $\mu_i$, $\{Z_j\}$ are samples from $\eta$, and $\lambda>0$ is a hyper-parameter weighing the intensity of regularization.

Algorithm \ref{al:three loops average weight} can be extended to obtain the barycenters for all weights in one shot. This extension is included in Sec. B of the supplementary material.
\begin{remark}
	It is tempting to combine the two minimization steps over $h$ and $g_i$ into one and reduce \eqref{eq:FICNN final object} into a min-max saddle point problem. The resulting algorithm only alternates between $f_i$ updates and $h, g_i$ updates instead of the three-way alternating in Algorithm \ref{al:three loops average weight}. However, in our implementations, we observed that this strategy is unstable.
\end{remark}
\begin{algorithm}[tb]
	\caption{Neural Wasserstein Barycenter (NWB)}
	\label{al:three loops average weight}
	\begin{algorithmic}
		\STATE {\bfseries Input:} Marginal dist. $\mu_{1:N}$, Generator dist. $\eta$, Batch size $M$
		\FOR{$k_3=1,\ldots,K_3$}
		\STATE {Sample batch {$\left\{Z_{j}\right\}_{j=1}^{M} \sim \eta$}}
		\STATE {Sample batch {$\left\{Y_{j}^{i}\right\}_{j=1}^{M} \sim \mu_i$}} for all $i=1, \cdots,N$
		\FOR{$k_2=1,\ldots,K_2$}
		\FOR{$k_1=1,\ldots,K_1$}
		\STATE {Update all $\theta_{g_i}$ to decrease \eqref{eq:three loops empirical FICNN counterpart}}
		\ENDFOR
		\STATE {Update all $\theta_{f_i}$ to increase  \eqref{eq:three loops empirical FICNN counterpart}}
		\STATE{Clip: $W_l =\max(W_l, 0)$ for all $\theta_{f_i}$}
		\ENDFOR
		\STATE{Update $\theta_h$ to decrease \eqref{eq:three loops empirical FICNN counterpart}}
		\ENDFOR
	\end{algorithmic}
\end{algorithm}

\paragraph{Computation complexity}
For our algorithm, as well as the algorithms recently proposed in \cite{korotin2021continuous,li2020continuous}, the computational complexity per iteration scales with $O(NMp)$ where $N$ is the number of marginals, $M$ is the batch-size, and $p$ is the size of the  network (the size of network scales almost linearly with dimension $d$). This should be compared with $O(NMK)$ for \citet{claici2018stochastic} where $K$ is the size of the support for barycenter, and $O(NnK)$ for \citet{CutDou14} where $n$ is the number of samples of the marginals. Although the size of the network is large, our approach is favored for large scale problems where the number of samples $n$ and the dimension $d$ are large (the number of atoms required to approximate a density scales exponentially with dimension).
\subsection{Recovering the barycenter}
Once Algorithm \ref{al:three loops average weight} converges, there are two distinct approaches to recover the Wasserstein barycenter: one through $h$ and one through $g_i$.

	{\bf Generative model $h_\sharp \eta$:} In our problem formulation \eqref{eq:FICNN final object}, the barycenter center is modeled by $h(Z)$ where $Z$ is sampled from a simple distribution $\eta$. That is, the barycenter $\nu$ is the pushforward of $\eta$ through the map $h$, denoted by $h_\sharp \eta$. Once the optimal $h$ is obtained, we can easily sample from the barycenter by sampling $Z_j$ from $\eta$ and apply the map $h(Z_j)$.

	{\bf Pushforward map $\nabla g_i \sharp \mu_i$:}\label{remark:pushforward} An alternative method to recover the barycenter is based on the fact, once Algorithm \ref{al:three loops average weight} converges, the pair $(f_i, g_i)$ solves the OT problem \eqref{eq:Amir's function} between the barycenter $\nu$ and the marginal $\mu_i$. As mentioned in Remark \ref{remark:map}, $\nabla f^*$ is the optimal map from marginal distribution to the barycenter. Moreover, the optimal $g_i$ is achieved at $g=f^*$. Hence, the pushforward of $\mu_i$ through map $\nabla g_i$ is the barycenter. Thus, to sample from the barycenter, we can sample $Y_j^i$ from a marginal and then apply map $\nabla g_i(Y_j^i)$. Note that this approach cannot generate more samples than those are already available in the marginals.

\subsection{Barycenter serving as GAN}\label{sec: algorithm wgan}
In case where there is only one marginal distribution, that is, $N=1$ in \eqref{eq:FICNN final object}, Algorithm \ref{al:three loops average weight} can be viewed as a type of generative adversarial network (GAN). More specifically, when $N=1$, the barycener $\nu$ coincides with the marginal distribution $\mu_1$. Given samples $\{Y_1^j\}$ from the marginal $\mu_1$, Algorithm \ref{al:three loops average weight} produces a generative model $h(Z)$ whose distribution matches the marginal $\mu_1$. Note that one can easily sample using $h(Z)$ and get samples that do not exist in the training data $\{Y_1^j\}$.

In fact, when $N=1$, Algorithm~\ref{al:three loops average weight} works very much like a Wasserstein Generative Adversarial Network (WGAN) which leverages the Wasserstein distance to distinguish fake and real samples in GAN. The original WGAN \cite{pmlr-v70-arjovsky17a} is based on the dual formulation for the Wasserstein-1 distance $W_1$. A heuristic weight clipping \cite{pmlr-v70-arjovsky17a} technique is used to enforce the Lipschitz condition on the potential function in the dual formulation. The WGAN was later improved in WGAN-GP \citep{gulrajani2017improved} via adding a gradient penalty term to promote the Lipschitz condition.
From this point of view, Algorithm \ref{al:three loops average weight} provides an alternative way to train the generative model with Wasserstein-2 metric (c.f. \citet{leygonie2019adversarial,korotin2021wasserstein,salimans2018improving,pmlr-v84-genevay18a}).


\section{Experiments}
\begin{figure*}[ht!]
	\begin{subfigure}{.2\textwidth}
		\centering
		\includegraphics[width=0.8\linewidth]{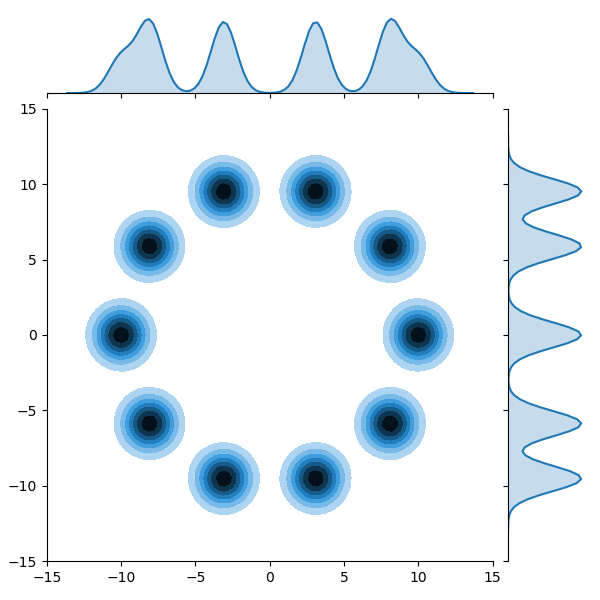}
		\caption{marginal $\mu_1$}
		\label{fig:sub-first}
	\end{subfigure}\hfill
	\begin{subfigure}{0.2\textwidth}
		\centering
		\includegraphics[width=0.8\linewidth]{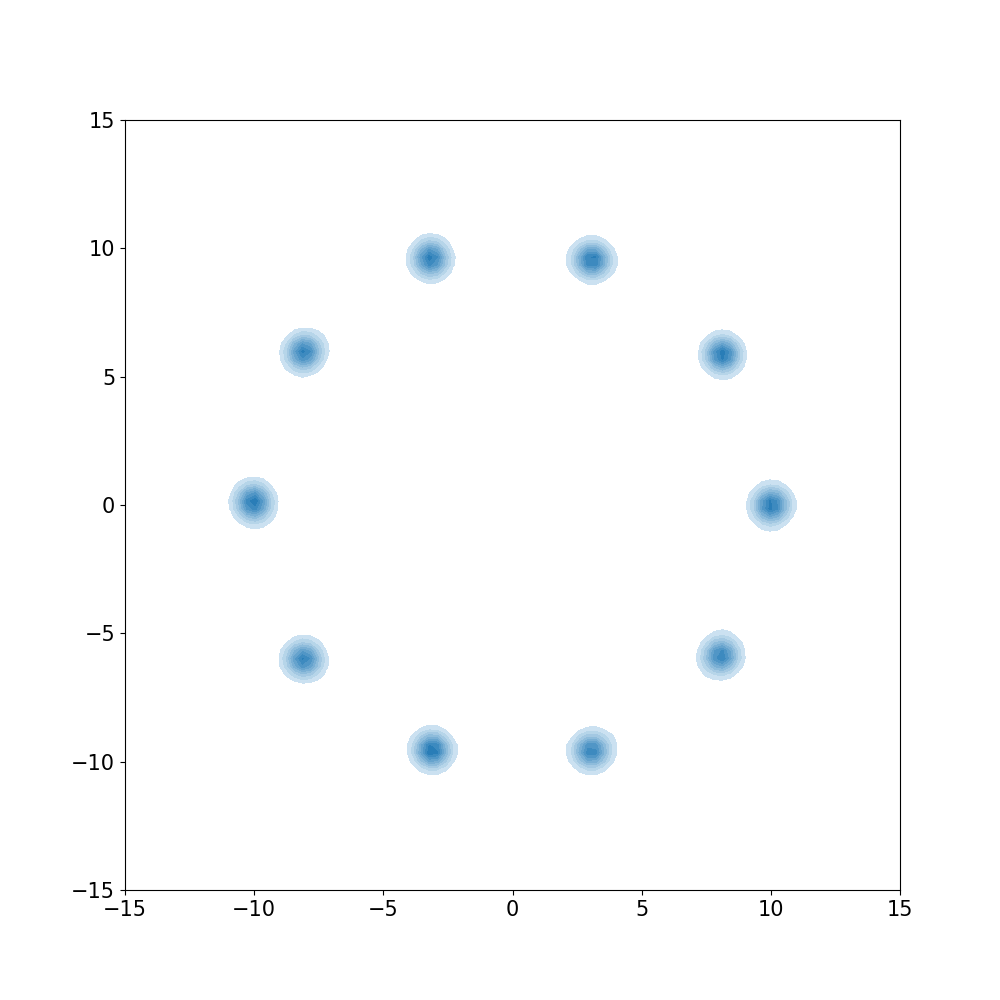}
		\caption{NWB (ours)}
		\label{fig:sub-second}
	\end{subfigure}\hfill
	\begin{subfigure}{0.2\textwidth}
		\centering
		\includegraphics[width=0.8\linewidth]{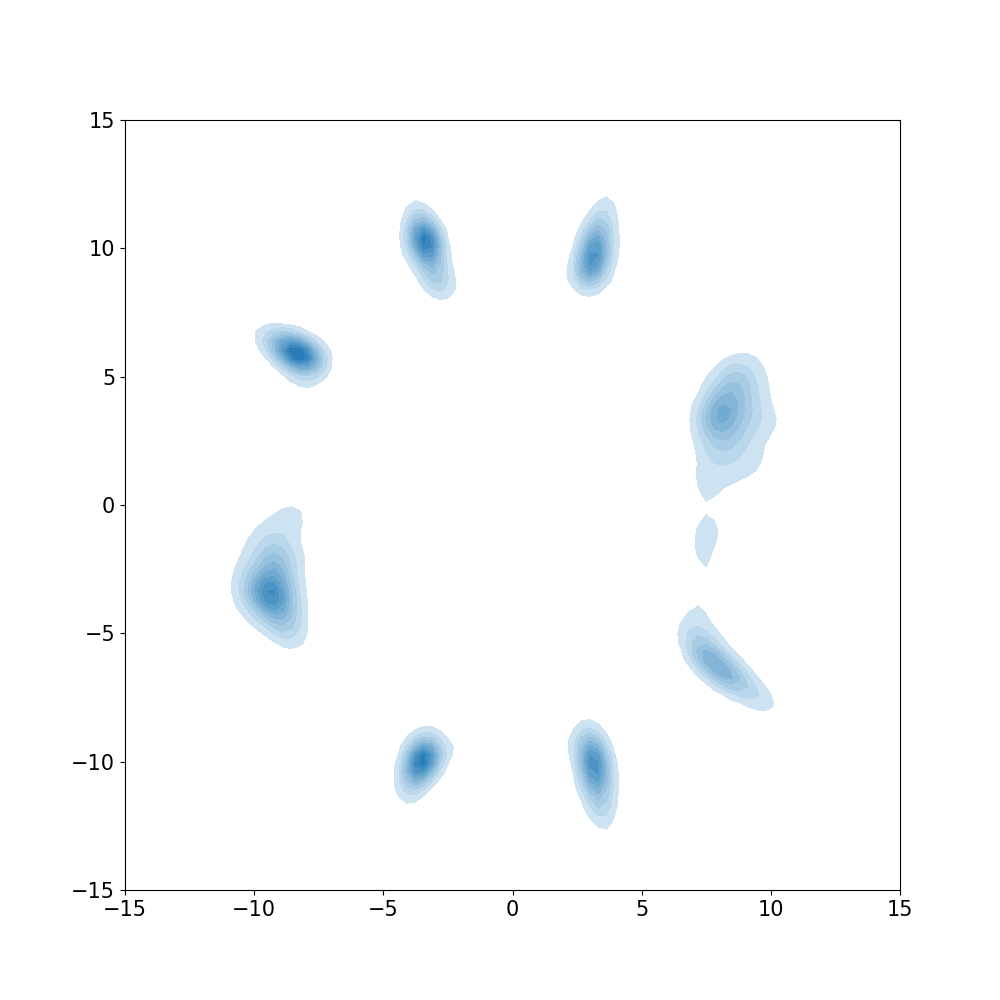}
		\caption{WGAN}
		\label{fig:sub-second}
	\end{subfigure}\hfill
	\begin{subfigure}{0.2\textwidth}
		\centering
		\includegraphics[width=0.8\linewidth]{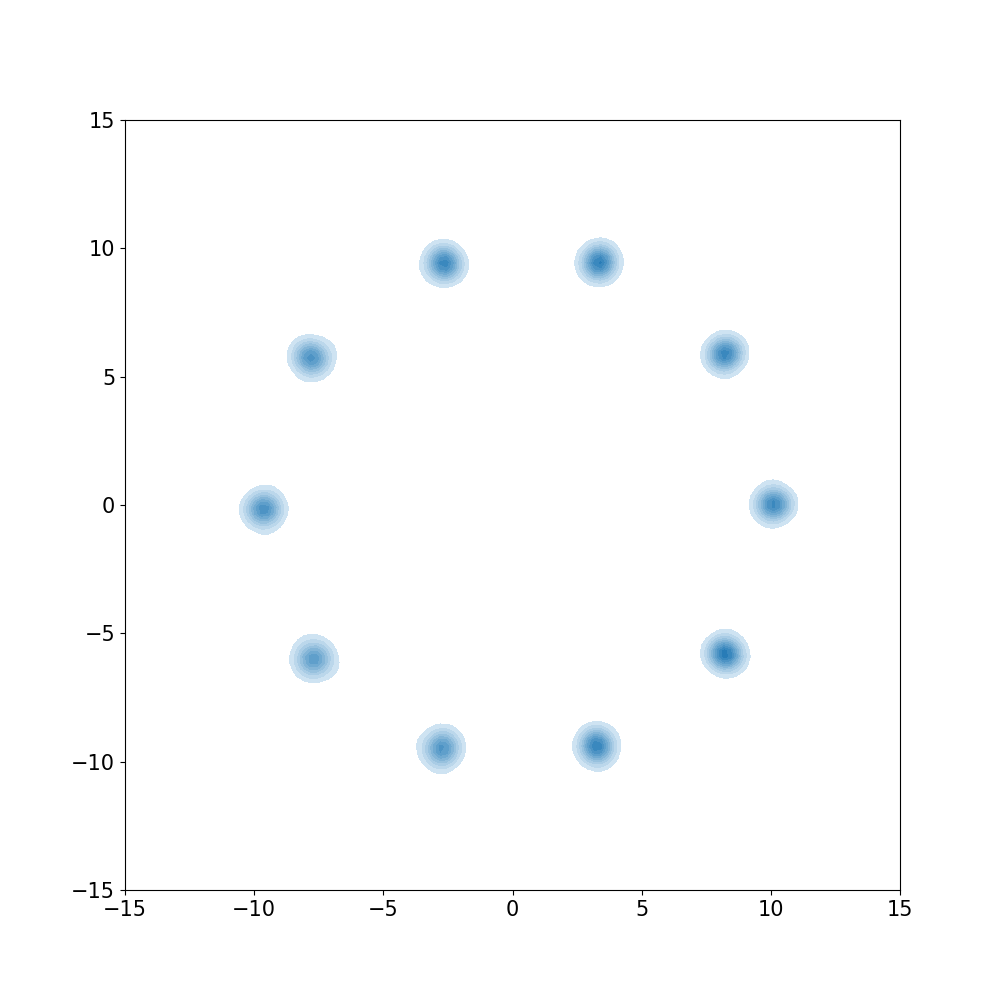}
		\caption{WGAN-GP}
		\label{fig:sub-second}
	\end{subfigure}\hfill
	\begin{subfigure}{0.2\textwidth}
		\centering
		\includegraphics[width=0.8\linewidth]{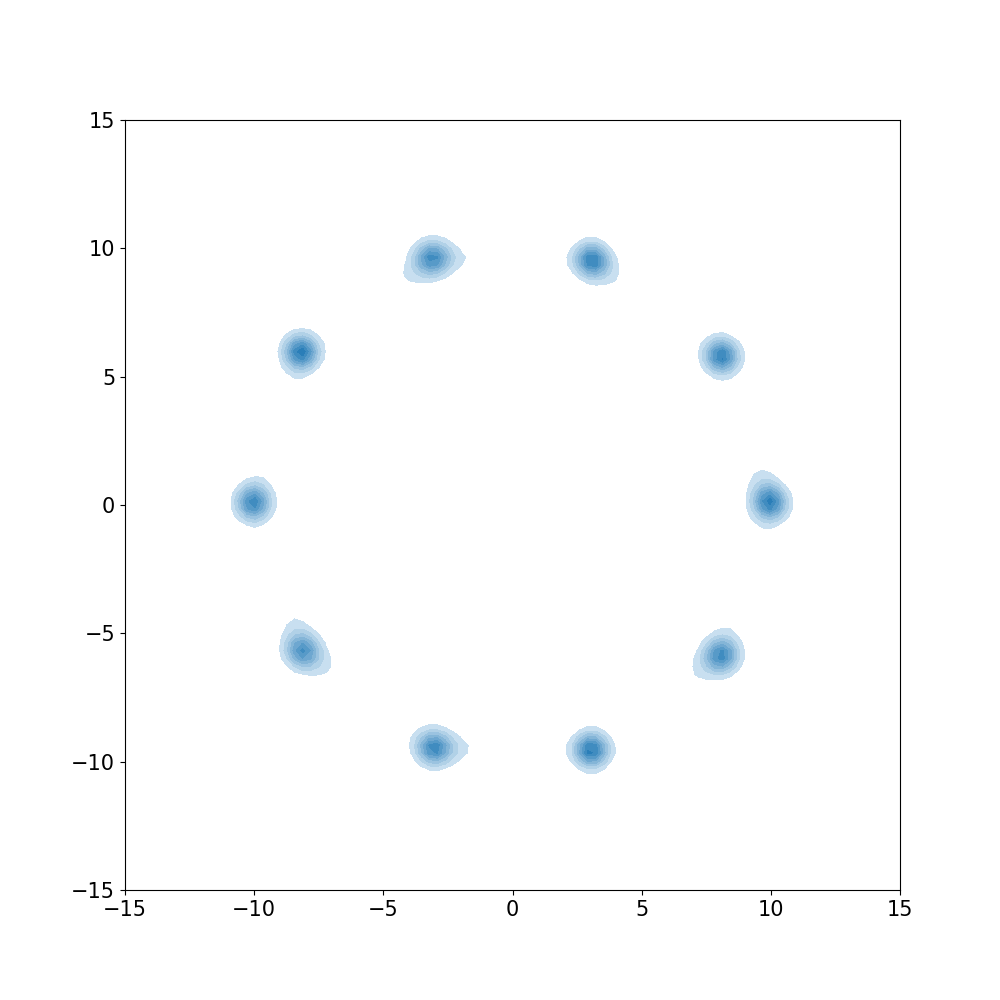}
		\caption{{W2GN}}
		\label{fig:sub-second}
	\end{subfigure}
	\caption{performance of our algorithm as GAN in single marginal case ($N=1$): learning Gaussian mixture}
	\label{fig:Gaussian mixture 1 marginal 2D results}
	\vskip -0.2in
\end{figure*}
In Section~\ref{sec:exp-2d}, we present numerical experiments on 2D/3D datasets which serve as proof of concept and qualitatively illustrate the performance of our approach.
In Section~\ref{sec:exp-high-dim}, we numerically study the effects of dimensionality and demonstrate the scalability of our algorithms to high-dimensional problems.
In Section~\ref{sec:bayesian} and \ref{sec:color}, we apply our algorithm in tasks such as Bayesian inference with large scale dataset and color transfer.
In Section~\ref{sec:gan}, we illustrate the ability of our algorithm to serve as a generative model.
The implementation details and further experiments are included in the supplementary materials.

For comparison, we choose the following state of the art algorithms:
(i) fast free-support Wasserstein barycenter (CDWB) \citep[Section 4.4]{CutDou14};
(ii) continuous Wasserstein barycenter without minimax optimization (CWB) \cite{korotin2021continuous};
(iii) continuous regularized Wasserstein barycenter (CRWB) \cite{li2020continuous}.
CWB and CRWB involve optimization over $N$ pairs of potentials $\{f_i,g_i\}$ as in NWB, and recover barycenter through ${\nabla g_i \sharp \mu_i}$.
The implementations of these algorithms are based on published code associated with the papers.

To evaluate the performance of these algorithms, we use the Bures-Wasserstein UVP \citep[Section 5]{korotin2021continuous}
\begin{equation}\label{eq:BW}
	\text{BW}_{2}^{2} \textendash \text{UVP}\left(\nu, \tilde{\nu}\right)
	\stackrel{\text { def }}{=}
	100 \frac{\text{BW}_{2}^{2}\left(\nu, \tilde{\nu}\right) }{\frac{1}{2} \text{Var}(\tilde{\nu} )} \%,
\end{equation}
where $\text{BW}_{2}^{2}\left(\nu, \tilde{\nu}\right)$ equals
\begin{equation}\nonumber
	\frac{1}{2}\| m_{\nu}-m_{\tilde{\nu}}\|^2+\left[
		\frac{1}{2}\text{Tr}\Sigma_{\nu}+
		\frac{1}{2}\text{Tr}\Sigma_{\tilde{\nu}}
		-
		\text{Tr}\left( \Sigma_{\nu} ^\frac{1}{2} \Sigma_{\tilde{\nu}} \Sigma_{\nu} ^\frac{1}{2}  \right)^\frac{1}{2}
		\right].
\end{equation}
Here $\nu$ is the estimated barycenter, $\tilde{\nu}$ is the exact barycenter, and $m_{\nu}, \Sigma_{\nu}$ are the mean and the convariance of the distribution $\nu$.
For barycenter given by pushforward $\nabla g_i \sharp \mu_i$, we
report the weighted average of $\text{BW}_{2}^{2} \textendash \text{UVP}$ scores from $N$ marginal distributions: $\sum_{i=1}^N a_i \text{BW}_{2}^{2} \textendash \text{UVP}\left( \nabla g_i \sharp \mu_i , \tilde{\nu}\right)$.
\jiaojiao{\paragraph{Hyper-parameter choice}
We choose a neural network architecture of $3 \sim 4$ hidden layers of size $1 \sim 2 $ times of input dimension in high dim cases and size $16\sim32$ in 2D/3D cases, with $\textbf{PReLU}$ activation function. We observed that the performance is not sensitive to number of hidden layers, but sensitive to the choice of activation function (ReLU and leaky ReLU do not perform as well as PReLU).
\paragraph{Training time:}
The training time for our method is almost once to twice longer than of CRWB and CWB due to the inner optimization cycles.
The time consumption for CDWB is shorter for small number of samples, as it does not involve  training neural networks.}

\subsection{Learning the Wasserstein Barycenter in 2D and 3D}\label{sec:exp-2d}
\begin{figure}[h]
	\centering
	\begin{subfigure}{.11\textwidth}
		\centering
		\begin{subfigure}{.45\textwidth}
			\centering
			\includegraphics[width=0.95\linewidth]{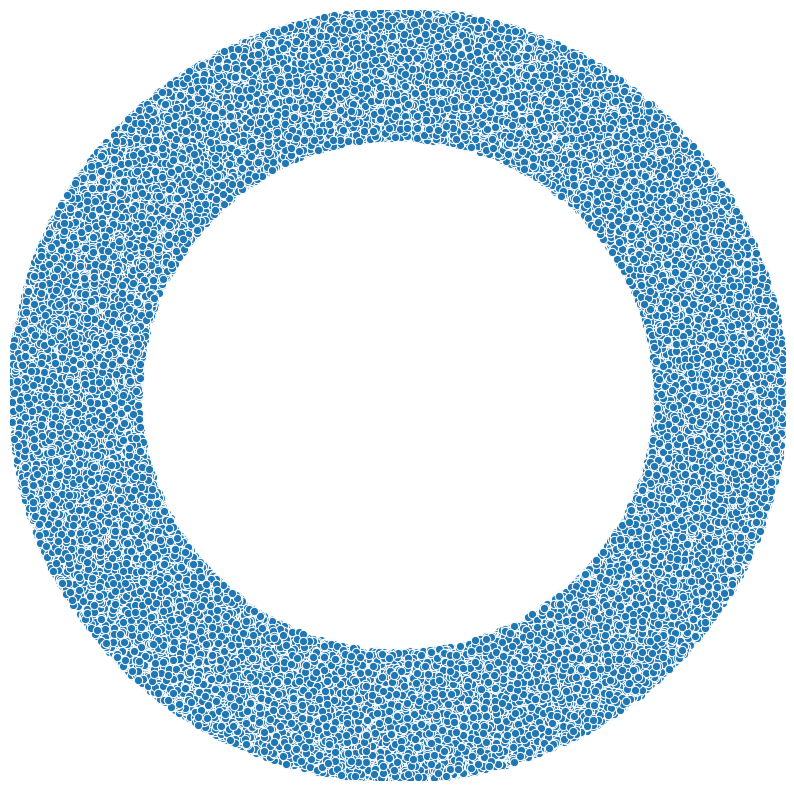}
			\includegraphics[width=0.95\linewidth]{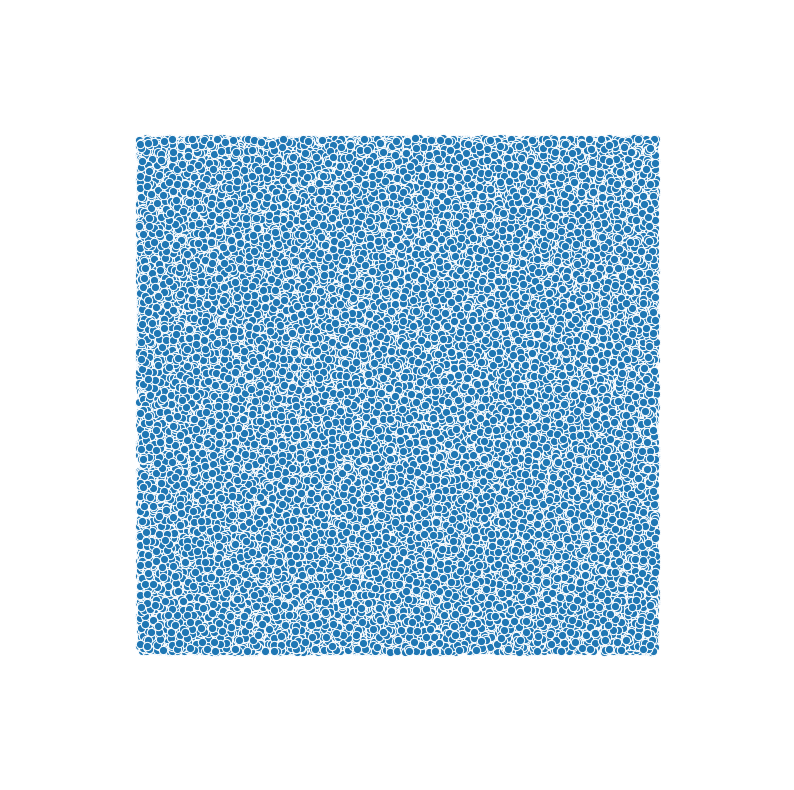}			
		\end{subfigure}
	\end{subfigure}
	\begin{subfigure}{.11\textwidth}
		\centering
		\includegraphics[width=0.95\linewidth]{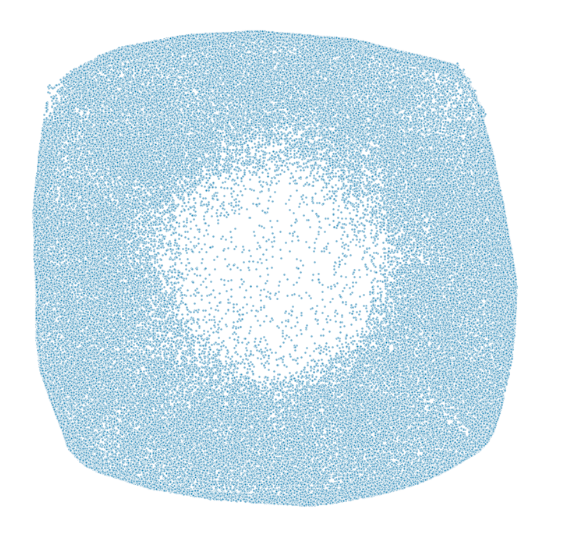}
	\end{subfigure}
	\begin{subfigure}{.11\textwidth}
		\centering
		\includegraphics[width=0.95\linewidth]{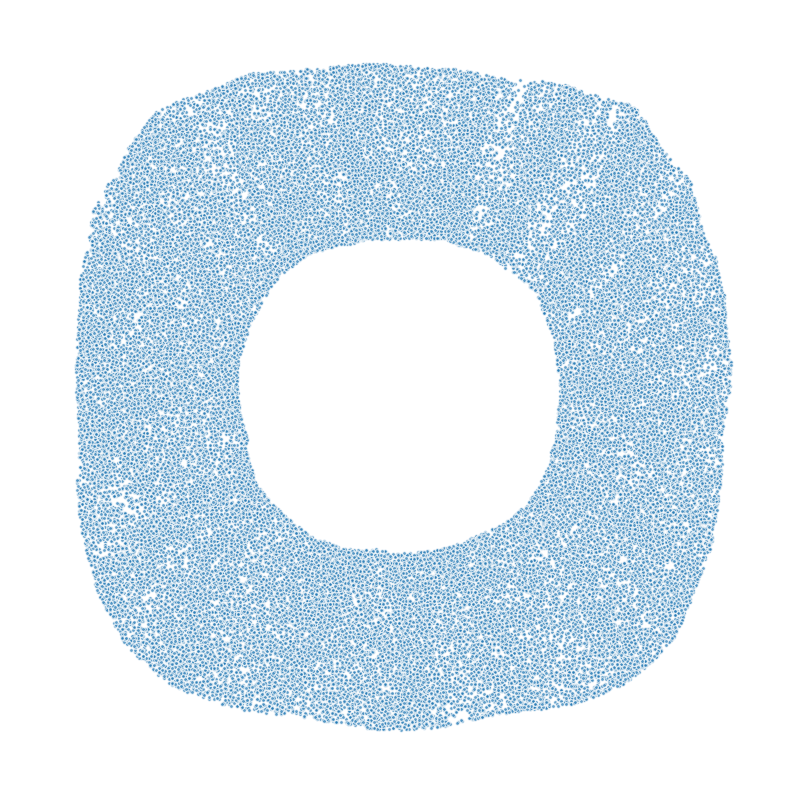}
	\end{subfigure}	
	\begin{subfigure}{.11\textwidth}
		\centering
		\includegraphics[width=0.95\linewidth]{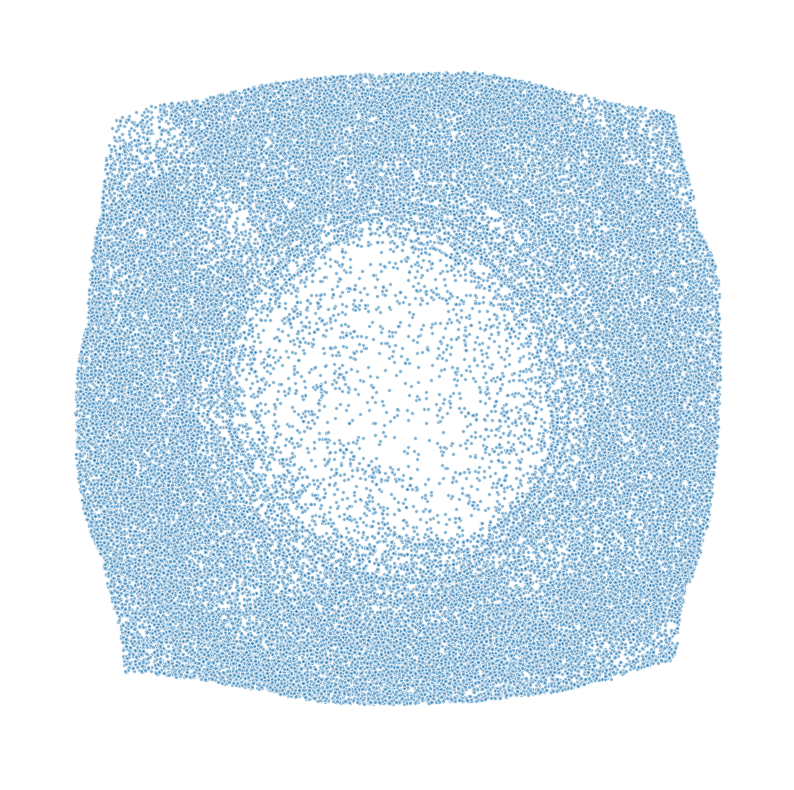}
	\end{subfigure}

	\begin{subfigure}{.11\textwidth}
		\centering
		\includegraphics[width=0.95\linewidth]{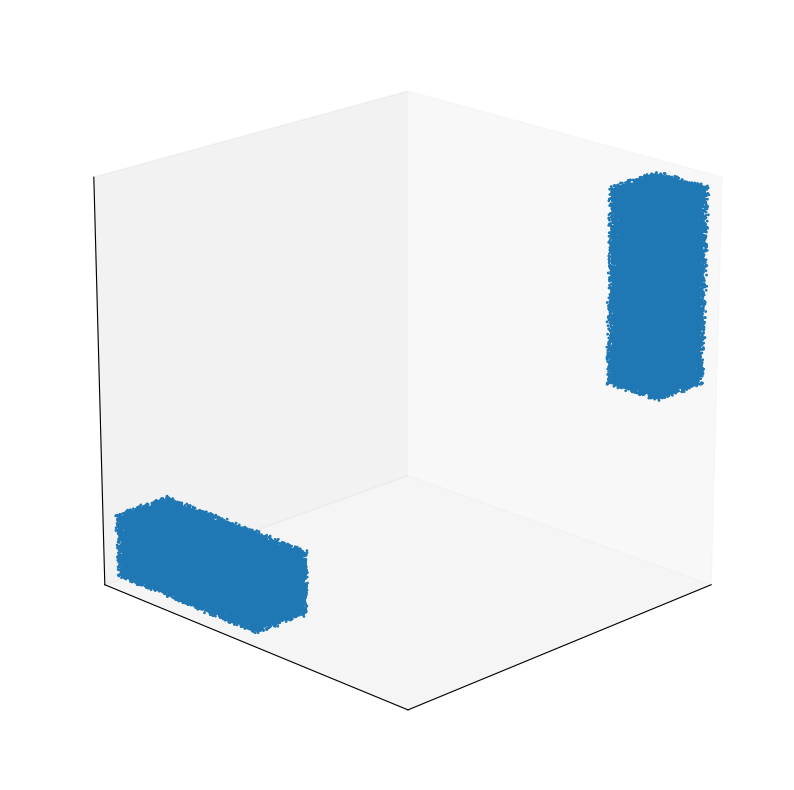}
	\end{subfigure}
	\begin{subfigure}{.11\textwidth}
		\centering
		\includegraphics[width=0.95\linewidth]{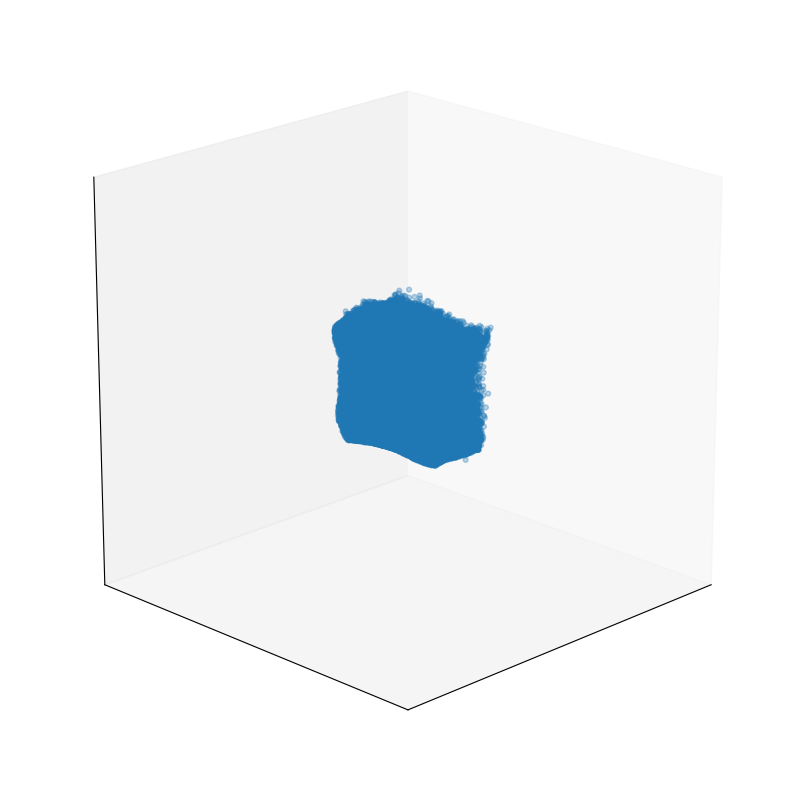}
	\end{subfigure}
	\begin{subfigure}{.11\textwidth}
		\centering
		\includegraphics[width=0.95\linewidth]{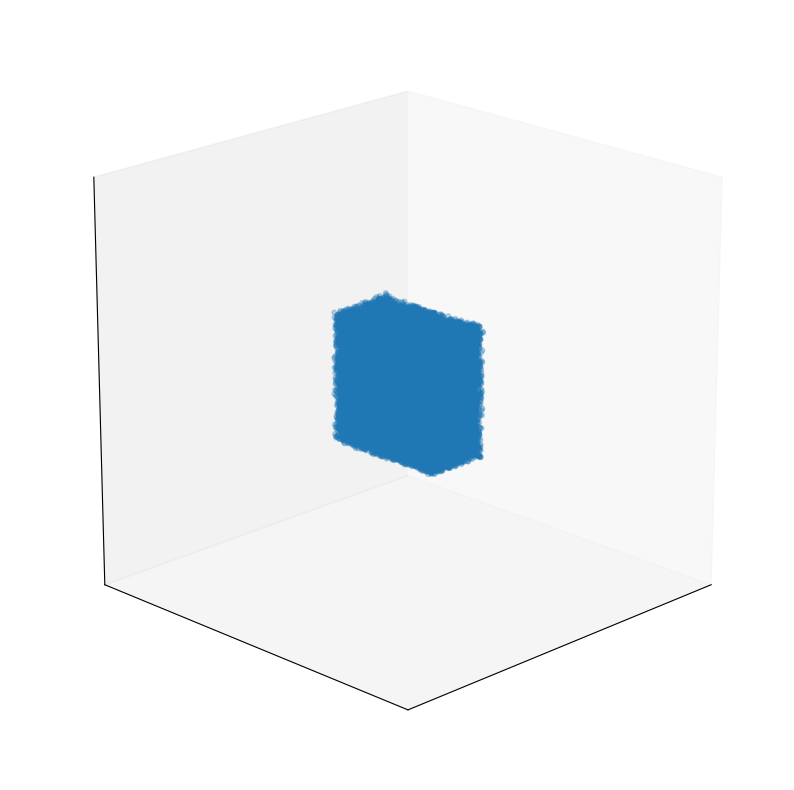}
	\end{subfigure}
	\begin{subfigure}{.11\textwidth}
		\centering
		\includegraphics[width=0.95\linewidth]{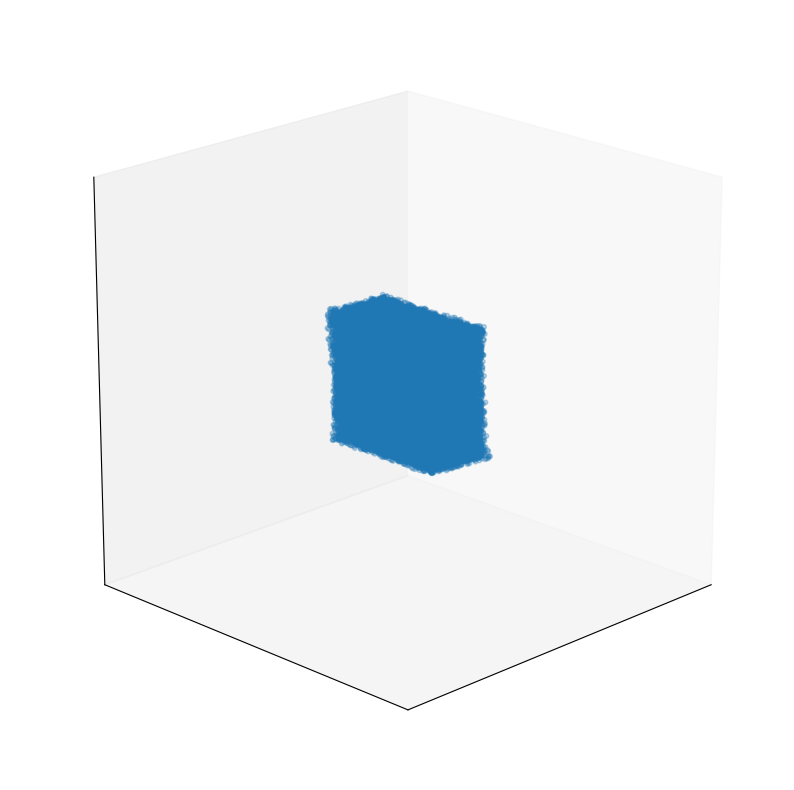}
	\end{subfigure}
	
	\begin{subfigure}{.11\textwidth}
		\centering
		\begin{subfigure}{.45\textwidth}
			\centering
			\includegraphics[width=0.95\linewidth]{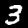}
			\includegraphics[width=0.95\linewidth]{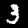}			
		\end{subfigure}
		\caption*{marginals}
	\end{subfigure}
	\begin{subfigure}{.11\textwidth}
		\centering
		\includegraphics[width=0.95\linewidth]{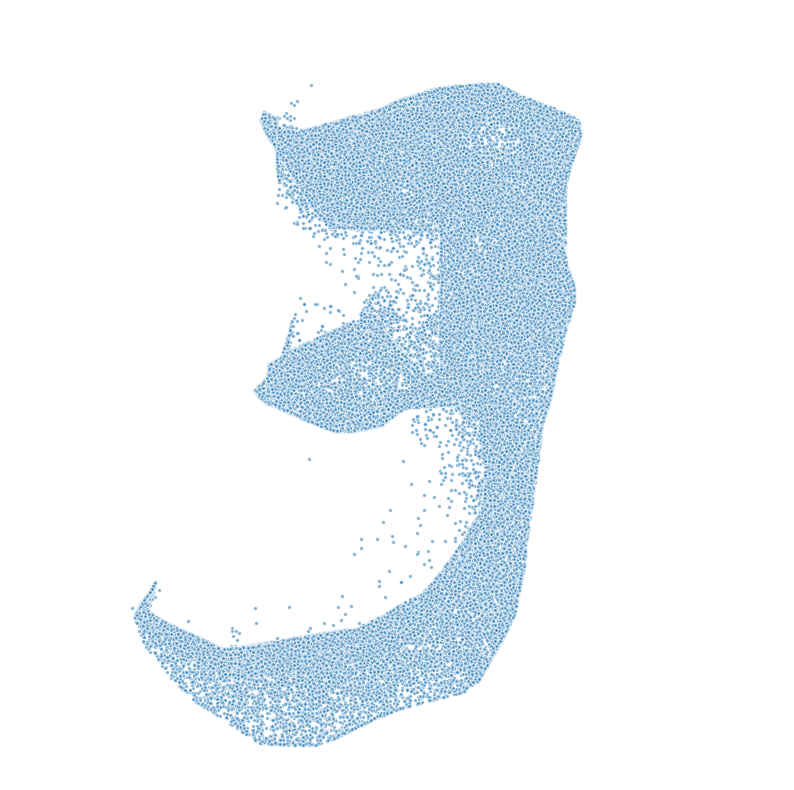}
		\caption{$h \sharp \eta$}
		\label{fig:nwb h}
	\end{subfigure}
	\begin{subfigure}{.11\textwidth}
		\centering
		\includegraphics[width=0.95\linewidth]{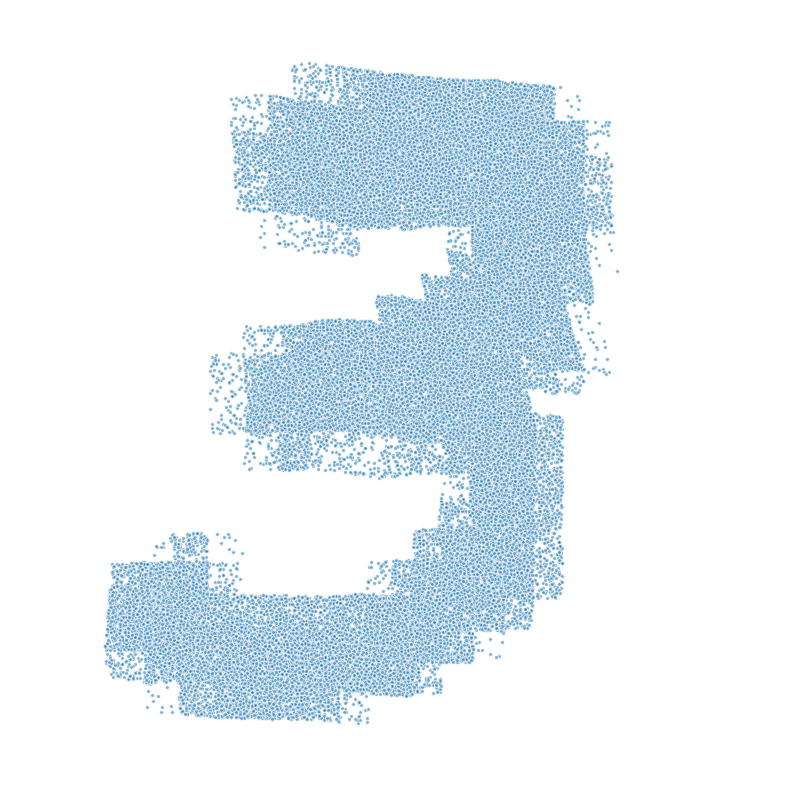}
		\caption{$\nabla g_1 \sharp \mu_1$}
		\label{fig:nwb g1}
	\end{subfigure}
		\begin{subfigure}{.11\textwidth}
		\centering
		\includegraphics[width=0.95\linewidth]{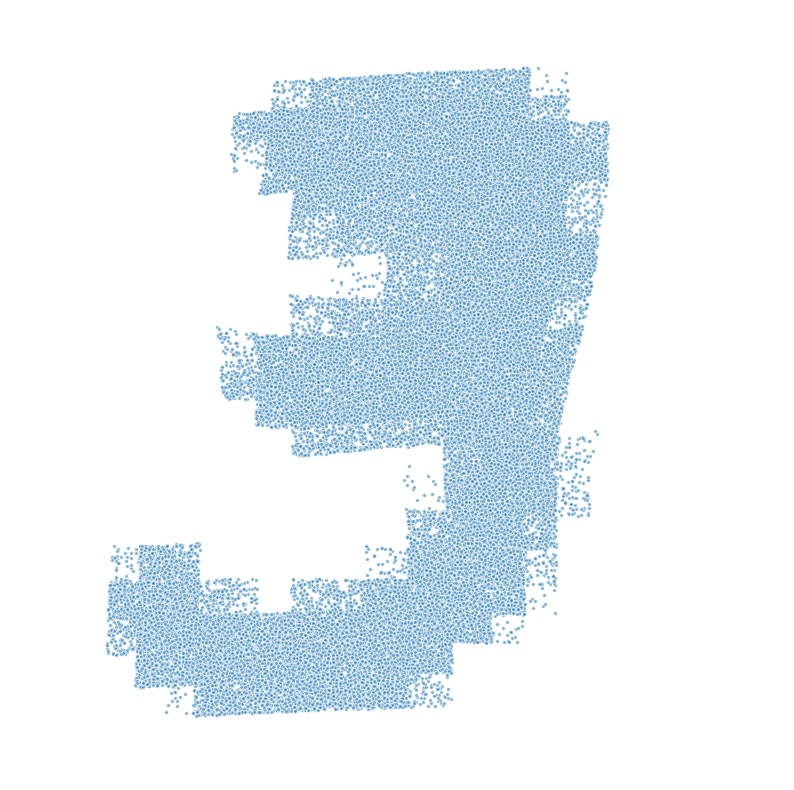}
		\caption{$\nabla g_2 \sharp \mu_2$}
		\label{fig:nwb g2}
	\end{subfigure}
	\caption{Qualitative results of our algorithm (NWB) in 2D and 3D settings. The first column contains the marginals, and the second and third columns contain the barycenter generated through $h\sharp \eta$ and $\nabla g_i \sharp \mu_i$ respectively.
		From top to bottom, the marginal distributions are uniformly supported on a ring and a square; two blocks; and white areas in digit images.
	}
	\label{fig:2or3d}
	\vskip -0.2in
\end{figure}

The qualitative performance of our algorithm in three benchmark  examples is depicted in Figure~\ref{fig:2or3d}. Each example is represented in a row. The first column contains the marginal distributions, and the second and third column contains the learned barycenter through $h\sharp\eta$ and $\nabla g_i \sharp\mu_i$ respectively. It is observed that both representations learn the barycenter qualitatively well, however, representation through $\nabla g_i \sharp\mu_i$ inherits the geometrical properties of the marginal distributions, highlighted with sharp boundaries in the first and second row and pixelated image in the third row. For comparison with CRWB, see~\citet[Figure 1]{li2020continuous}.

\subsection{Scalability with the dimension} \label{sec:exp-high-dim}
\begin{figure}[h]
	\centering
	\begin{subfigure}{0.23\textwidth}
		\centering
		\includegraphics[trim={20pt 0 0 0},width=1\linewidth]{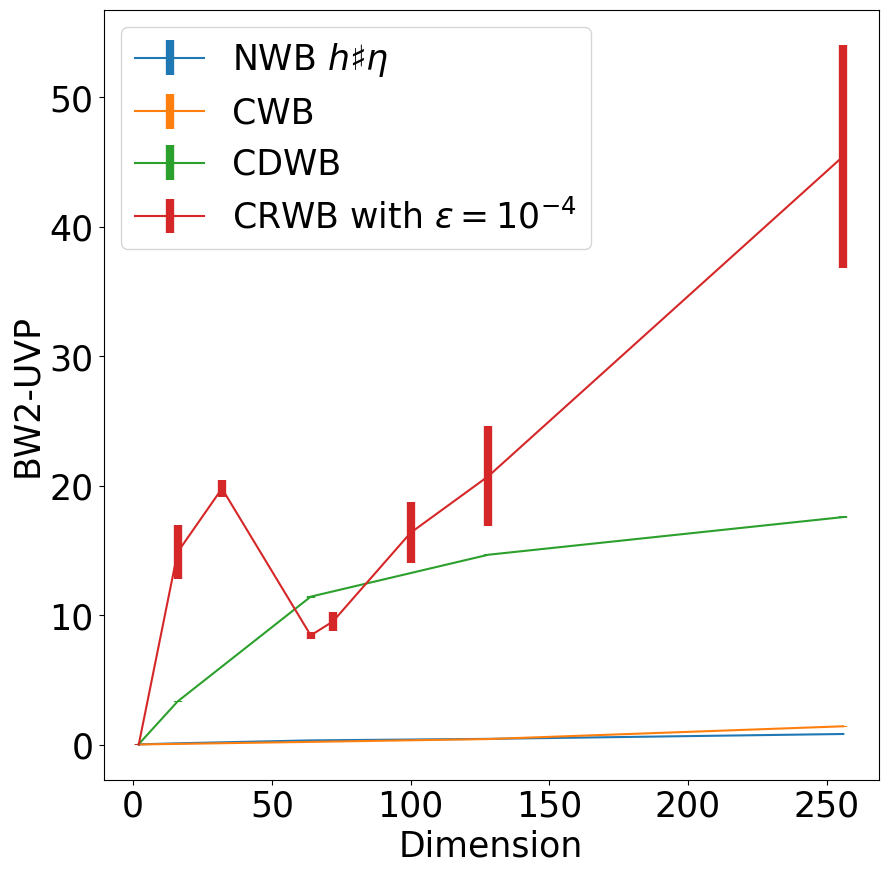}
	\end{subfigure}
	\begin{subfigure}{0.23\textwidth}
		\centering
		\includegraphics[width=1\linewidth]{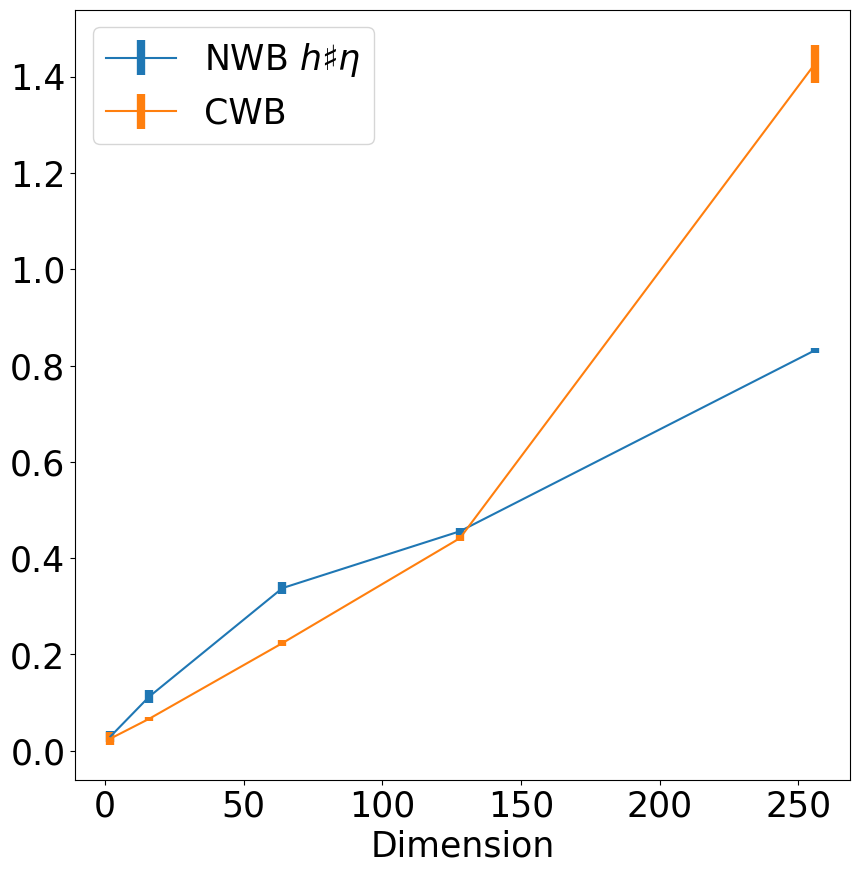}
	\end{subfigure}
	\caption{Numerical result for scalability of the error with dimension for  estimating barycenter of Gaussian  distributions. The error criteria is~\eqref{eq:BW}. We generate $10^4$ samples from barycenter for NWB, CWB, CRWB algorithms and $1500$ samples for CDWB (the maximum number it can generate). The plot on the left includes the results for all four methods and the plot on the right highlights  the  detailed difference between CWB and NWB.
	}
	\label{fig:Gaussian 3 marginal highD results}
	\vskip -0.15in
\end{figure}

{\bf Gaussian:} We study the performance of our proposed algorithm in learning the barycenter of three Gaussian marginal distributions as dimension grows. The Gaussian marginal distributions have zero mean and a random non-diagonal covariance matrix whose conditional number is less than $10$. The exact barycenter of Gaussian distributions is available to serve as the baseline in evaluating the Bures-Wasserstein UVP error criteria~\eqref{eq:BW}.
The results are displayed in Figure~\ref{fig:Gaussian 3 marginal highD results}.
It is observed that the estimation error of NWB and CWB exhibit a slow rate of growth with respect to the dimension compared with CDWB and CRWB. \jiaojiao{The NWB and CWB algorithms are quite comparable in performance, but still different with respect to the optimization landscape. The optimization landscape of CWB is sensitive to the choice of regularization parameters (see Appendix \ref{sec:landscape}). We conjecture that the observed error curve of CWB in Figure 5, is due to the effect of regularization terms in distorting the optimization landscape which becomes more severe as dimension grows.}
Note that the error curve for CRWB contains irregularities and high variance that are probably due to the regularization term.

\begin{figure}[h]
	\centering
	\begin{subfigure}{0.2\textwidth}
		\centering
		\includegraphics[width=1\linewidth]{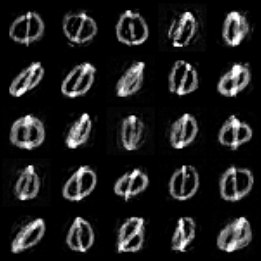}
		\caption{NWB $h \sharp \eta$}
	\end{subfigure}
	\begin{subfigure}{0.2\textwidth}
		\centering
		\includegraphics[width=1\linewidth]{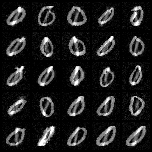}
		\caption{NWB $\nabla g_i \sharp \mu_i$}
	\end{subfigure}

	\begin{subfigure}{0.2\textwidth}
		\centering
		\includegraphics[width=1\linewidth]{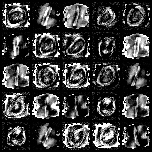}
		\caption{CRWB}
	\end{subfigure}
	\begin{subfigure}{0.2\textwidth}
		\centering
		\includegraphics[width=1\linewidth]{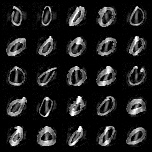}
		\caption{CWB}
	\end{subfigure}
	\caption{Learning the barycenter of MNIST 0 and 1 digits ($784$-dimensional problem)}
	\label{fig:0-1}
	\vskip -0.2in
\end{figure}

{\bf MNIST:} To further investigate the performance of our algorithm in high dimension setting with real dataset, we use the MNIST data set. We consider the task of learning the barycenter of two marginal distributions.
The first marginal $\mu_1$ is an empirical distribution of digit 0 samples and the second marginal $\mu_2$ is of digit 1.
Each image has 28 $\times$ 28 pixels, yielding a 784-dimensional problem. The result of learning the barycenter is depicted in Figure~\ref{fig:0-1}.
Both our algorithm NWB and CWB give reasonable results, with slightly sharper boundaries in NWB, whereas CRWB does not perform well in this high dimension setting. Note that the images in panel (a) are genuinely new samples generated using the trained generator, while the images in other panels are pushforward of marginal samples.

\begin{figure}[h]
	\centering
	\begin{subfigure}{0.2\textwidth}
		\centering
		\includegraphics[width=1\linewidth]{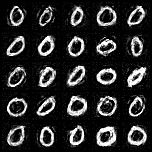}
		\caption{backward to '0'}
		\label{fig:backward0}
	\end{subfigure}
	\begin{subfigure}{0.2\textwidth}
		\centering
		\includegraphics[width=1\linewidth]{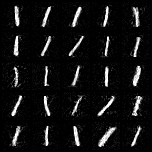}
		\caption{backward to '1'}
		\label{fig:backward1}
	\end{subfigure}
	\caption{Generating digit 0 and 1 from random input $Z \sim \mathcal{N} (\b0,I)$ using our architecture with the map $\nabla f_i( h(Z))$.}
	\label{fig:backward}
	\vskip -0.2in
\end{figure}

In order to demonstrate the inner-workings of our algorithm and its ability to learn the structure of barycenter, we implement the following experiment. We generate fresh samples from the barycenter using the generator $h(Z)$, where $Z \sim \mathcal{N} (\b0,I)$, and push-forward it through the maps $\nabla f_1(h(Z))$ and $\nabla f_2(h(Z))$. It is expected that through this procedure, we would recover the digits 0 and 1, because $\nabla f_i$ represent the optimal transport map from Barycenter to the marginal $\mu_i$ (see Remark~\ref{remark:map}). The experimental result confirms our expectation as shown in Figure~\ref{fig:backward}. This implies that our proposed framework can serve as a generative model, not only for barycenter distribution, but also for the marginal distributions, by taking a random Gaussian random variable as input and output samples from marginal distributions.


\subsection{Subset posterior aggregation}\label{sec:bayesian}
MCMC Bayesian inference is often carried out on splitted datasets in big data setting. However, the subset posterior distributions need to be merged into a single posterior to reflect the entire dataset property. This subset posterior aggregation scheme has been shown as an advantageous substitute to the full posterior \cite{srivastava2015wasp} \cite{SriLiDun18}. The barycenter of subset posteriors is proved to converge to the full data posterior \cite{SriLiDun18}.
Similar to \cite{li2020continuous}, we consider the Poisson regression for predicting the hourly number of bike hires with predictors such as the season and the weather conditions\footnote{\url{http://archive.ics.uci.edu/ml/datasets/Bike+Sharing+Dataset}}. We consider the posterior on the 8-dimensional coefficients for the Poisson regression model. We randomly split the data into 5 equal-size subsets and simultaneously use the stochastic approximation trick~\cite{minsker2014scalable} to promise the subset posterior samples do not vary consistently from the full posterior in covariance. We obtain $10^5$ samples from each subset posterior using the PyMC3 library \cite{pymc3}.

We use the full posterior samples as the ground truth and report the Bures-Wasserstein UVP to compare the estimated barycenter and the ground truth.
The results are shown in Table \ref{tab:bayesian_inference}. All methods approach the true barycenter well (UVP $< 1 \%$) and the performance of NWB is better than or on par with existing algorithms.
\begin{table}[H]
	\caption{Comparison of UVP for recovered barycenters in our subset posterior aggregation task}
	\label{tab:bayesian_inference}
	\begin{center}
		\begin{small}
			\begin{sc}
				\begin{tabular}{lcccc}
					\toprule
					Metric                                                & NWB $h\sharp \eta$  & CDWB & CWB  & CRWB \\
					\midrule
					$\mathrm{BW}_{2}^{2} \textendash \mathrm{UVP} ,{\%} $ & 0.06 & 0.26 & 0.07 & 1.67 \\
					\bottomrule
				\end{tabular}
			\end{sc}
		\end{small}
	\end{center}
	\vskip -0.1in
\end{table}
\vspace{-0.6cm}
\subsection{Color palette averaging} \label{sec:color}
Color transfer is a  method to change the appearance of a source image according to the color pattern of a target image \cite{reinhard2001color}. Given several images, we can solve for Wasserstein barycenter to aggregate color palettes of images to achieve color transfer among them. Given an RGB image $\mathcal{I} $, its color palette is the empirical distribution $\mu(\mathcal{I})= \sum_{k=1}^K \frac{1}{K} \delta_{p_k}$ where $\{p_k\}$ represents the pixels $ \in [0,1]^3$ and $\delta_{p_k}$ is the Dirac distribution concentrated on $p_k$. In our example, each image contains $1980 \times 1024$ pixels, so the number of samples for each marginal distribution is more than 2 million. \footnote{The pictures are downloaded from \url{https://wallpaperaccess.com/}}.

\begin{figure}[h]
	\centering
	\begin{subfigure}{.5\textwidth}
		\centering
		\includegraphics[width=0.3\linewidth]{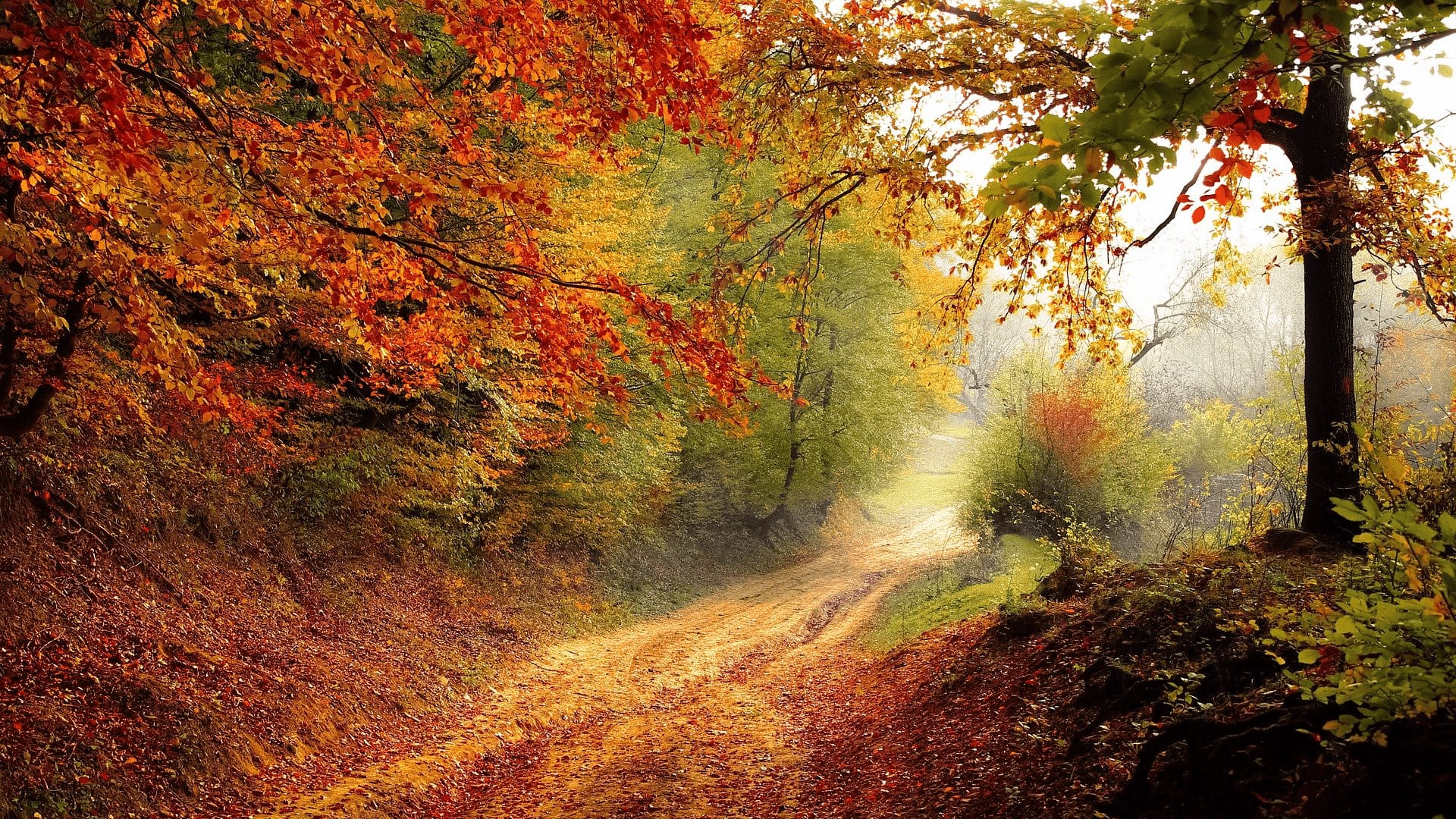}
		\includegraphics[width=0.3\linewidth]{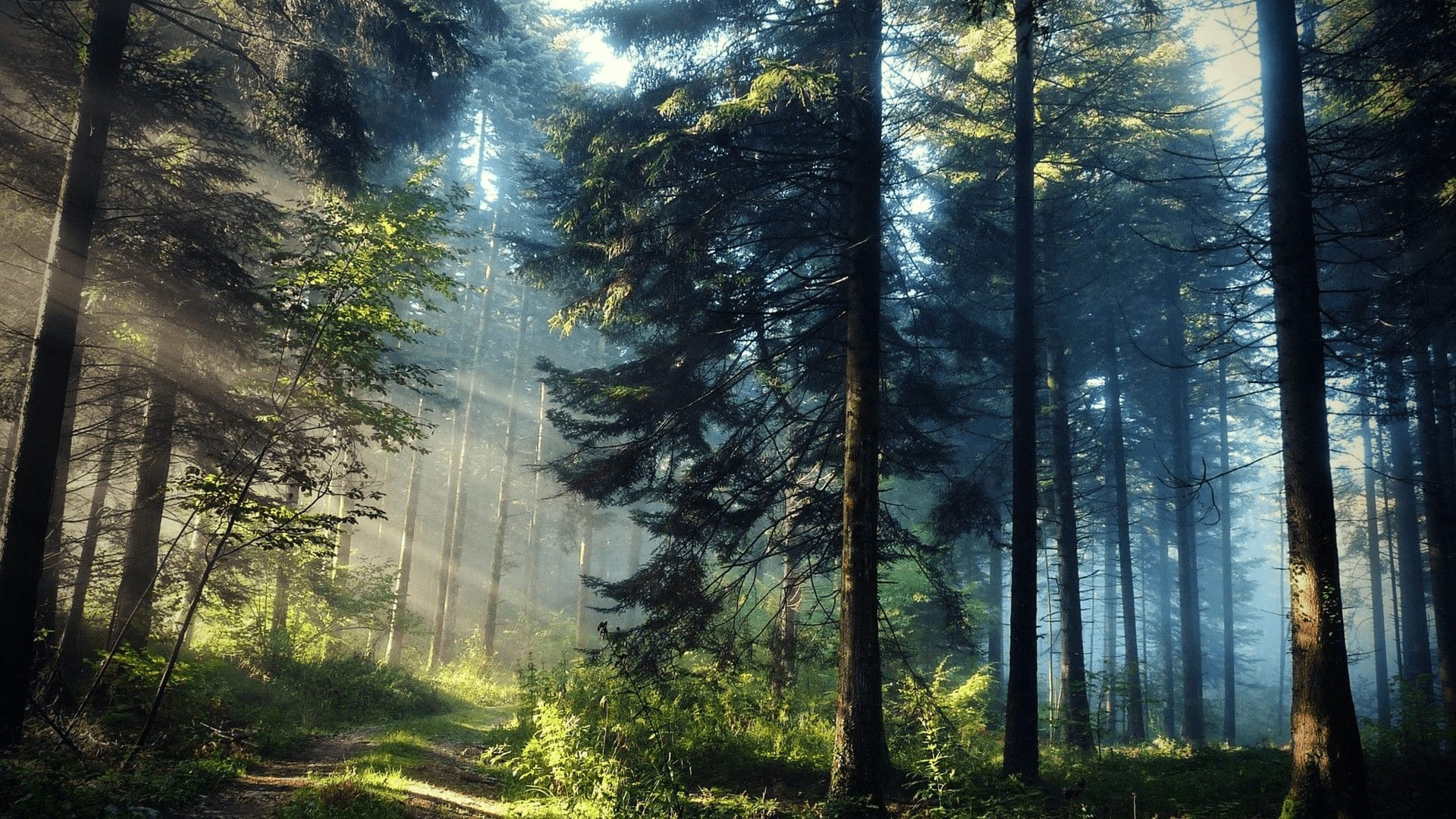}
		\includegraphics[width=0.3\linewidth]{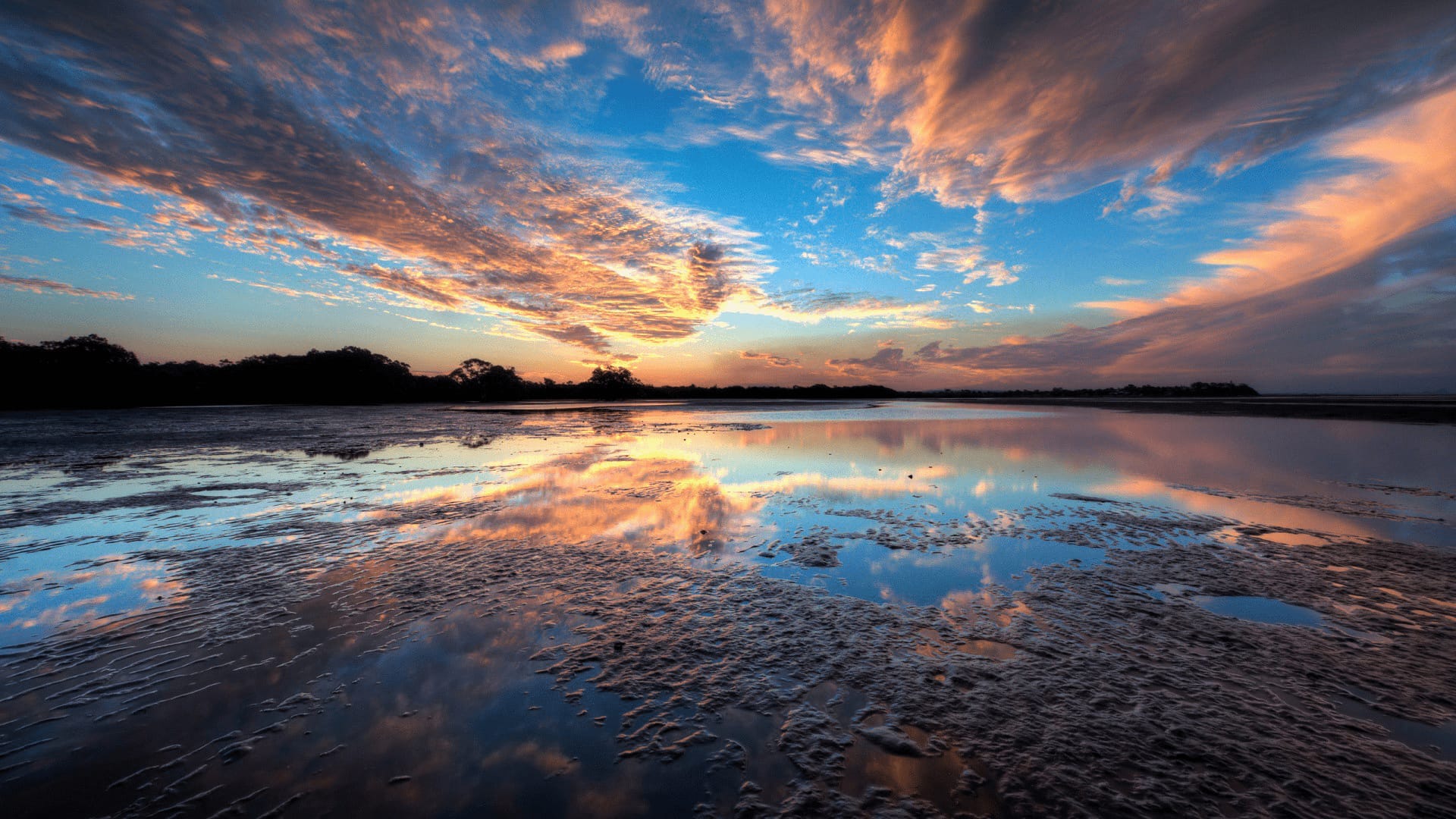}
		\caption*{Source images $\{\mI_i\}$}
	\end{subfigure}
	\begin{subfigure}{.5\textwidth}
		\centering
		\includegraphics[width=.3\linewidth]{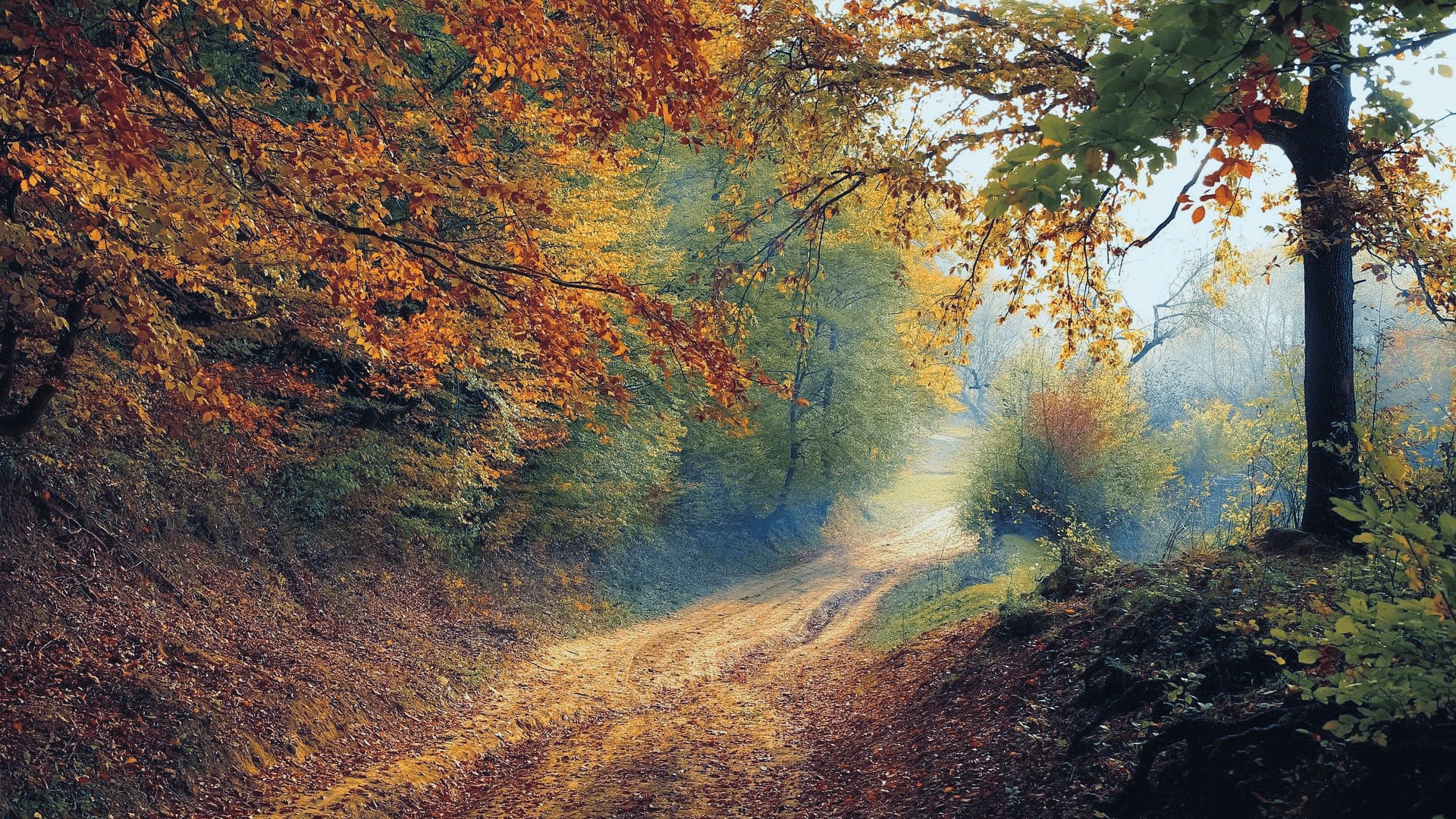}
		\includegraphics[width=.3\linewidth]{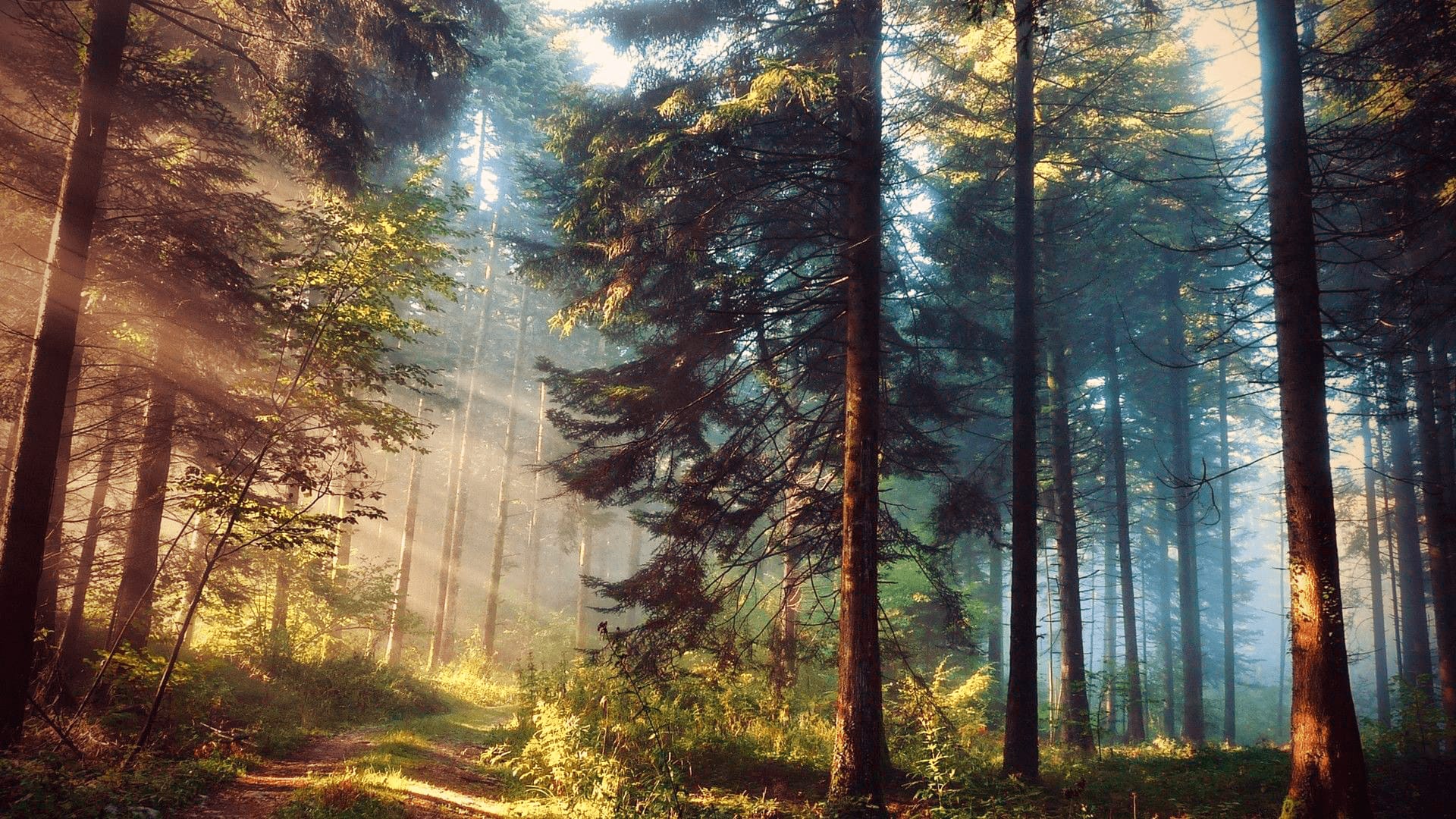}
		\includegraphics[width=.3\linewidth]{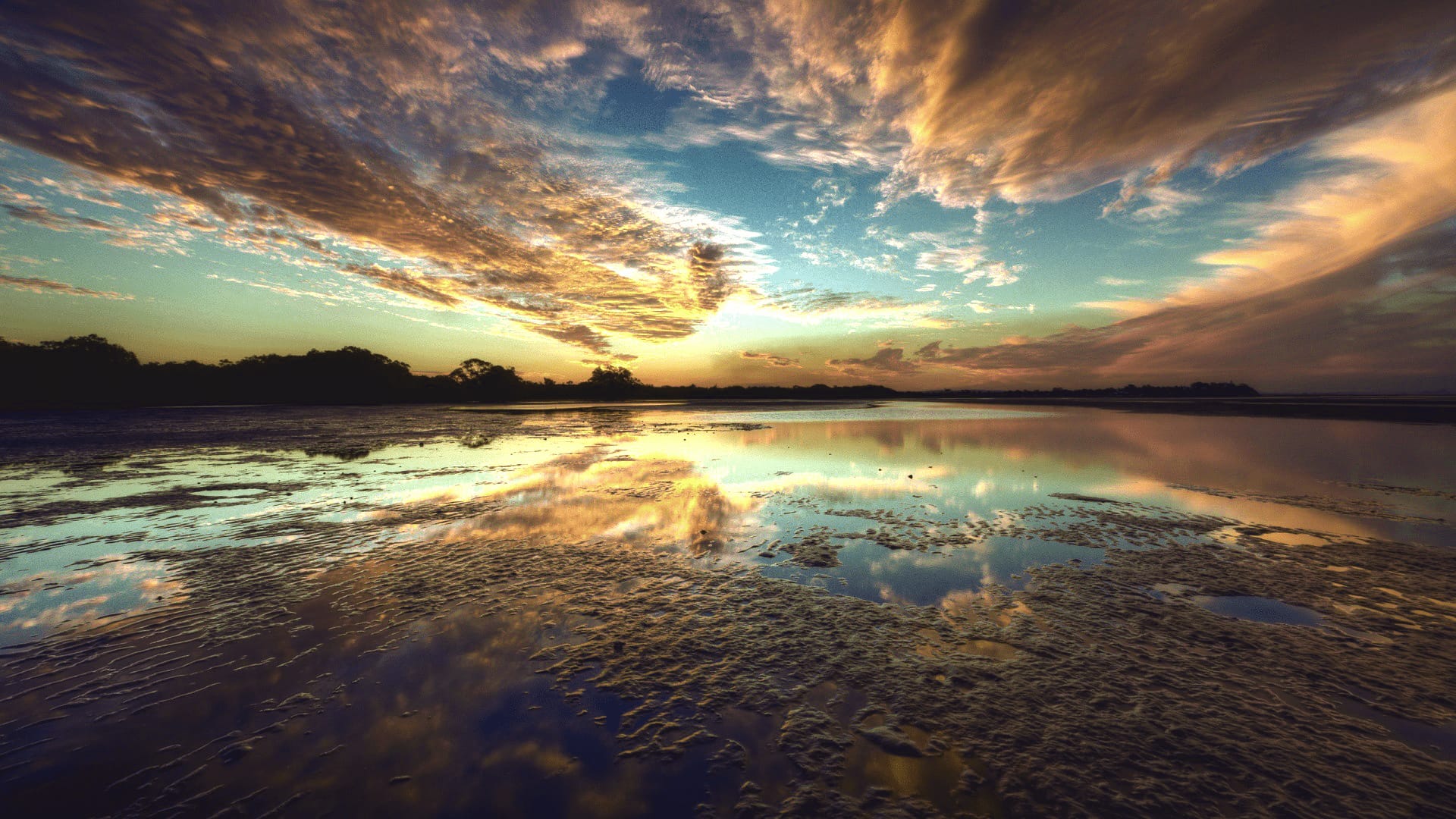}
		\caption*{Pushforward images}
	\end{subfigure}
	\caption{Qualitative results by pixel-wise pushforward of the source images }
	\label{fig:pushforward images}
	\vskip -0.2in
\end{figure}

\begin{figure}[h]
	\centering
	\begin{subfigure}{.5\textwidth}
		\centering
		\includegraphics[width=0.3\linewidth]{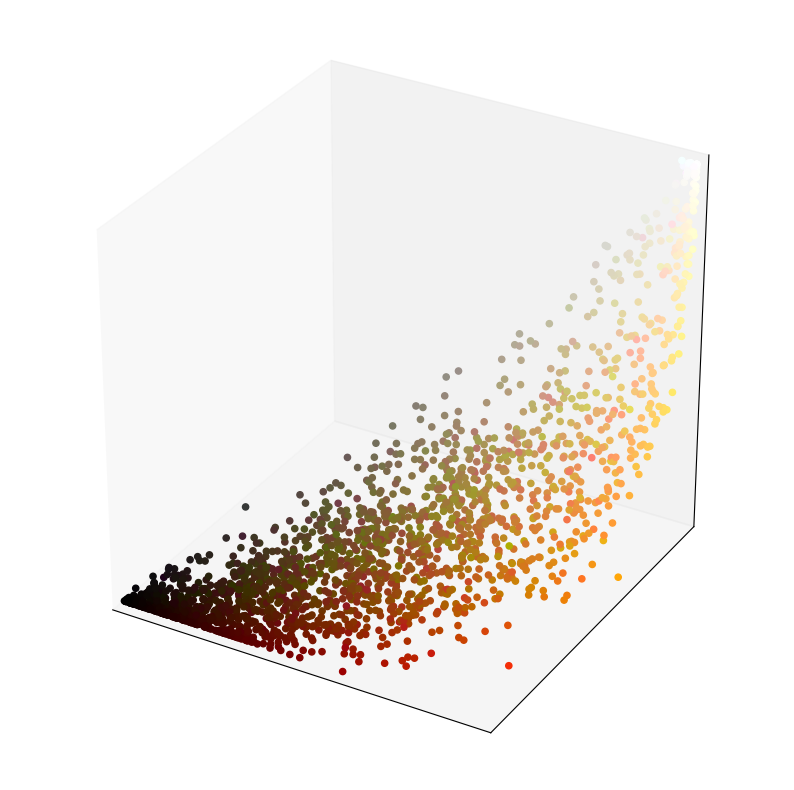}
		\includegraphics[width=0.3\linewidth]{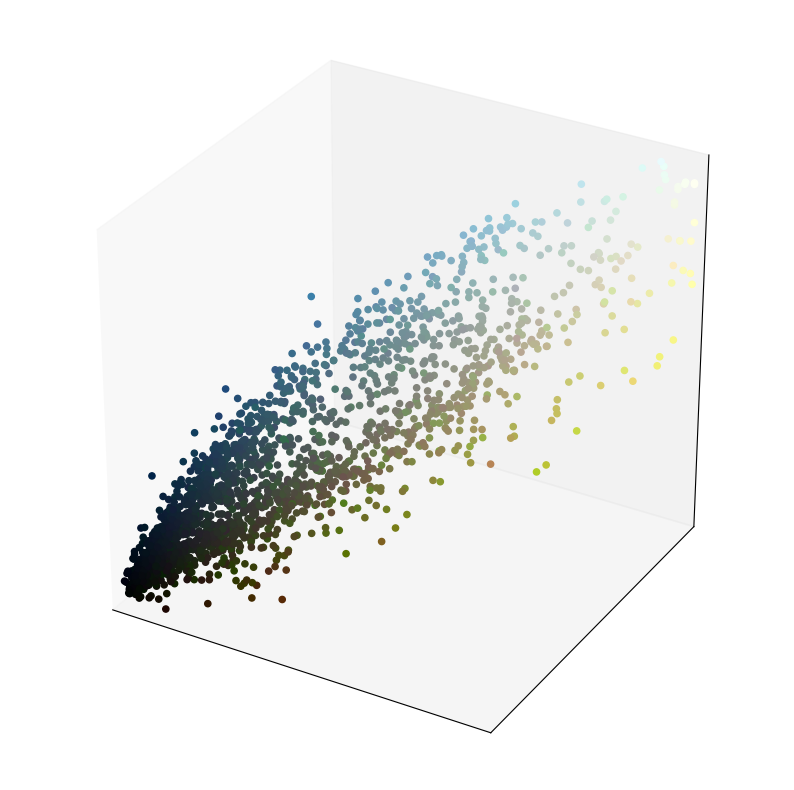}
		\includegraphics[width=0.3\linewidth]{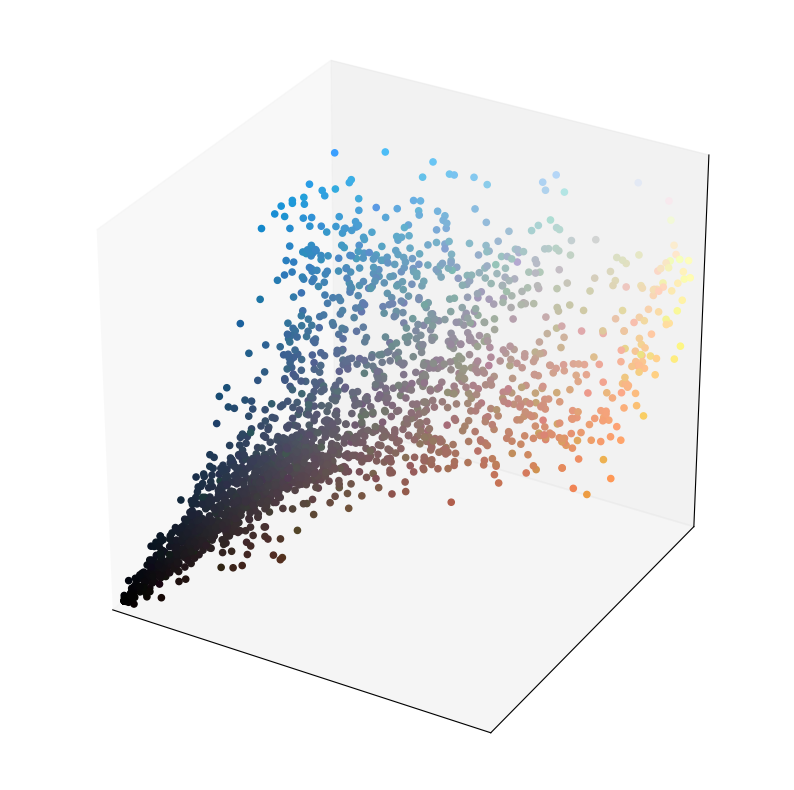}
		\caption{ source images: $\{\mu(\mI_i)\}$}
\end{subfigure}

	\begin{subfigure}{.5\textwidth}
		\centering
		\begin{subfigure}{.24\textwidth}
			\centering
			\includegraphics[width=1\linewidth]{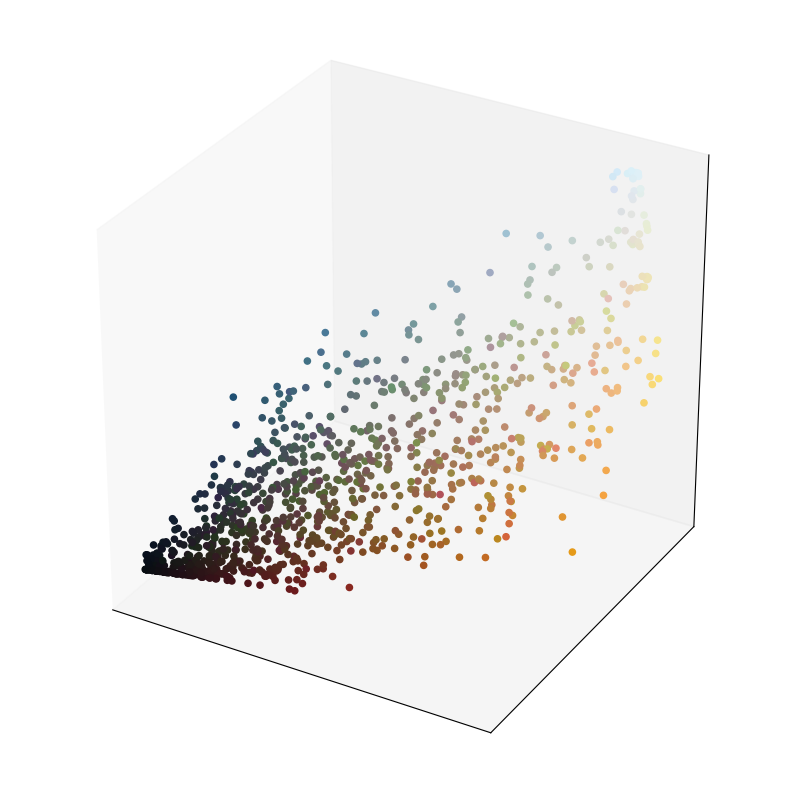}
			\caption{ $\nabla g_1 \sharp \mu(\mI_1)$}
			\label{fig:rgb cloud pushf1}
		\end{subfigure}
		\begin{subfigure}{.24\textwidth}
			\centering
			\includegraphics[width=1\linewidth]{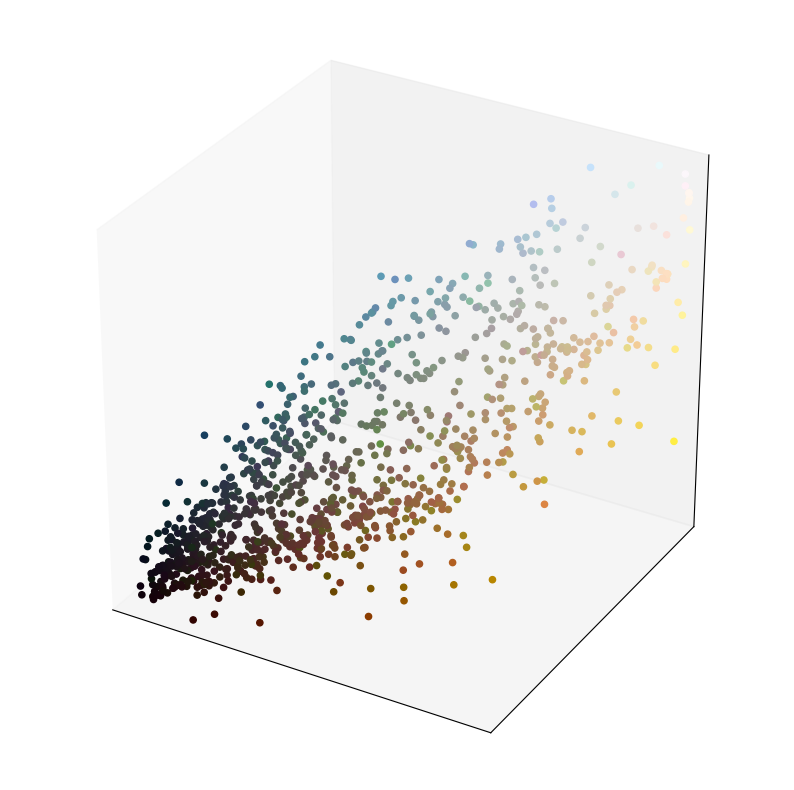}
			\caption{ $\nabla g_2 \sharp \mu(\mI_2)$}
		\end{subfigure}
		\begin{subfigure}{.24\textwidth}
			\centering
			\includegraphics[width=1\linewidth]{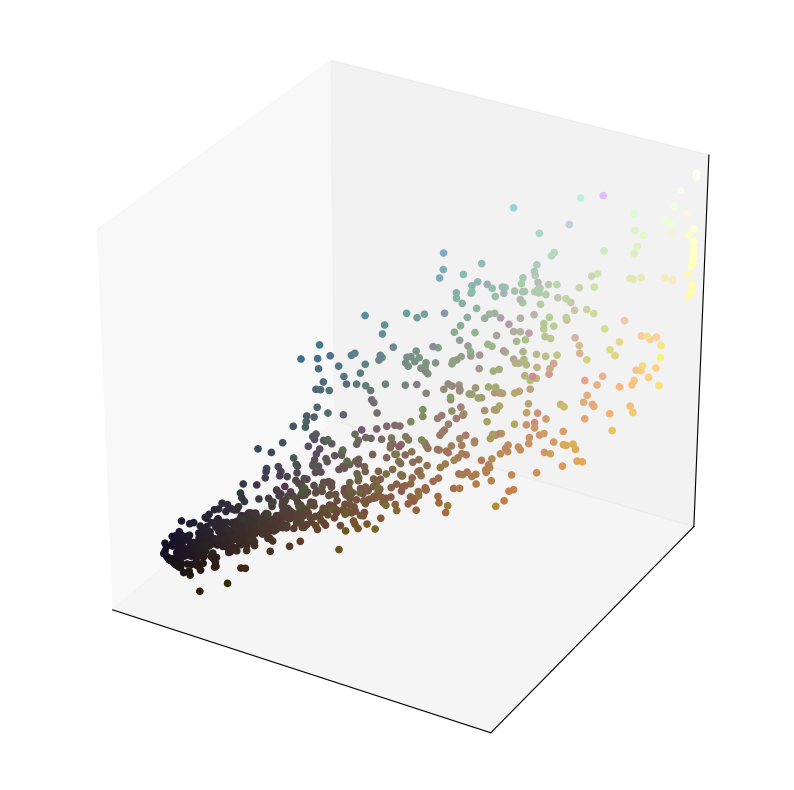}
			\caption{ $\nabla g_3 \sharp \mu(\mI_3)$}
		\end{subfigure}
		\begin{subfigure}{.24\textwidth}
			\centering
			\includegraphics[width=1\linewidth]{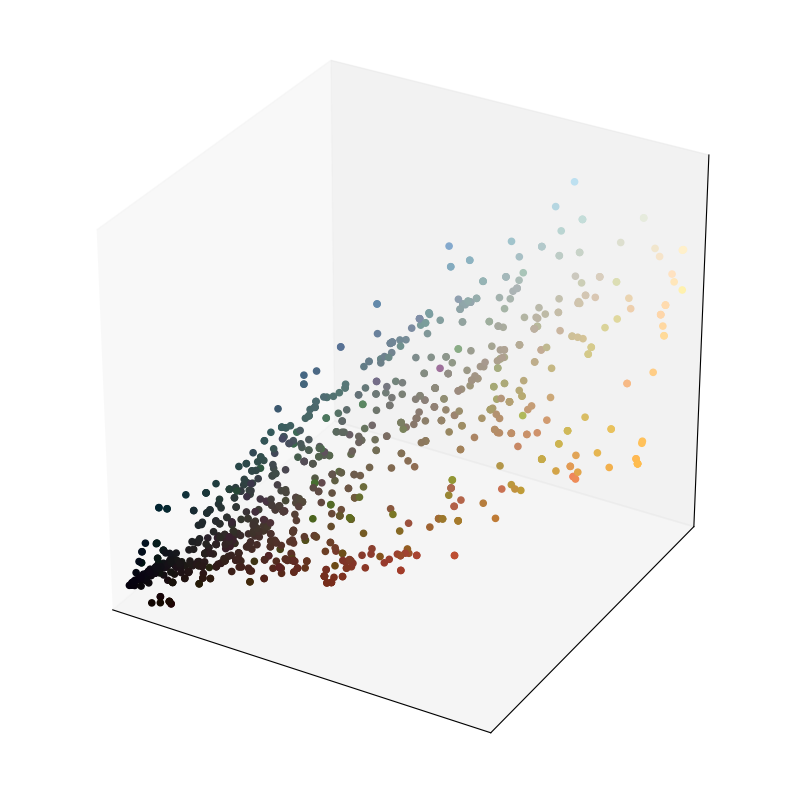}
			\caption{ $h \sharp \eta$ }
			\label{fig:rgb cloud generative nn}
		\end{subfigure}
	\end{subfigure}
	\caption{Color palettes of source images and barycenter}
	\label{fig:color palettes}
	\vskip -0.2in
\end{figure}

In Figure \ref{fig:pushforward images}, the upper panel shows the original images $\{\mI_1, \mI_2, \mI_3 \}$, and the bottom panel shows pixel-wise ``pushforward" images.
Figure \ref{fig:color palettes} shows the RGB cloud to visualize the color palettes of images. In Figure \ref{fig:pushforward images}, the appearance of the pushforward images are different from the source images thanks to the color averaging: the leaves in the first picture become greener and darker, the sunbeams in the second picture become more red, and the sky in the last picture receives an orange color toning.


\vspace{-0.1cm}
\subsection{Serving as a Generative Adversarial Model in the one marginal setting}\label{sec:gan}
\begin{figure}[h]
	\centering
	\begin{subfigure}{.23\textwidth}
		\centering
		\includegraphics[width=1\linewidth]{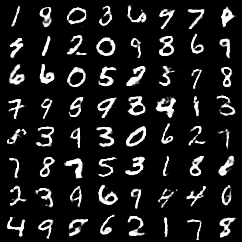}
		\caption{NWB $h \sharp \eta$}
		\label{fig:ours mnist0-9}
	\end{subfigure}
	\begin{subfigure}{.23\textwidth}
		\centering
		\includegraphics[width=1\linewidth]{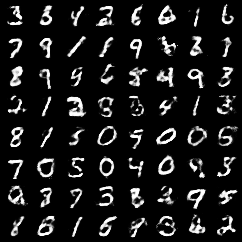}
		\caption{WGAN}
		\label{fig:mnist wgan}
	\end{subfigure}

	\begin{subfigure}{.23\textwidth}
		\centering
		\includegraphics[width=1\linewidth]{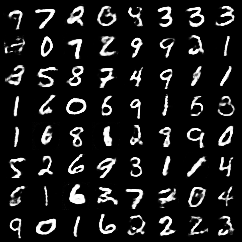}
		\caption{WGAN-GP}
		\label{fig:mnist wgan-gp}
	\end{subfigure}
	\begin{subfigure}{.23\textwidth}
		\centering
		\includegraphics[width=1\linewidth]{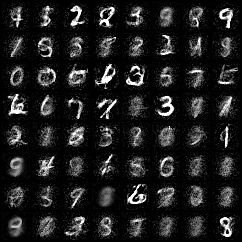}
		\caption{W2GN}
		\label{fig:mnist w2gn}
	\end{subfigure}
	\caption{Performance of our algorithm as GAN in single marginal case: learning the MNIST digit dataset}
	\label{fig:mnist GAN}
	\vskip -0.2in
\end{figure}
We study the performance of our proposed algorithm in the case of one marginal distribution, where it behaves as a  generative adversarial network using the $W_2$ metric. For comparison, we use WGAN \cite{pmlr-v70-arjovsky17a} and WGAN-GP \cite{gulrajani2017improved}, which are based on $W_1$ metic, and W2GN \cite{korotin2021wasserstein}, which is based on $W_2$ metric. Note that our goal is not to provide a competitive GAN algorithm, but to demonstrate the ability of our algorithm in performing as GAN.


We first consider an example of learning a Gaussian mixture model with $10$  components shown in Figure \ref{fig:Gaussian mixture 1 marginal 2D results}. 
It can be seen that NWB avoids mode collapse.
We then investigate the performance of our algorithm NWB in learning MNIST digits dataset (784 dim, 60000 sample size). From Figure \ref{fig:mnist GAN}, it is observed that our algorithm could output all the digits from 0 to 9 without mode collapse and the quality is on par with WGAN and WGAN-GP.
W2GN seeks an optimal transport map from 784 dim standard Gaussian to MNIST; the generated digits are of poor quality. Note that in \citet{korotin2021wasserstein} W2GN was tested on MNIST dataset but the optimal transport is addressed in the feature/latent space (see Section 5.2, Section C.7 in \citet{korotin2021wasserstein}), which is of much lower dimension than the pixel space (784 dim).




\vspace{-0.3cm}
\section{Conclusion}\label{sec:concl}
During the last decade, many algorithms have been proposed for Wasserstein Barycenter estimation. A majority of these algorithms are designed for discrete setting (either discretization of space or discretization of distribution from samples). There are several algorithms that are designed for semi-discrete setting, in the sense that even though the marginal distributions are continuous, the barycenter computed from the algorithms is supported on finite points. More recently, two algorithms \citep{korotin2021continuous,li2020continuous} have been proposed to approximate the barycenter by learning the optimal transport maps from the marginal distributions to the barycenter using samples from the marginal distributions.
Compared to all these existing Wasserstein Barycenter estimation algorithm, the NWB algorithm we develop is the only algorithm that gives a continuous representation of the barycenter through a generative model and is capable of generating infinitely many samples from the barycenter.

\section*{Acknowledgments}
The authors would like to thank the anonymous reviewers for useful comments. JF and YC are supported in part by grants NSF CAREER ECCS-1942523, NSF CCF-2008513, and NSF CCF-1740776. \jiaojiao{The authors also thank Qinsheng Zhang for fruitful discussions.}

\bibliography{w2_barycenter}
\bibliographystyle{icml2021}

\onecolumn
\clearpage

\appendix
\section{Proof of Proposition 1}
 {\it Proof.} For fixed $\nu$ that is absolutely continuous with respect to the Lebesgue measure, and $f_i, i = 1,\ldots, N$, the solution to the inner-loop minimization problems over $g_i$ are clearly $g_i^\star = f_i^*, i = 1,\ldots, N$. The problem \eqref{eq:precise final target} then becomes
\[
  \underset{\nu}{\min} \sum_{i=1}^{N} a_i \left\{\underset{f_i \in \textbf{CVX}} {\sup}\{-\mathbb{E}_{\nu}[f_i(X)]-\mathbb{E}_{\mu_i}[f_i^*(Y)]\} + C_{\nu, \mu_i}\right\}.
\]
In view of \eqref{eq:extract two momentum}, it boils down to
\[
  \min_\nu ~ \sum_{i=1}^{N} a_i W_{2}^{2}\left(\nu, \mu_{i}\right),
\]
which is exactly the Wasserstein barycenter problem \eqref{eq:initial target}. Since all the marginal distributions $\mu_i$ are absolutely continuous with respect to the Lebesgue measure, their barycenter exists and is unique. This completes the proof. $\hfill\square$

\section{Neural Wasserstein Barycenter-F}
We consider a more challenging Wasserstein barycenter problem with free weights. More specifically, given a set of marginal distribution $\mu_i,\, i = 1, \ldots, N$, we aim to compute their Wasserstein barycenter for all the possible weights. Of course, we can utilize Algorithm \ref{al:three loops average weight} to solve fixed weight Wasserstein barycenter problem \eqref{eq:FICNN final object} for different weight $a$ separately. However, this will be extremely expensive if the number of weights is large. It turns out that Algorithm \ref{al:three loops average weight} can be adapted to obtain the barycenters for all weights in one shot. \begin{wrapfigure}[10]{r}{7cm}
  \centering
  \includegraphics[width=0.8\linewidth]{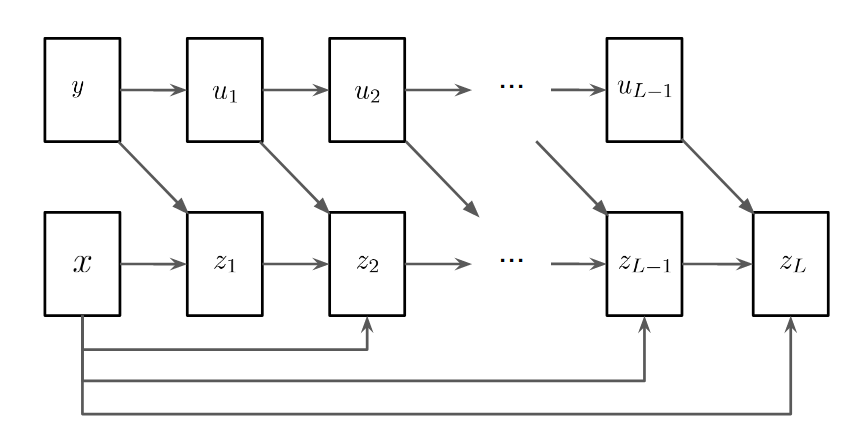}
  \caption{Partially ICNN structure }
  \label{fig:PICNN}
\end{wrapfigure} To this end, we include the weight $a$ as an input to all the neural networks $f_i,g_i$ and $h$, rendering maps $h(z,a;\theta_h), f_i(x,a;\theta_{f_i}), g_i(y,a;\theta_{g_i})$. For each fixed weight $a$, the networks $f_i,g_i$ and $h$ with this $a$ as an input solves the Barycenter problem with this weight. Apparently, $f_i, g_i$ are only required to be convex with respect to samples, not the weight $a$. Therefore, we use PICNN \citep[Section 3.2]{AmoXuKol17} instead of FICNN for as network architectures. PICNN is an extension of FICNN that is capable of modeling functions that are convex with respect to parts of the variable.The architecture of PICNN is depicted in Figure \ref{fig:PICNN}. It is a $L$-layer architecture with inputs $(x,y)$. Under some proper assumptions on the weights (the feed-forward weights $\{W_l^{(z)}\}$ for $z$ are non-negative) and activation functions of the network, the map $(x, y) \rightarrow f(x, y ; \theta):=z_{L}$ is convex over $x$. We refer the reader to \citep{AmoXuKol17} for more details. The problem then becomes
\begin{equation} \label{eq:weighted problem with PICNN}
  \underset{h}{\min} \underset{f_{i} \in \textbf{PICNN}}{\sup}\,
  \underset{{g_{i} \in \textbf{PICNN}} }{\inf}
  \mE_U\{-\mathbb{E}_{\eta}[f_i(h(Z,a),a)]-\mathbb{E}_{\mu_i}[\langle Y, \nabla g_i(Y,a)\rangle-f_i(\nabla g_i(Y,a),a)]
  + \frac{1}{2}\mE_\eta [\|h(Z,a)\|^2]\}
\end{equation}
where $U$ is a probability distribution on the probability simplex, from which the weight $a$ is sampled. In our experiment, we used uniform distribution, but it can be any distribution that is simple to sample from, e.g., Dirichlet distribution. Effectively, the objective function in \eqref{eq:weighted problem with PICNN} amounts to the total Wasserstein cost over all the possible weights. Our formulation makes it ideal to implement stochastic gradient descent/ascent algorithm and solve the problem jointly in one training. As in the fixed weights setting, the (partial) convexity constraints of $\{g_i\}$ can be replaced by a penalty term. For batch implementation, in each batch, we randomly choose one $a\in U$ and $M$ samples $\{Y_j^i\}$ from $\mu_j$ and $\{Z_j\}$ from $\eta$. The unbiased batch estimation of the objective in \eqref{eq:weighted problem with PICNN} reads
\begin{equation}\label{eq:batchBII}
  \sum_{i=1}^{N}a_i\{
  J(\theta_{f_i}, \theta_{g_i},\theta_{h})+R\left(\theta_{g_i}\right)\}
  + \frac{1}{2M}\sum_{j=1}^{M} ||h(Z_j,a)||^2,
\end{equation}
where $$J\!=\!\frac{1}{M}\sum_{j=1}^{M}
  [f_i\left(\nabla g_i\left(Y_{j}^i,a\right),a\right)\!-\!\left\langle Y_{j}^i, \nabla g_i\left(Y_{j}^i,a\right)\right\rangle
  \!-\!f_i\left(h(Z_j,a),a\right)],$$
and $R\left(\theta_{g_i}\right)\!\!=\!\!\lambda \sum_{W_{l}^{(z)} \in \theta_{g_i}}\left\|\max \left(-W_{l}^{(z)}, 0\right)\right\|_{F}^{2}$.
By alternatively updating $h, f_i, g_i$ we establish Neural Wasserstein Barycenter-F (NWB-F) (Algorithm \ref{alg:BII}).

\begin{minipage}{.9\textwidth}
  \begin{algorithm}[H]
    \caption{Neural Wasserstein Barycenter-F}
    \label{alg:BII}
    \begin{algorithmic}
      \STATE \textbf{Input} Marginal dist. $\mu_{1:N}$, Generator dist. $\eta$, Batch size $M$, weight dist. $U$
      \FOR{$k_3=1,\ldots,K_3$}
      \STATE {Sample $a \sim U$}
      \STATE {Sample batch {$\left\{Z_{j}\right\}_{j=1}^{M} \sim \eta$}}
      \STATE {Sample batch {$\left\{Y_{j}^{i}\right\}_{j=1}^{M} \sim \mu_i$}}
      \FOR{$k_2=1,\ldots,K_2$}
      \FOR{$k_1=1,\ldots,K_1$}
      \STATE {Update all $\theta_{g_i}$ to decrease \eqref{eq:batchBII}}
      \ENDFOR
      \STATE {Update all $\theta_{f_i}$ to increase \eqref{eq:batchBII}}
      \STATE{Clip: $W_l^{(z)} =\max(W_l^{(z)}, 0)$ for all $\theta_{f_i}$}
      \ENDFOR
      \STATE{Update $\theta_h$ to decrease \eqref{eq:batchBII}}
      \ENDFOR
    \end{algorithmic}
  \end{algorithm}%
\end{minipage}

\begin{figure}[h]
  \centering
  \includegraphics[width=0.95\linewidth]{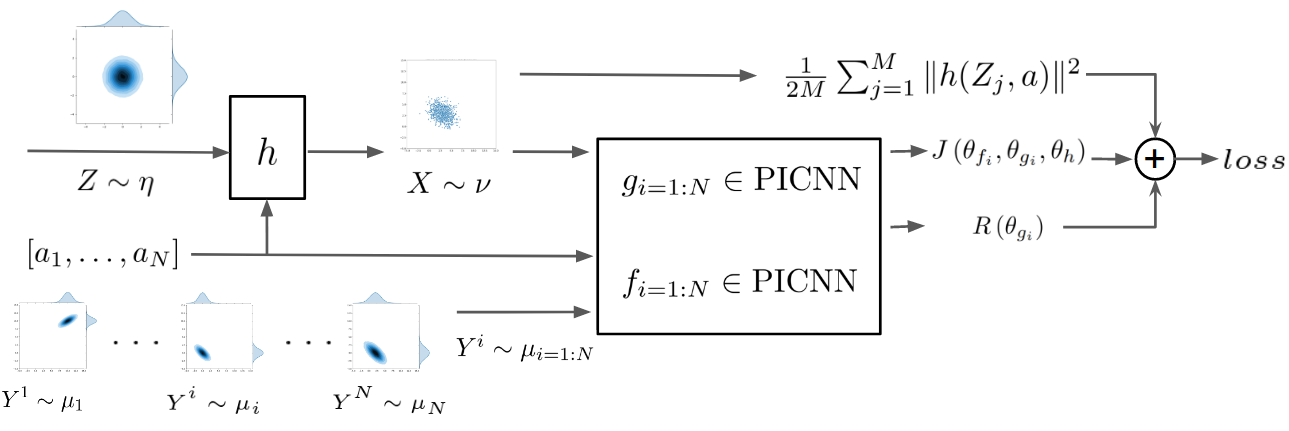}
  \caption{Block diagram for Neural Wasserstein Barycenter-F Algorithm}\label{fig:diagram2}
\end{figure}

The block diagram for Neural Wasserstein Barycenter-F (Algorithm 2) is shown in Figure~\ref{fig:diagram2}.

\subsection{Supportive experiments for NWB-F}
In this part, we evaluate the performance of NWB-F which is an algorithm to calculate the Wasserstein barycenter of a given set of marginals for all weights in one shot. Departing from NWB, the networks $f_i$ and $g_i$ are of PICNN structure. We carry out 3 sets of experiments when the marginal distributions are Gaussian, Gaussian mixtures and sharp distributions. In these experiments, NWB-F converges after 15000 outer cycle iterations.
\begin{wrapfigure}[9]{r}{9cm}
  \centering
  \begin{subfigure}{.21\textwidth}
    \centering
    \includegraphics[width=0.95\linewidth]{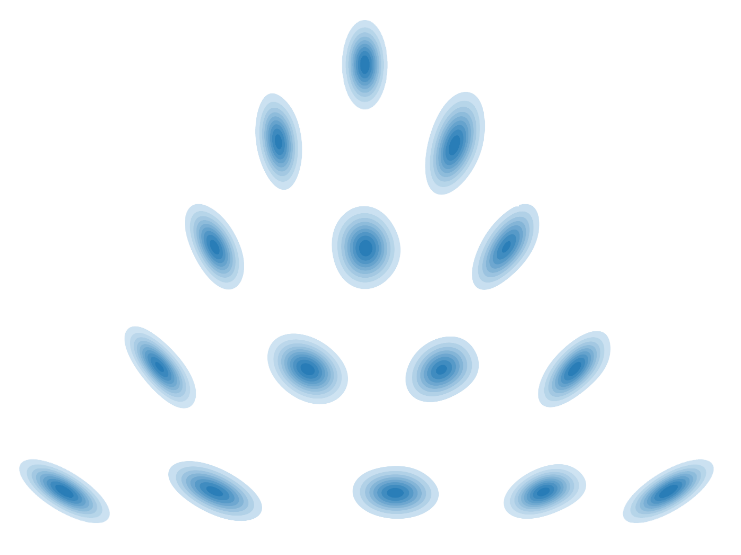}
    \caption{Ours NWB-F}
    \label{fig:line2d_sto}
  \end{subfigure}
  \begin{subfigure}{.21\textwidth}
    \centering
    \includegraphics[width=0.95\linewidth]{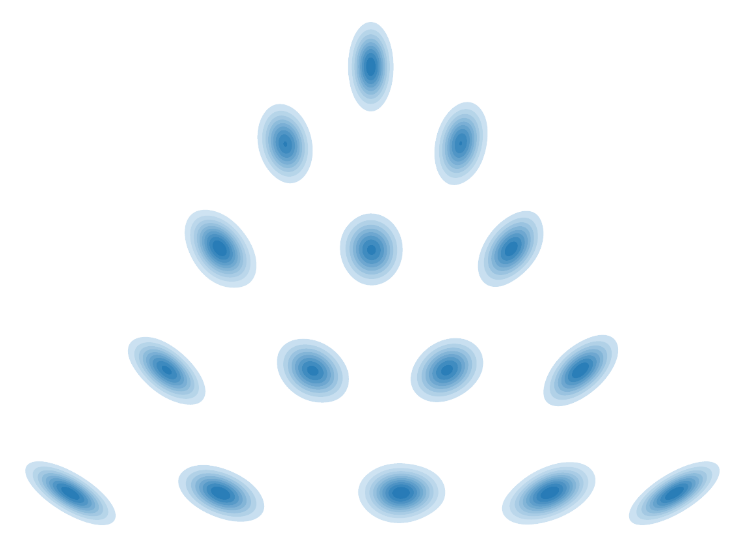}
    \caption{Ground truth}
    \label{fig:free weight b}
  \end{subfigure}
  \caption{Barycenter with different weights using NWB-F.}
  \label{fig:free weight}
\end{wrapfigure}

\paragraph{Gaussian marginal}
We present the experimental result of implementing NWB-F (Algorithm~\ref{alg:BII}) to compute the Wasserstein barycenter for all combinations of weights with a single training. The result for the case of Gaussian marginal distributions, and 12 combination of weight values, is depicted in Figure~\ref{fig:free weight}. For comparison, we have included the exact barycenter. It is qualitatively observed that our approach is able to compute the Wasserstein barycenter for the selected weight combinations in comparison to exact barycenter. The three marginals are
\[
  \mu_1\!\!=\!\! N(\begin{bmatrix}
    0 \\ 0
  \end{bmatrix}\!,\!
  \begin{bmatrix}
    0.5 & 0 \\
    0   & 2
  \end{bmatrix}),~~
  \mu_2\!\! =\!\! N(\begin{bmatrix}
    0 \\ 0
  \end{bmatrix}\!,\!
  \begin{bmatrix}
    2 & 1 \\
    1 & 1
  \end{bmatrix}),~~
  \mu_3 = N(\begin{bmatrix}
    0 \\ 0
  \end{bmatrix}\!,\!
  \begin{bmatrix}
    2  & -1 \\
    -1 & 1
  \end{bmatrix}).
\]
To quantitatively verify the performance of NWB-F, we compare the barycenters to ground truth with several different weight in terms of KL-divergence. The resulting error is respectively $0.0235$ for $a=[0.5,0.25,0.25]$, $0.0153$ for $a=[0.25,0.5,0.25]$, and $0.0114$ for $a=[0.25,0.25,0.5]$. The error of results using NWB-F is consistently small among different weight combinations.

The networks $f_i$ and $g_i$ each has 3 layers and the generative network $h$ has 4 layers.
All networks have 12 neurons for each hidden layer. Learning rate is 0.001. The inner loop iteration numbers are $K_1=6$ and $K_2=4$. The batch size is $M=100$.

\begin{figure}
  \centering
  \begin{subfigure}{.175\textwidth}
    \centering
    \includegraphics[width=0.95\linewidth]{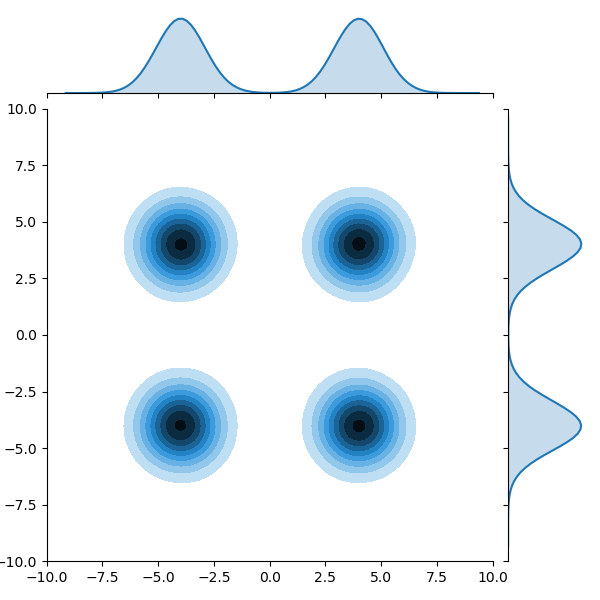}
    \caption*{Marginal 1}
  \end{subfigure}
  \begin{subfigure}{.35\textwidth}
    \centering
    \includegraphics[width=0.95\linewidth]{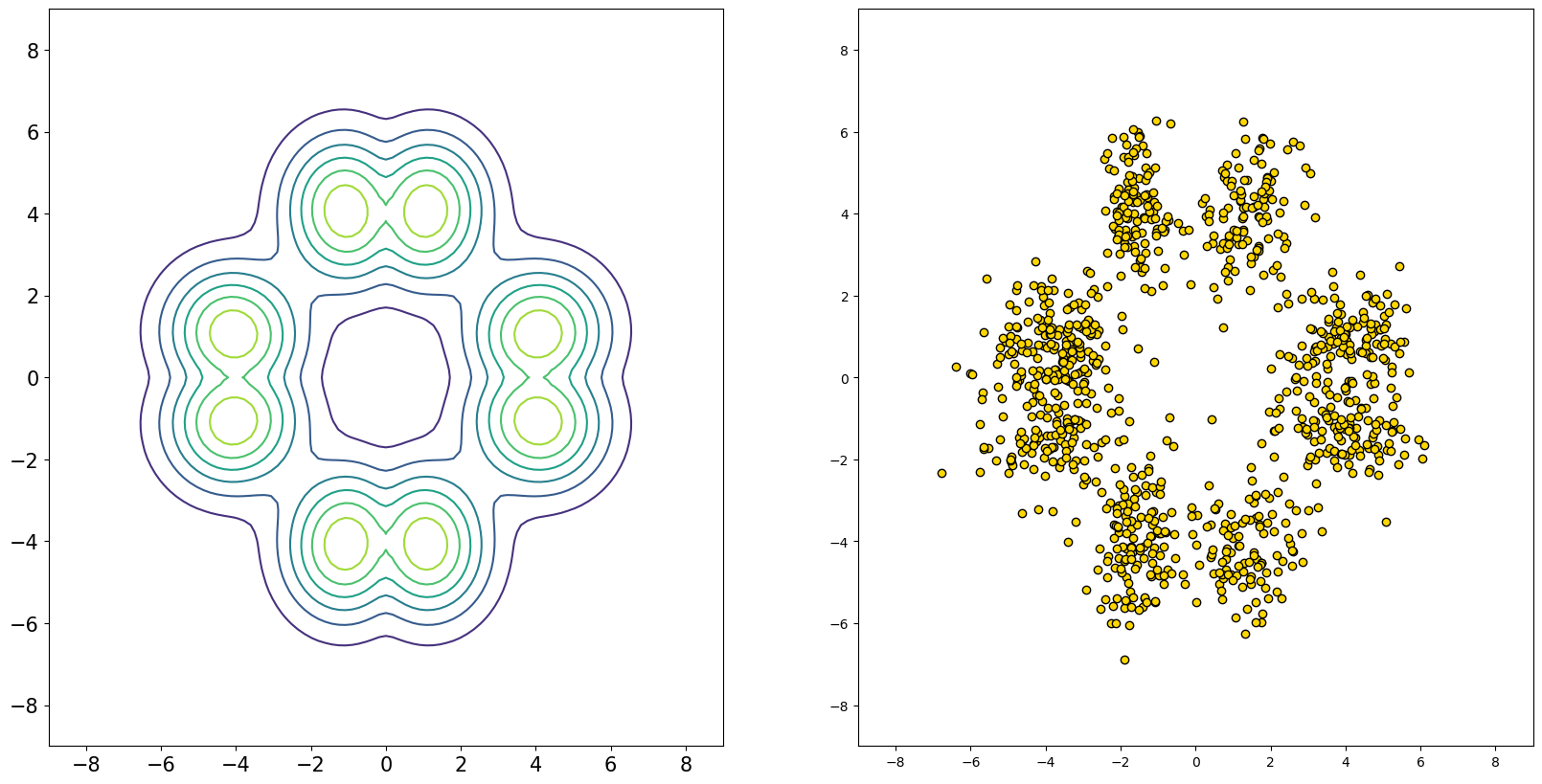}
    \caption*{weight [0.2,0.8]}
  \end{subfigure}
  \begin{subfigure}{.35\textwidth}
    \centering
    \includegraphics[width=0.95\linewidth]{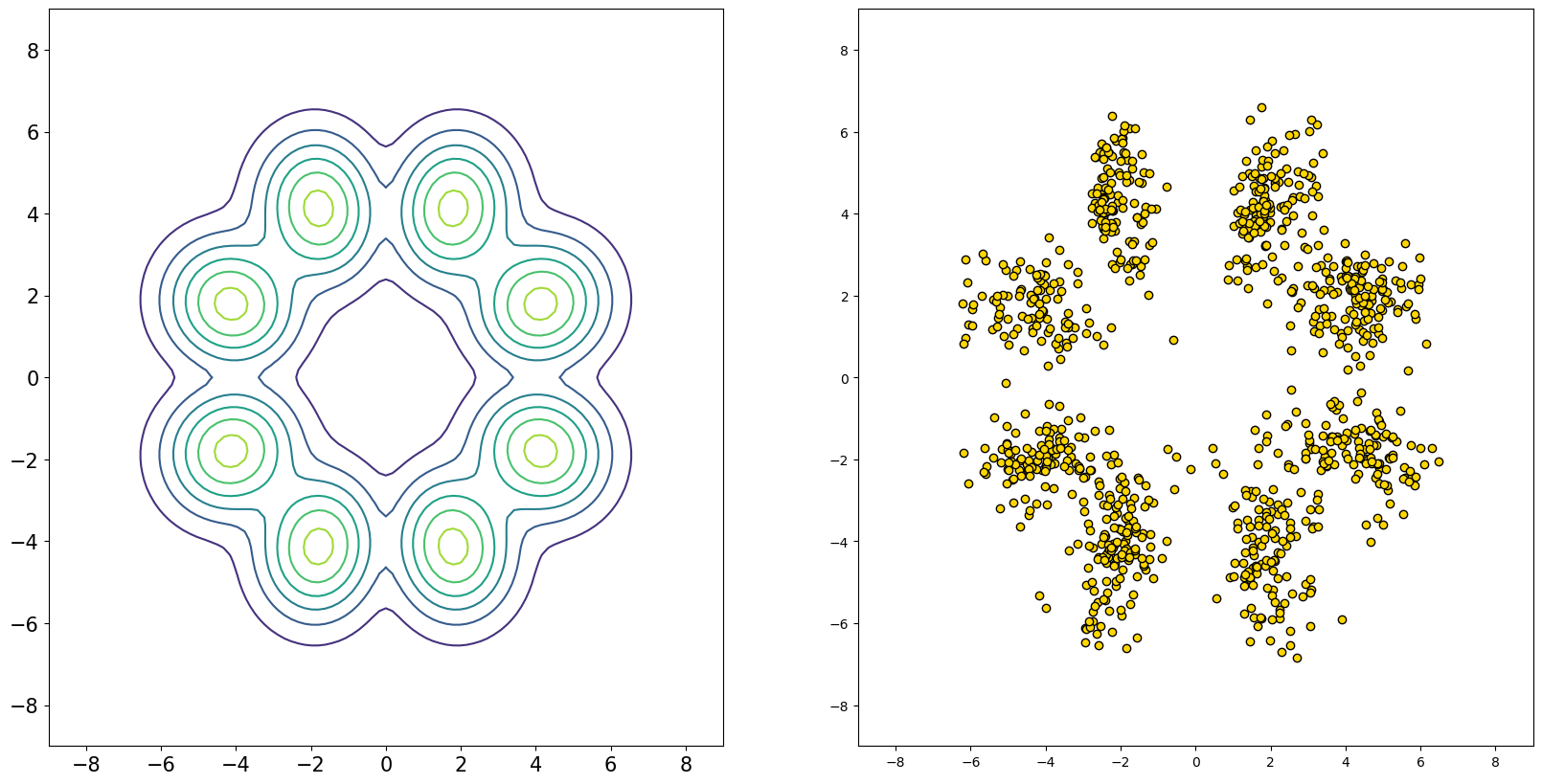}
    \caption*{weight [0.4,0.6]}
  \end{subfigure}

  \begin{subfigure}{.175\textwidth}
    \centering
    \includegraphics[width=0.95\linewidth]{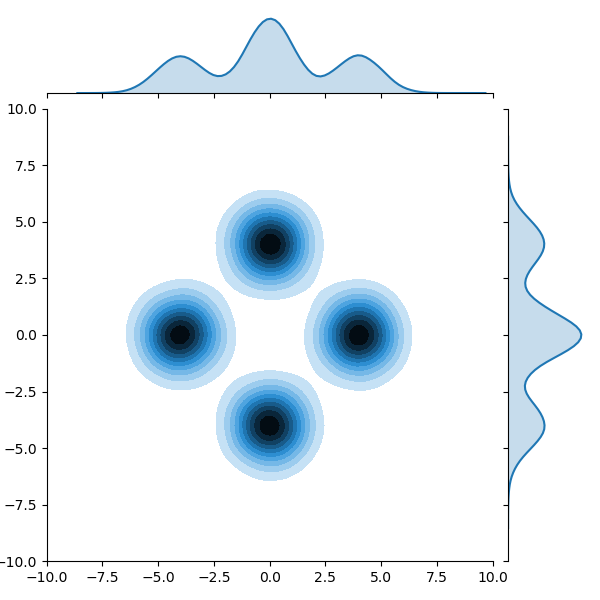}
    \caption*{Marginal 2}
  \end{subfigure}
  \begin{subfigure}{.35\textwidth}
    \centering
    \includegraphics[width=0.95\linewidth]{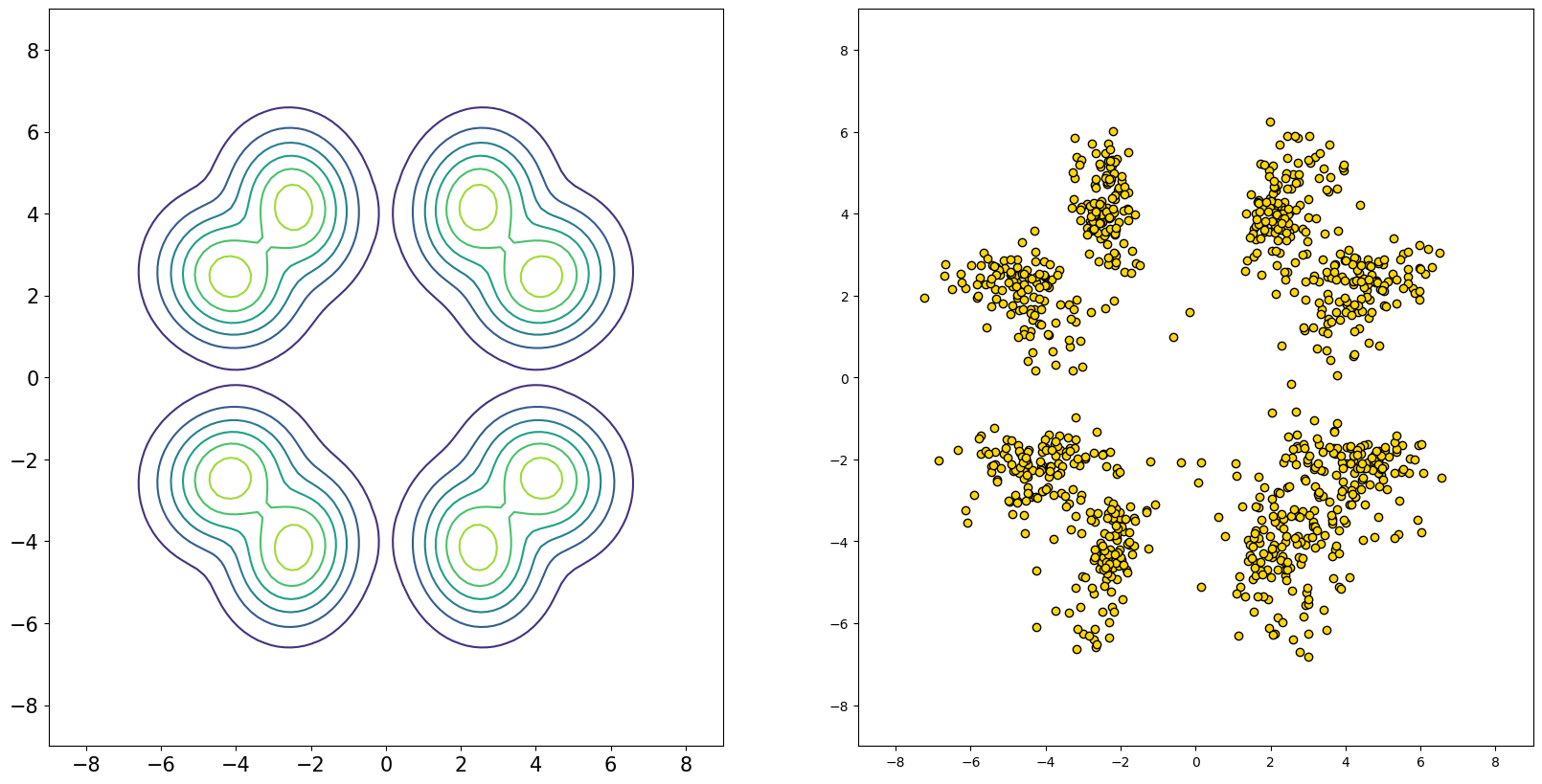}
    \caption*{weight [0.6,0.4]}
  \end{subfigure}
  \begin{subfigure}{.35\textwidth}
    \centering
    \includegraphics[width=0.95\linewidth]{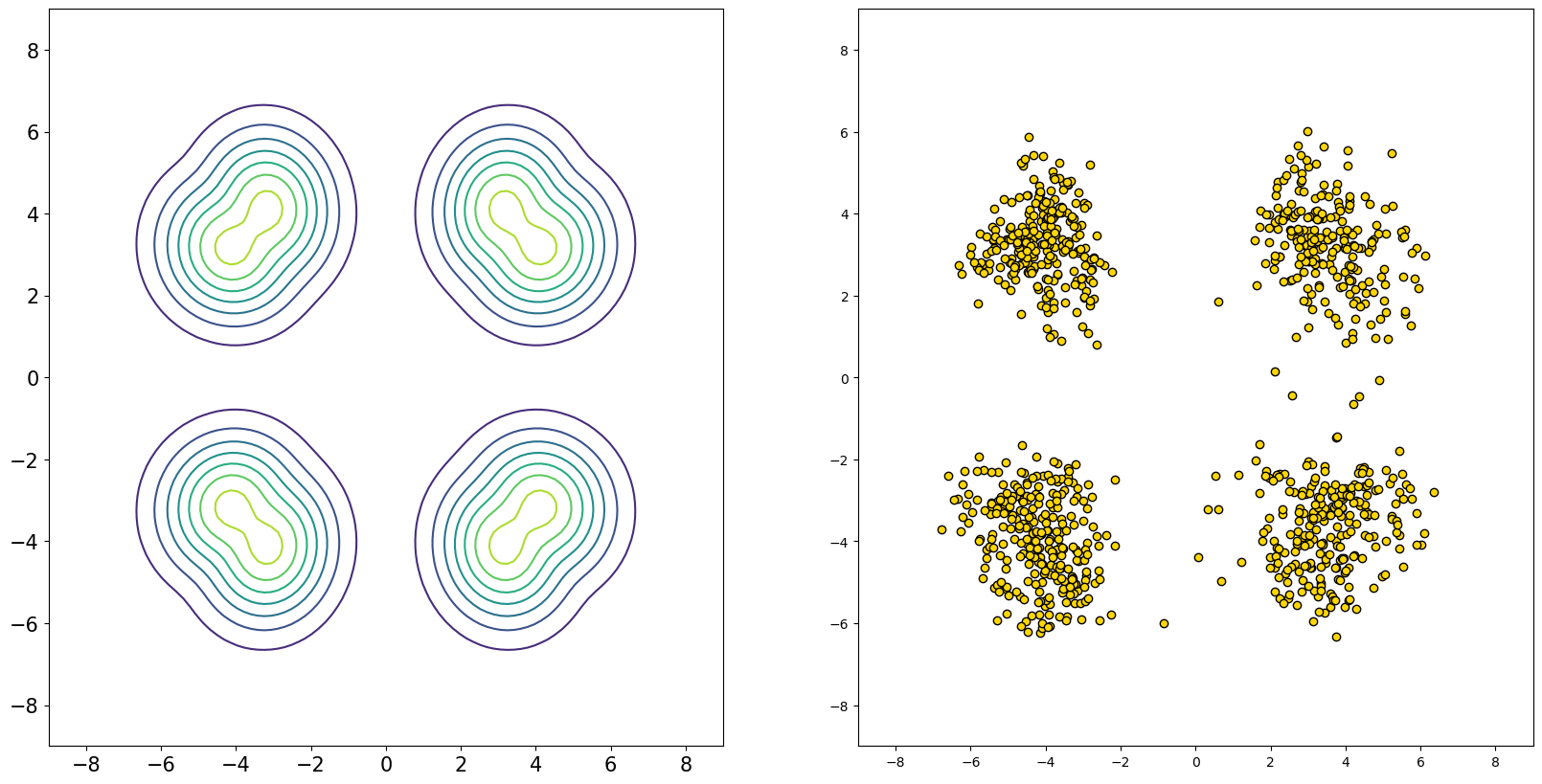}
    \caption*{weight [0.8,0.2]}
  \end{subfigure}
  \caption{Barycenter with different weights using NWB-F. For each subfigure, the plot on the left is obtained using \citet{SolDeGui15} and the plot on the right is obtained using NWB-F.}
  \label{fig:free weight mixture}
\end{figure}

\paragraph{Gaussian mixture marginal}
We apply NWB-F to obtain the Wasserstein barycenter for all weights in one shot. The first marginal is a uniform combination of the Gaussian distributions
\[
  N(\begin{bmatrix}
    4 \\ 4
  \end{bmatrix}\!,\!
  \begin{bmatrix}
    1 & 0 \\
    0 & 1
  \end{bmatrix}),~~
  N(\begin{bmatrix}
    4 \\ -4
  \end{bmatrix}\!,\!
  \begin{bmatrix}
    1 & 0 \\
    0 & 1
  \end{bmatrix}),~~
  N(\begin{bmatrix}
    -4 \\ -4
  \end{bmatrix}\!,\!
  \begin{bmatrix}
    1 & 0 \\
    0 & 1
  \end{bmatrix}),~~
  N(\begin{bmatrix}
    -4 \\ 4
  \end{bmatrix}\!,\!
  \begin{bmatrix}
    1 & 0 \\
    0 & 1
  \end{bmatrix}).
\]
The second marginal is a uniform combination of the Gaussian distributions
\[
  N(\begin{bmatrix}
    0 \\ 4
  \end{bmatrix}\!,\!
  \begin{bmatrix}
    1 & 0 \\
    0 & 1
  \end{bmatrix}),~~
  N(\begin{bmatrix}
    4 \\ 0
  \end{bmatrix}\!,\!
  \begin{bmatrix}
    1 & 0 \\
    0 & 1
  \end{bmatrix}),~~
  N(\begin{bmatrix}
    0 \\ -4
  \end{bmatrix}\!,\!
  \begin{bmatrix}
    1 & 0 \\
    0 & 1
  \end{bmatrix}),~~
  N(\begin{bmatrix}
    -4 \\ 0
  \end{bmatrix}\!,\!
  \begin{bmatrix}
    1 & 0 \\
    0 & 1
  \end{bmatrix}).
\]
The experiment results are depicted in Figure \ref{fig:free weight mixture} in comparison with Convolutional Wasserstein Barycenter \cite{SolDeGui15}. We remark that this is not a fair comparison since NWB-F obtained all the barycenters with different weights in one shot while \citet{SolDeGui15} has to be run separately for each weight. Nevertheless, NWB-F generates reasonable results.

The networks $f_i$ and $g_i$ each has 5 layers and the generative network $h$ has 6 layers.
All networks have 12 neurons for each hidden layer. Batch normalization is used in $h$.
Learning rate is 0.001. The inner loop iteration numbers are $K_1=10$ and $K_2=6$. The batch size is $M=100$.
\paragraph{Sharp line marginal}

\begin{figure}[h]
  \begin{subfigure}{1\textwidth}
    \centering
    \includegraphics[width=0.95\linewidth]{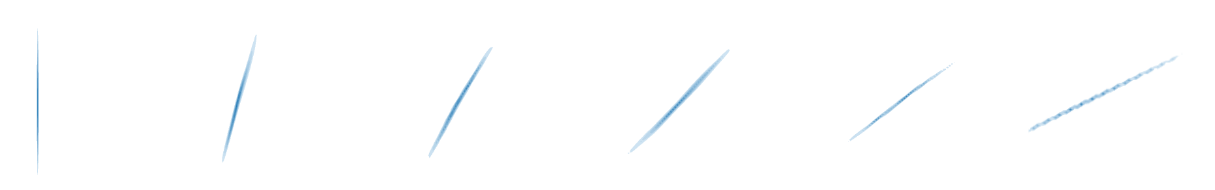}
    \caption{Ours NWB-F}
    \label{fig:free weight b}
  \end{subfigure}

  \begin{subfigure}{1\textwidth}
    \centering
    \includegraphics[width=0.95\linewidth]{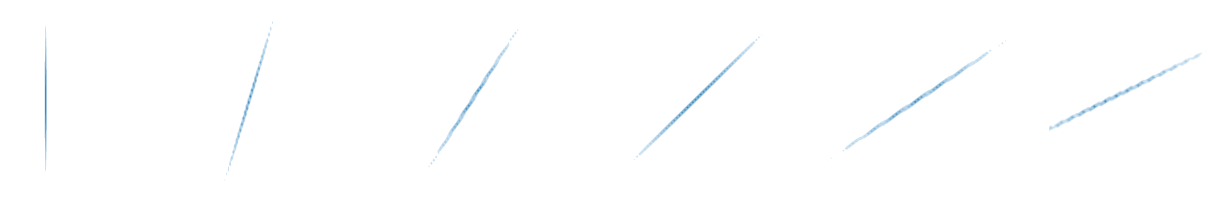}
    \caption{Ground truth}
    \label{fig:line2d_sto}
  \end{subfigure}

  \caption{Barycenter with different weights using NWB-F. For each subfigure, the two plots at both ends are given marginal distributions supported on two line segments.}

  \label{fig:free weight line}
\end{figure}

Given two marginals supported on two lines, we apply NWB-F to obtain the Wasserstein barycenter for all weights in one shot. Note that these Wasserstein barycenters in fact constitute the Wasserstein geodesic between the two distributions.
The networks $f_i$ and $g_i$ each has 4 layers and the generative network $h$ has 4 layers.
All networks have 12 neurons for each hidden layer. Batch normalization is used in $h$.
Learning rate is 0.001. The inner loop iteration numbers are $K_1=6$ and $K_2=4$. The batch size is $M=100$. The experiment results are depicted in Figure \ref{fig:free weight line} in comparison with ground truth results.

\section{Experiment details for NWB and more supportive experiments}
In this section, we provide the experiment details as well as more supportive experimental results of NWB. Some common experiment setup for NWB is: \\
1) All $f_i$ and $g_i$ networks use CELU activation function while the $h$ network uses \textbf{PReLU} \cite{he2015delving} activation function.\\
2) The weight $\lambda=0.1$ for the regularizer $R\left(\theta_{g_i}\right)=\lambda \sum_{W_{l} \in \theta_{g_i}}\left\|\max \left(-W_{l}, 0\right)\right\|_{F}^{2}$.\\
3) All optimizers are Adam.\\
4) All $h$ used in this article are vanilla feedforward networks. \\
5) All input Gaussian distribution $\eta$ has zero mean an identity covariance. \\
6) The inner loop iteration numbers are $K_1=6$ and $K_2=4$.\\
7) The batch size is $M=100$ unless further specified. \\
8) NWB converges after \textbf{15000} outer cycle iterations unless further specified.

We also note that the value of the evalutation metric $\text{BW}_{2}^{2} \textendash \text{UVP}$ is sensitive to the number of samples. To be consistent, we draw 10000 samples from each method to calculate $\text{BW}_{2}^{2} \textendash \text{UVP}$.
\subsection{Learning the Gaussian mixture Wasserstein Barycenter}
\begin{figure}[h]
  \centering
  \begin{subfigure}{.19\textwidth}
    \centering
    \includegraphics[width=1\linewidth]{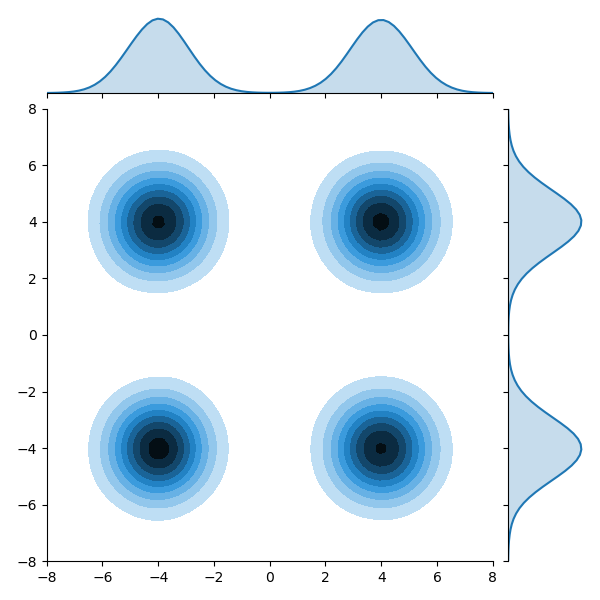}
    \caption{Marginal 1}
  \end{subfigure}
  \begin{subfigure}{.19\textwidth}
    \centering
    \includegraphics[width=1\linewidth]{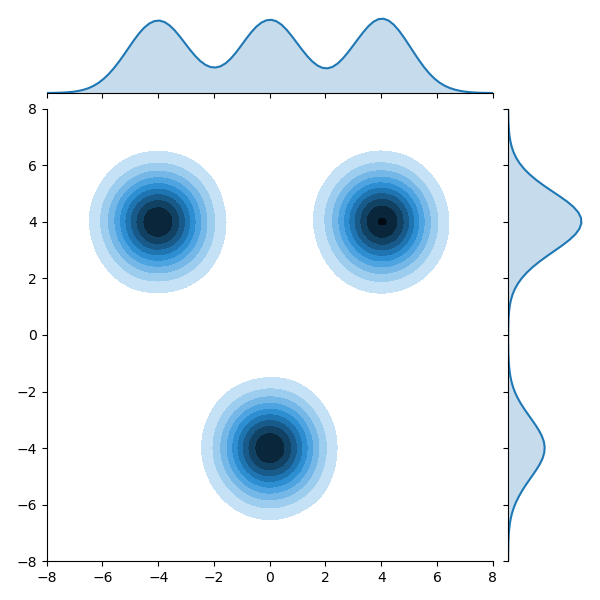}
    \caption{Marginal 2}
  \end{subfigure}
  \begin{subfigure}{.19\textwidth}
    \centering
    \includegraphics[width=1\linewidth]{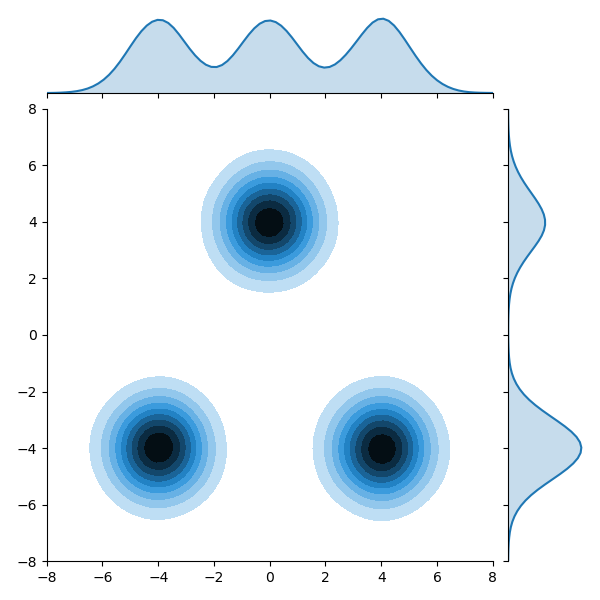}
    \caption{Marginal 2}
  \end{subfigure}
  \begin{subfigure}{.19\textwidth}
    \centering
    \includegraphics[width=1\linewidth]{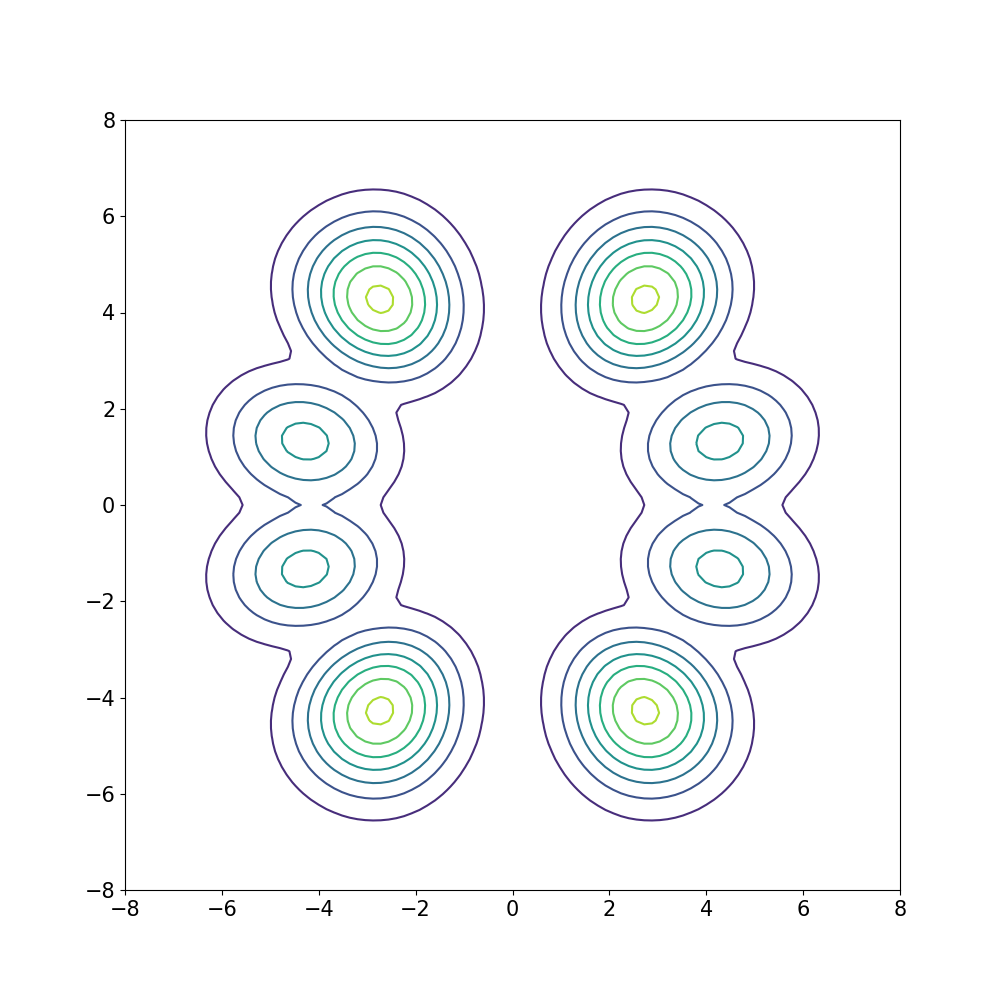}
    \caption{\citet{SolDeGui15}}
  \end{subfigure}
  \begin{subfigure}{.19\textwidth}
    \centering
    \includegraphics[width=1\linewidth]{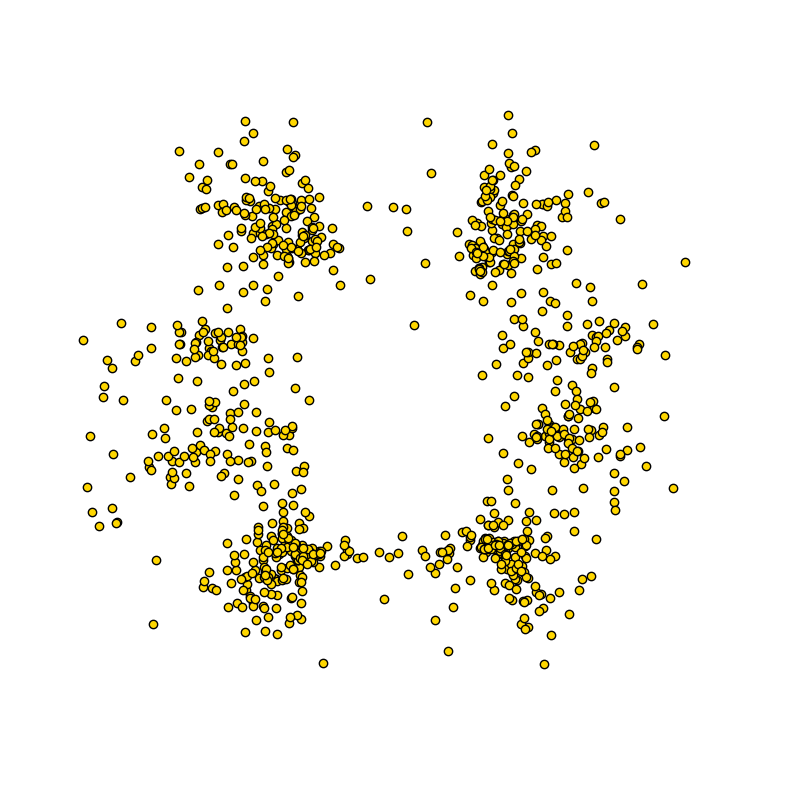}
    \caption{ NWB $h\sharp \eta$}
  \end{subfigure}

  \begin{subfigure}{.19\textwidth}
    \centering
    \includegraphics[width=1\linewidth]{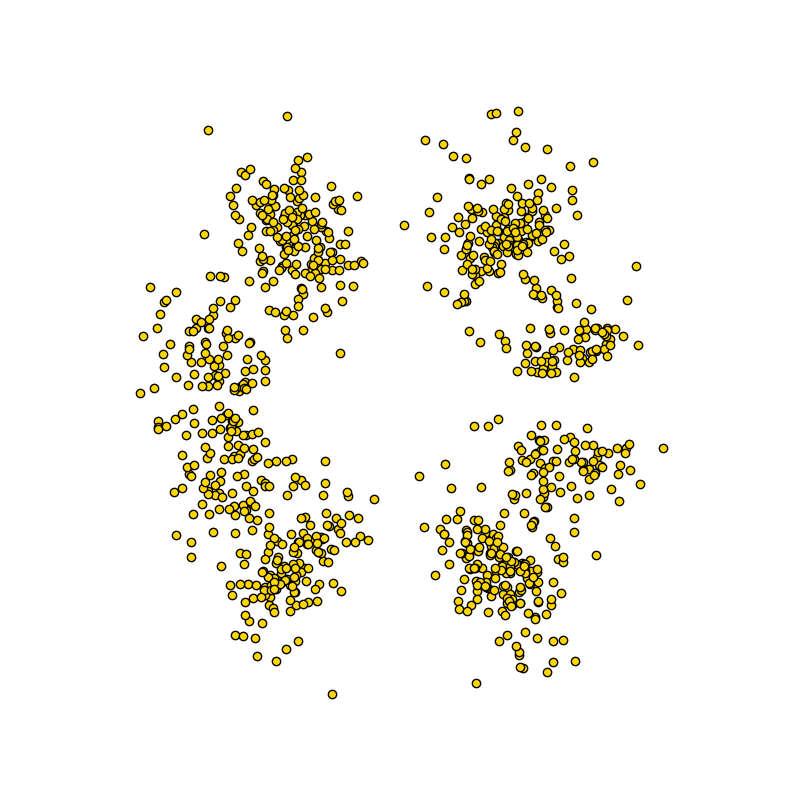}
    \caption{{NWB $\nabla g_1 \sharp \mu_1$}}
  \end{subfigure}
  \begin{subfigure}{.19\textwidth}
    \centering
    \includegraphics[width=1\linewidth]{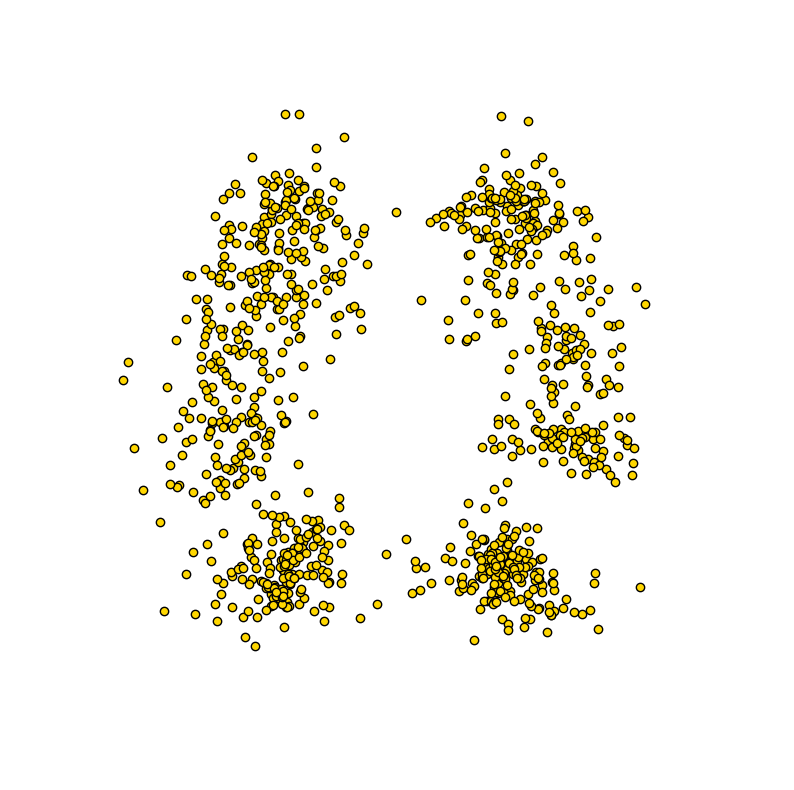}
    \caption{{NWB $\nabla g_2 \sharp \mu_2$}}
  \end{subfigure}
  \begin{subfigure}{.19\textwidth}
    \centering
    \includegraphics[width=1\linewidth]{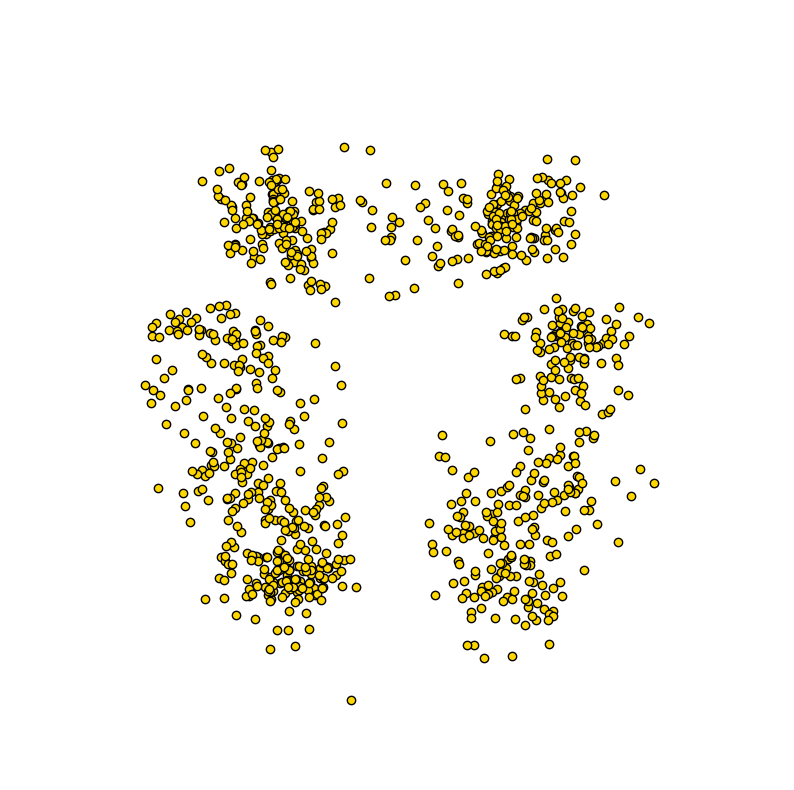}
    \caption{{NWB $\nabla g_3 \sharp \mu_3$}}
  \end{subfigure}
  \begin{subfigure}{.19\textwidth}
    \centering
    \includegraphics[width=1\linewidth]{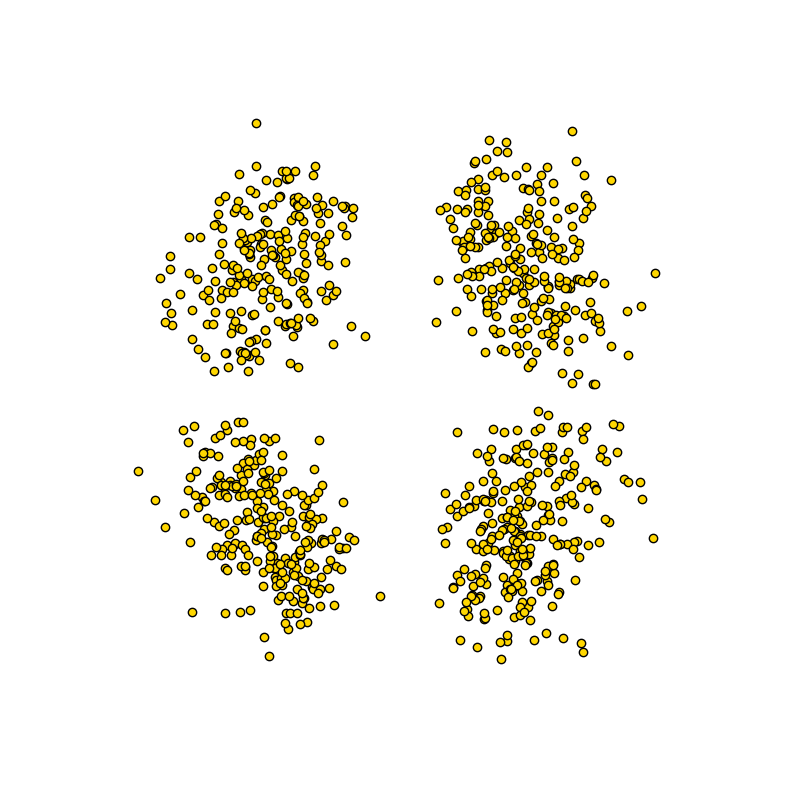}
    \caption{{CWB $\nabla g_1 \sharp \mu_1$}}
  \end{subfigure}
  \begin{subfigure}{.19\textwidth}
    \centering
    \includegraphics[width=1\linewidth]{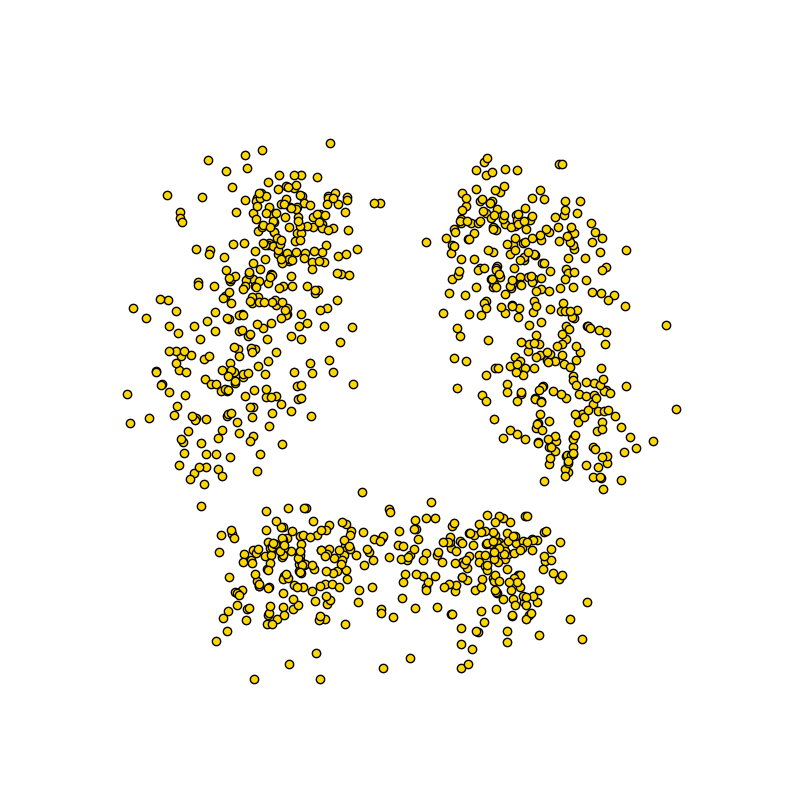}
    \caption{{CWB $\nabla g_2 \sharp \mu_2$}}
  \end{subfigure}

  \begin{subfigure}{.19\textwidth}
    \centering
    \includegraphics[width=1\linewidth]{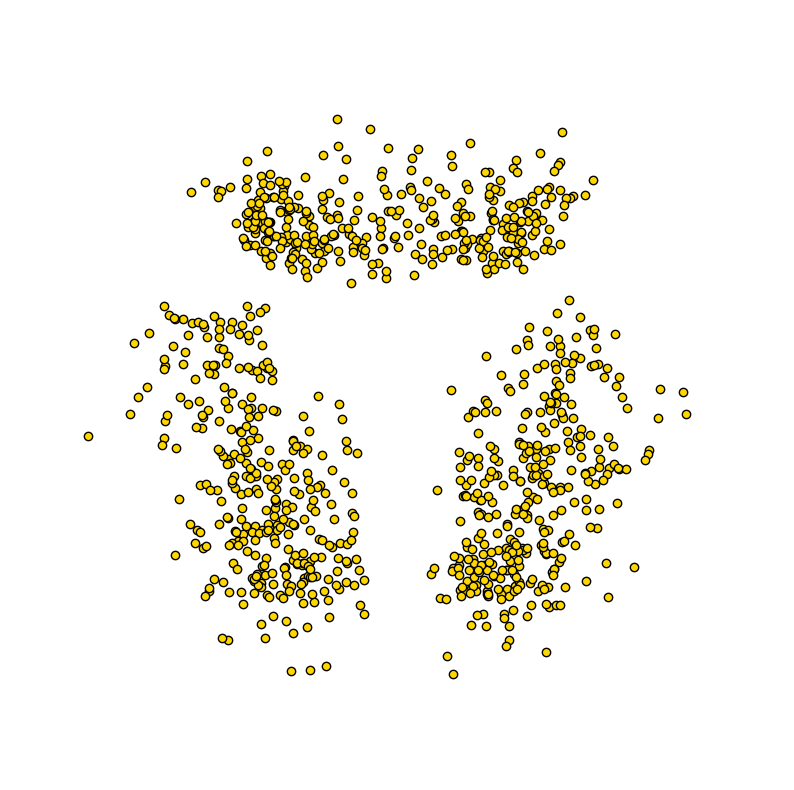}
    \caption{{CWB $\nabla g_3 \sharp \mu_3$}}
  \end{subfigure}
  \begin{subfigure}{.19\textwidth}
    \centering
    \includegraphics[width=1\linewidth]{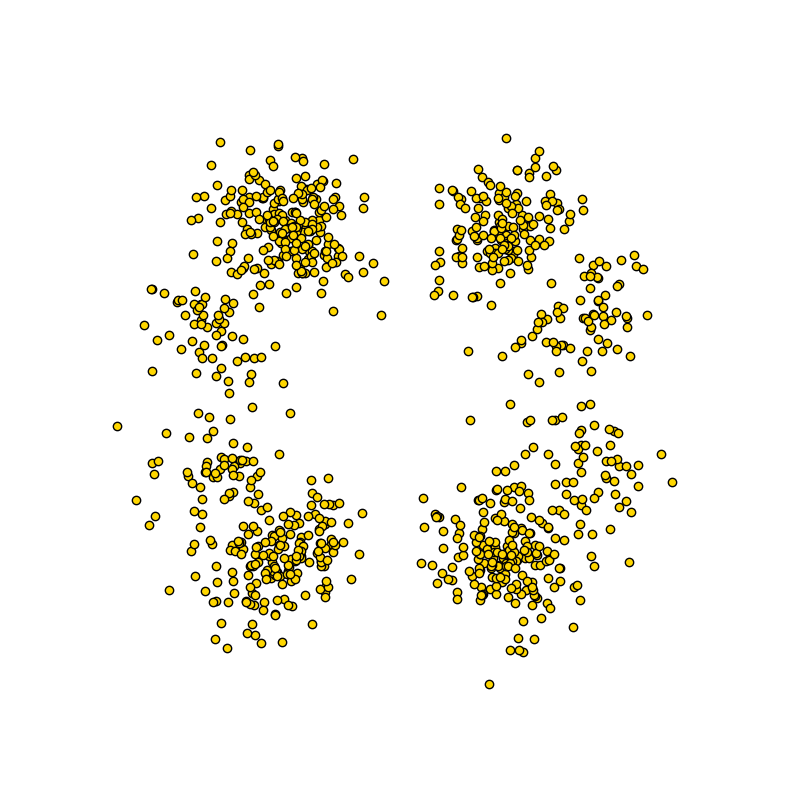}
    \caption{{CRWB $\nabla g_1 \sharp \mu_1$}}
  \end{subfigure}
  \begin{subfigure}{.19\textwidth}
    \centering
    \includegraphics[width=1\linewidth]{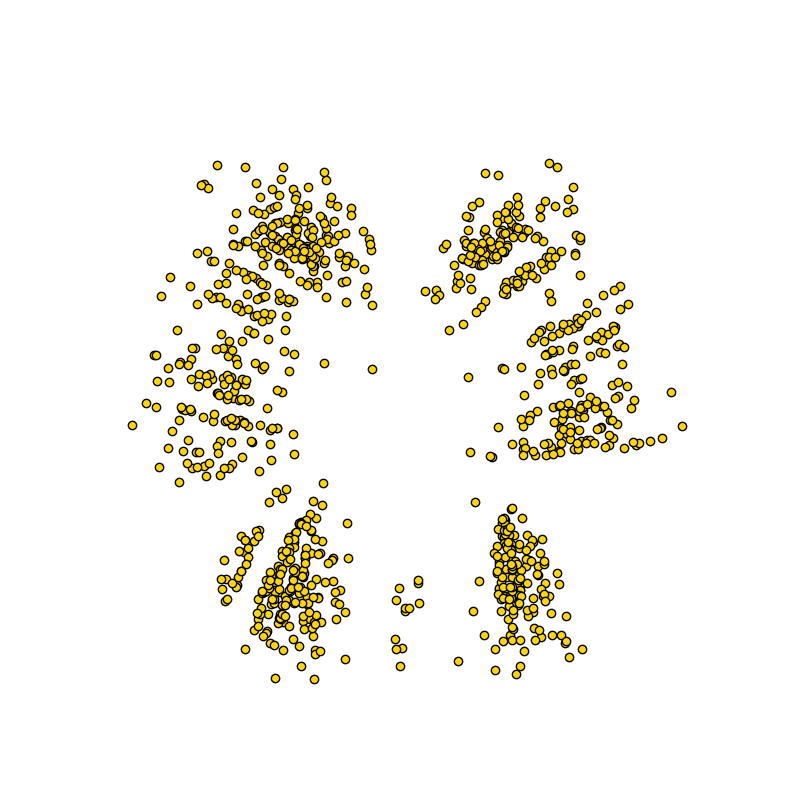}
    \caption{{CRWB $\nabla g_2 \sharp \mu_2$}}
  \end{subfigure}
  \begin{subfigure}{.19\textwidth}
    \centering
    \includegraphics[width=1\linewidth]{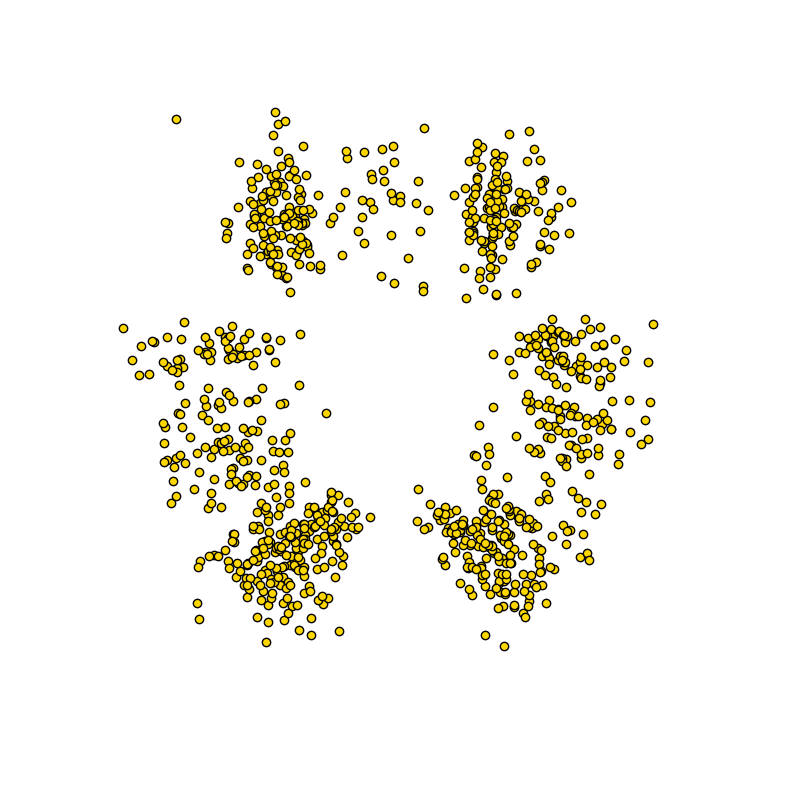}
    \caption{{CRWB $\nabla g_3 \sharp \mu_3$}}
  \end{subfigure}
  
    \begin{subfigure}{0.19\textwidth}
        \centering
        \includegraphics[width=1\linewidth]{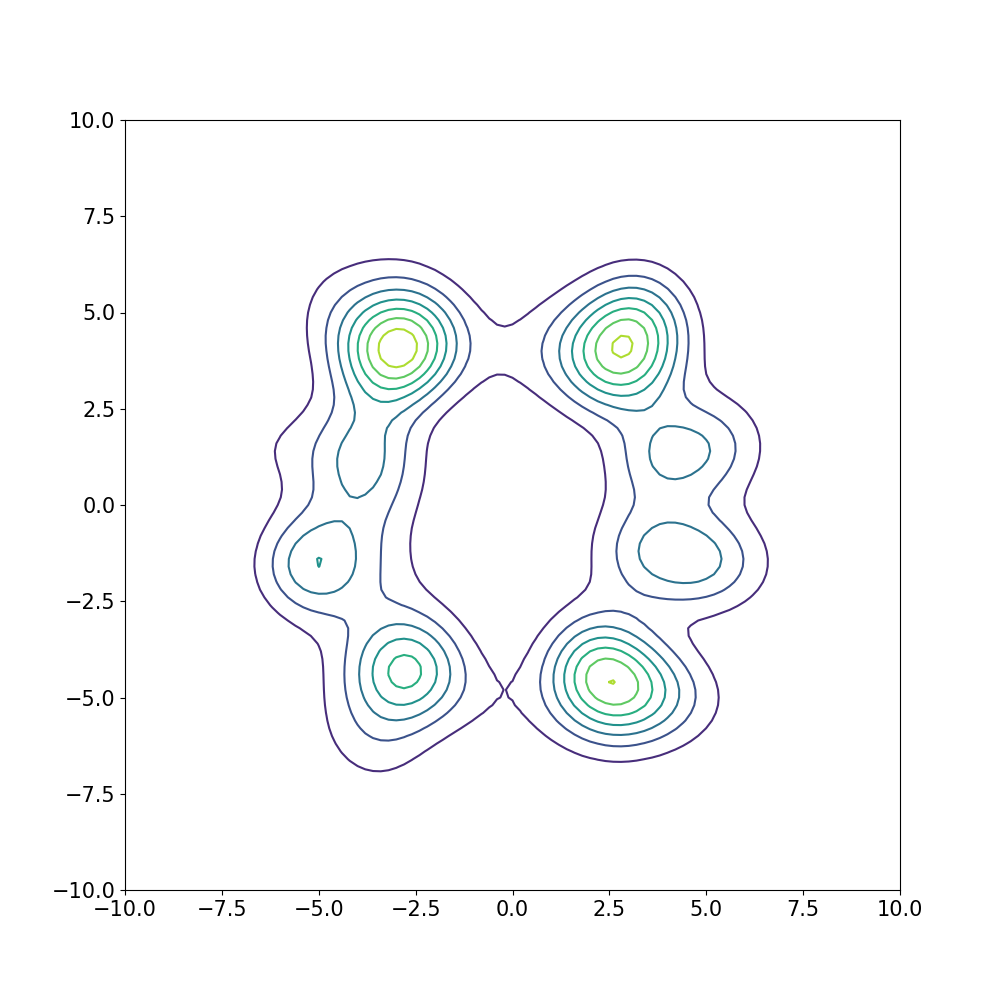}
        \caption{NWB $h \sharp \eta$}
    \end{subfigure}
    \begin{subfigure}{0.19\textwidth}
        \centering
        \includegraphics[width=1\linewidth]{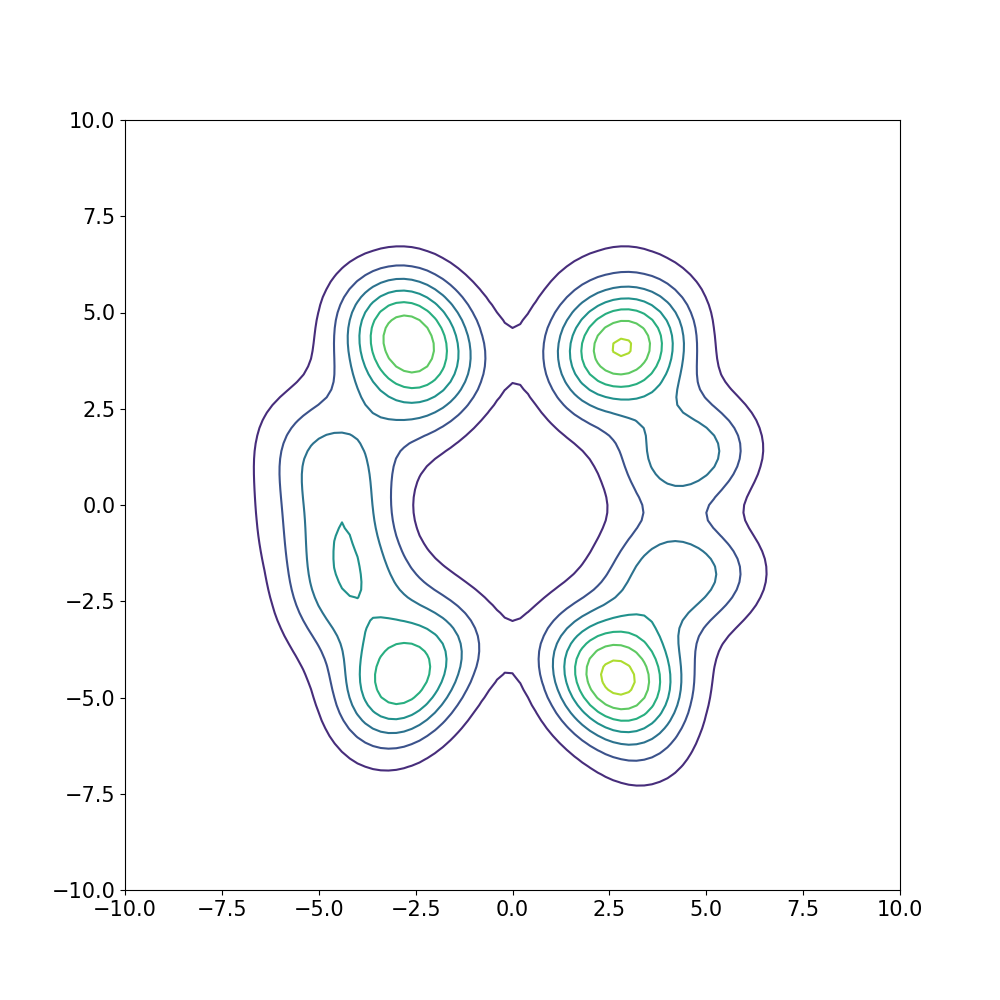}
        \caption{NWB $\nabla g_1 \sharp \mu_1$}
    \end{subfigure}
\begin{subfigure}{0.19\textwidth}
        \centering
        \includegraphics[width=1\linewidth]{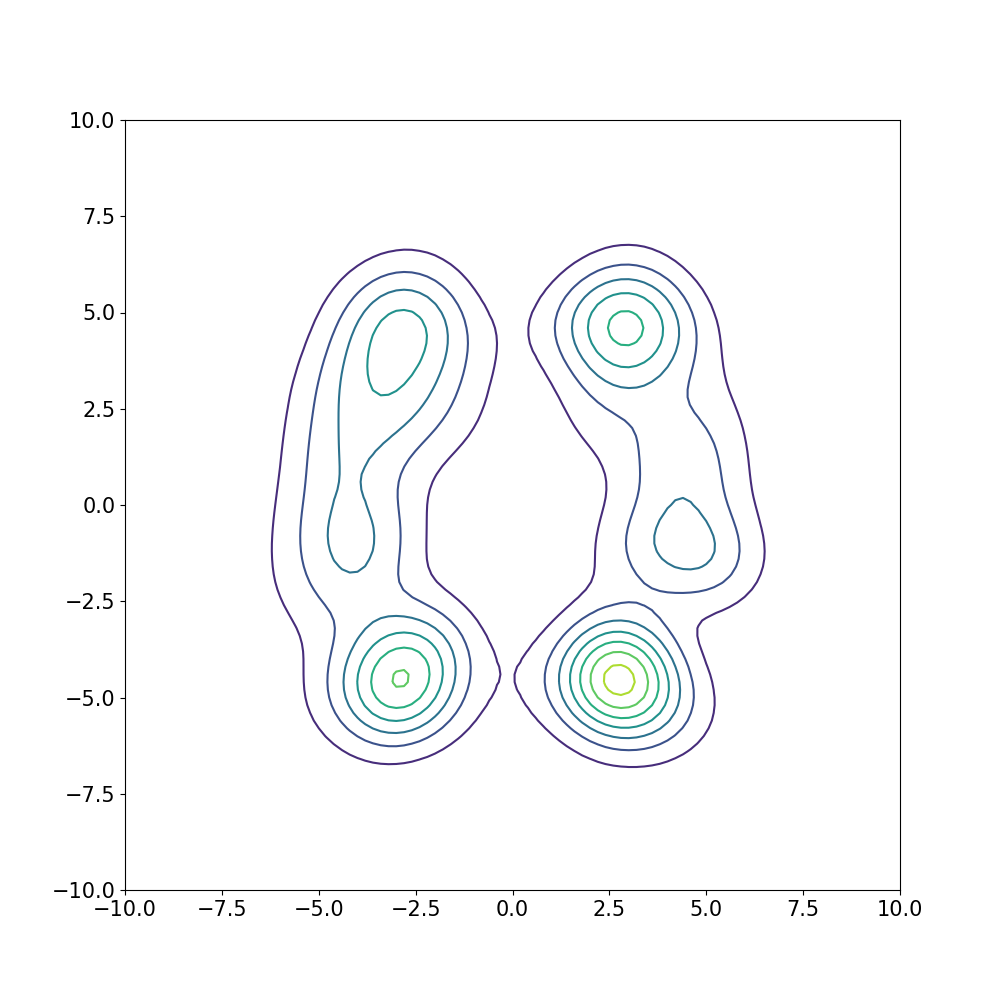}
        \caption{NWB $\nabla g_2 \sharp \mu_2$}
    \end{subfigure}
    \begin{subfigure}{0.19\textwidth}
        \centering
        \includegraphics[width=1\linewidth]{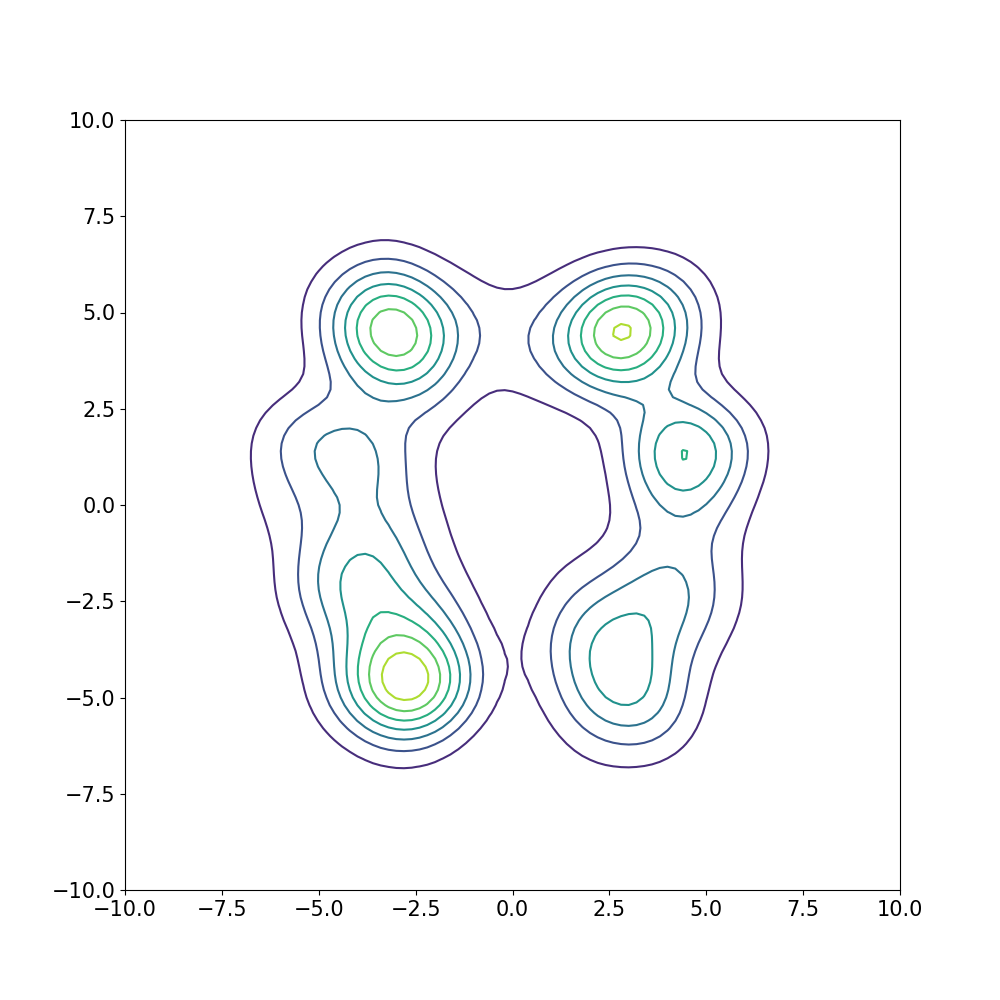}
        \caption{NWB $\nabla g_3 \sharp \mu_3$}
    \end{subfigure}  
    \begin{subfigure}{0.19\textwidth}
        \centering
        \includegraphics[width=1\linewidth]{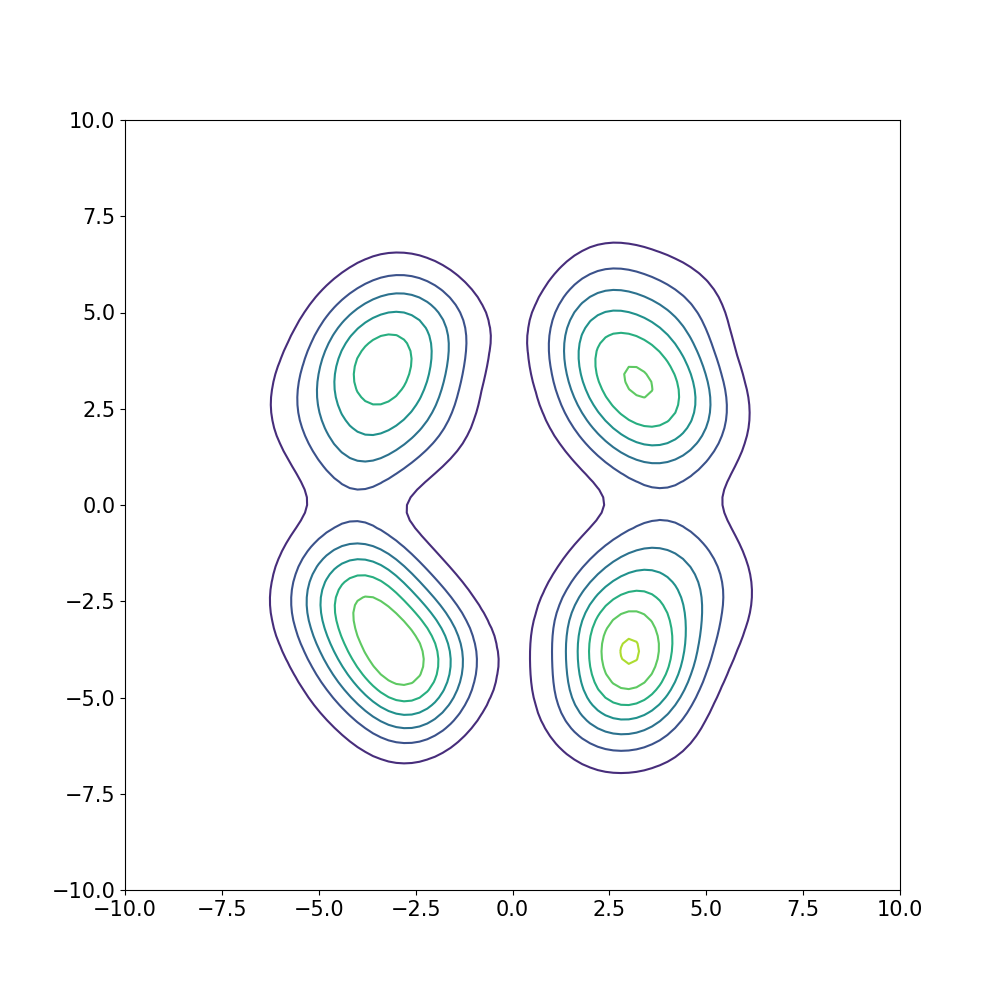}
        \caption{CWB $\nabla g_1 \sharp \mu_1$}
    \end{subfigure}
    
    \begin{subfigure}{0.19\textwidth}
        \centering
        \includegraphics[width=1\linewidth]{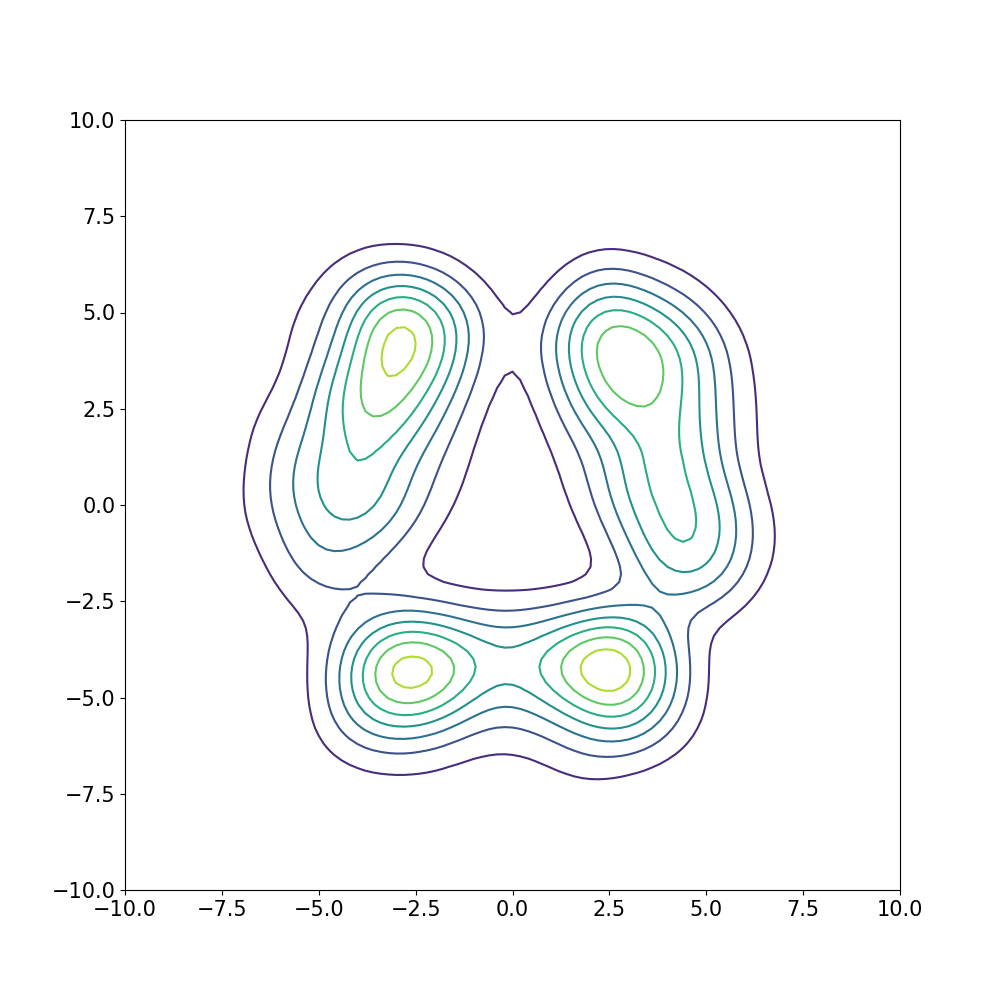}
        \caption{CWB $\nabla g_2 \sharp \mu_2$}
    \end{subfigure}
    \begin{subfigure}{0.19\textwidth}
        \centering
        \includegraphics[width=1\linewidth]{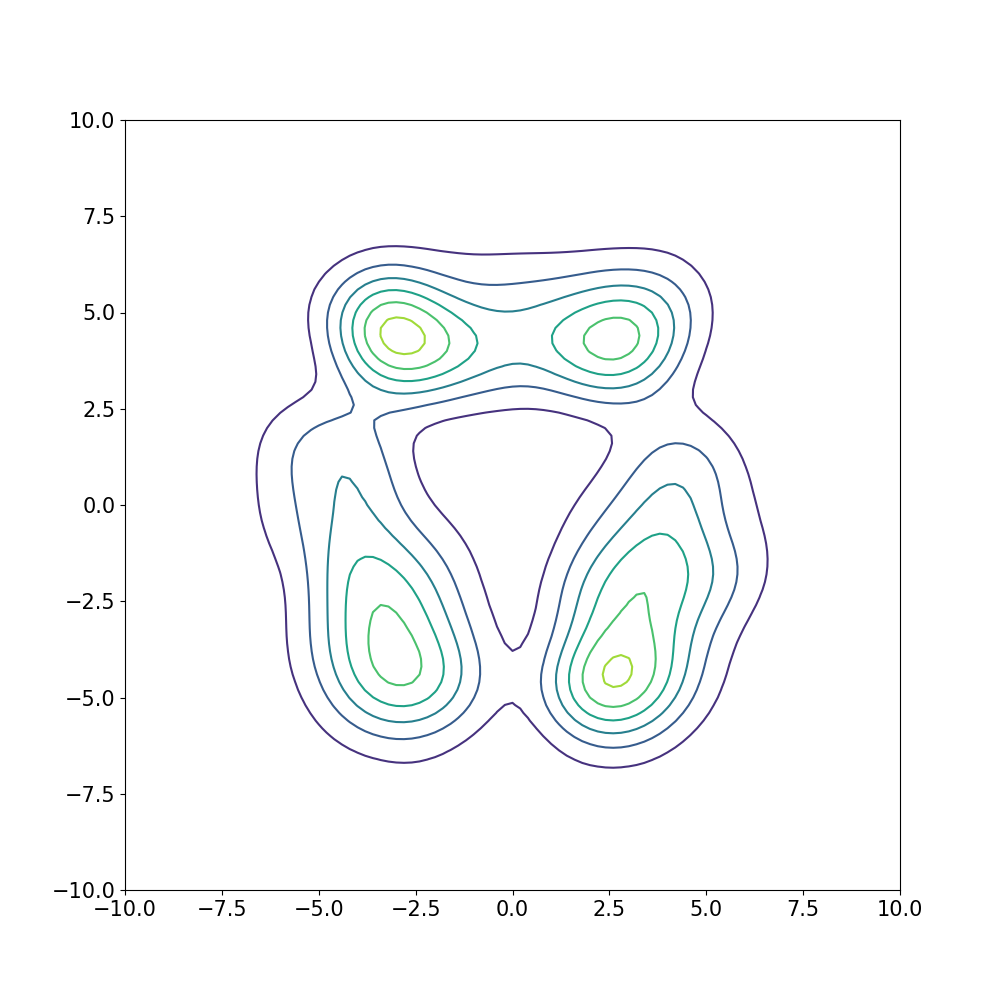}
        \caption{CWB $\nabla g_3 \sharp \mu_3$}
    \end{subfigure}
    \begin{subfigure}{0.19\textwidth}
        \centering
        \includegraphics[width=1\linewidth]{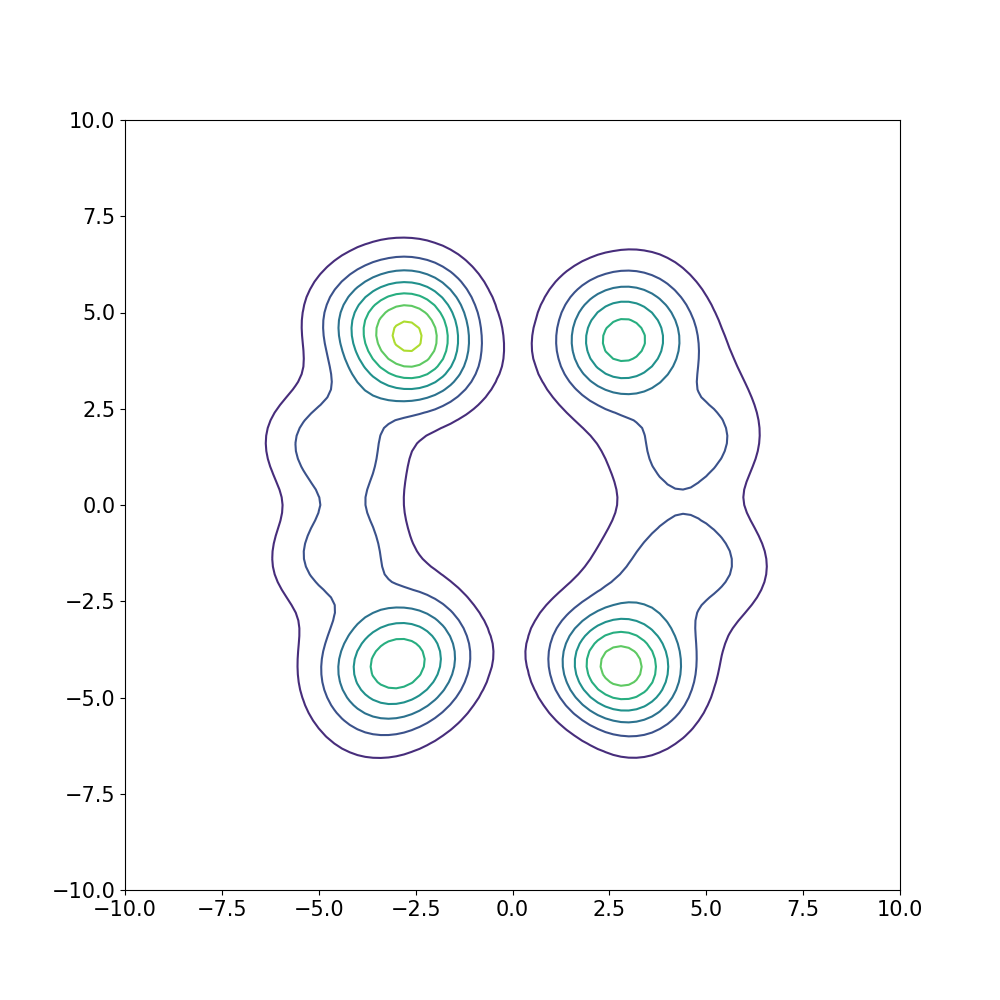}
        \caption{CRWB $\nabla g_1 \sharp \mu_1$}
    \end{subfigure}
    \begin{subfigure}{0.19\textwidth}
        \centering
        \includegraphics[width=1\linewidth]{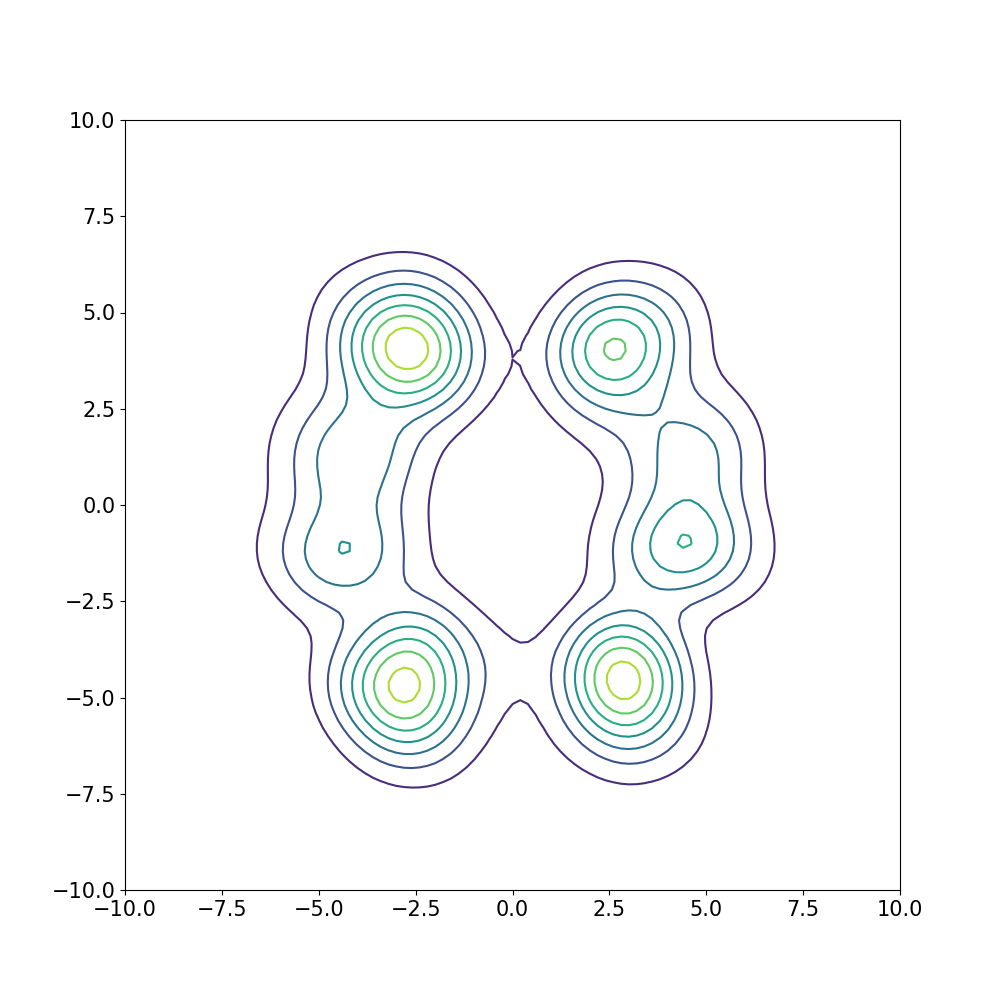}
        \caption{CRWB $\nabla g_2 \sharp \mu_2$}
    \end{subfigure}
    \begin{subfigure}{0.19\textwidth}
        \centering
        \includegraphics[width=1\linewidth]{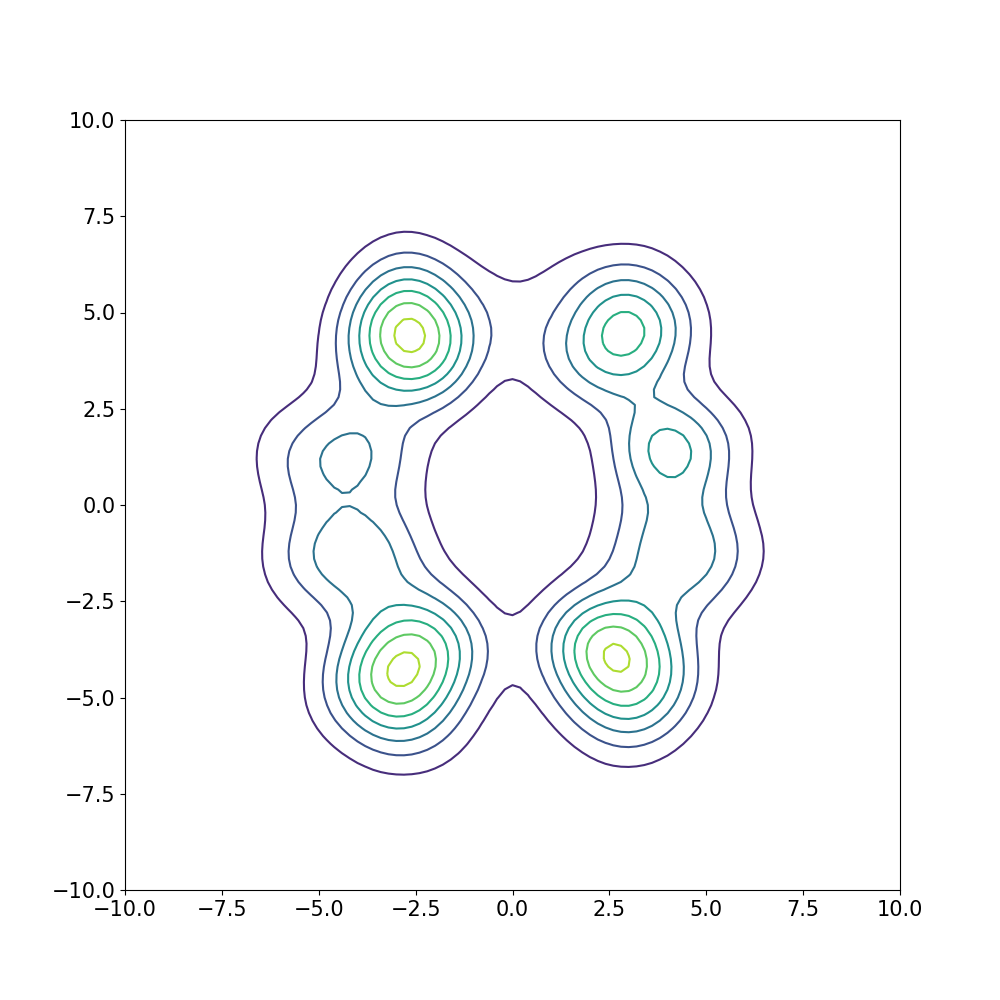}
        \caption{CRWB $\nabla g_3 \sharp \mu_3$}
    \end{subfigure}    
   \caption{Wasserstein barycenter of two Gaussian mixture marginals. \jiaojiao{Both of scatter plots and the level sets are exhibited.}}
  \label{fig:Gaussian mixture 3 marginal 2D results}
\end{figure}

In Figure \ref{fig:Gaussian mixture 3 marginal 2D results}, we further test NWB with 3 marginals of Gaussian mixtures. The first marginal is a uniform combination of 4 Gaussian components
\[
  N(\begin{bmatrix}
    4 \\ 4
  \end{bmatrix}\!,\!
  \begin{bmatrix}
    1 & 0 \\
    0 & 1
  \end{bmatrix}),~~
  N(\begin{bmatrix}
    4 \\ -4
  \end{bmatrix}\!,\!
  \begin{bmatrix}
    1 & 0 \\
    0 & 1
  \end{bmatrix}),~~
  N(\begin{bmatrix}
    -4 \\ -4
  \end{bmatrix}\!,\!
  \begin{bmatrix}
    1 & 0 \\
    0 & 1
  \end{bmatrix}),~~
  N(\begin{bmatrix}
    -4 \\ 4
  \end{bmatrix}\!,\!
  \begin{bmatrix}
    1 & 0 \\
    0 & 1
  \end{bmatrix}).
\]
The second marginal is a uniform combination of 3 Gaussian components
\[
  N(\begin{bmatrix}
    4 \\ 4
  \end{bmatrix}\!,\!
  \begin{bmatrix}
    1 & 0 \\
    0 & 1
  \end{bmatrix}),~~
  N(\begin{bmatrix}
    -4 \\ 4
  \end{bmatrix}\!,\!
  \begin{bmatrix}
    1 & 0 \\
    0 & 1
  \end{bmatrix}),~~
  N(\begin{bmatrix}
    0 \\ -4
  \end{bmatrix}\!,\!
  \begin{bmatrix}
    1 & 0 \\
    0 & 1
  \end{bmatrix}).
\]
The third marginal is a uniform combination of 3 Gaussian components
\[
  N(\begin{bmatrix}
    4 \\ -4
  \end{bmatrix}\!,\!
  \begin{bmatrix}
    1 & 0 \\
    0 & 1
  \end{bmatrix}),~~
  N(\begin{bmatrix}
    -4 \\ -4
  \end{bmatrix}\!,\!
  \begin{bmatrix}
    1 & 0 \\
    0 & 1
  \end{bmatrix}),~~
  N(\begin{bmatrix}
    0 \\ 4
  \end{bmatrix}\!,\!
  \begin{bmatrix}
    1 & 0 \\
    0 & 1
  \end{bmatrix}).
\]
For NWB, all the networks have 10 neurons for each hidden layer. The networks $f_i$ and $g_i$ each has 4 layers and the generative network $h$ has 6 layers. The initial learning rate is 0.001 and the learning rate drops $90$ percent every 20 epochs. For \citet{SolDeGui15}, the regularization intensity is set to 0.004.

We draw 1000 samples for each scatter plot. It can be seen that NWB can better capture the different modes of the distributions.

\subsection{Learning barycenters with sharp marginal distributions}
\begin{figure}[h]
  \centering
  \begin{subfigure}{.156\textwidth}
    \centering
    \includegraphics[width=0.95\linewidth]{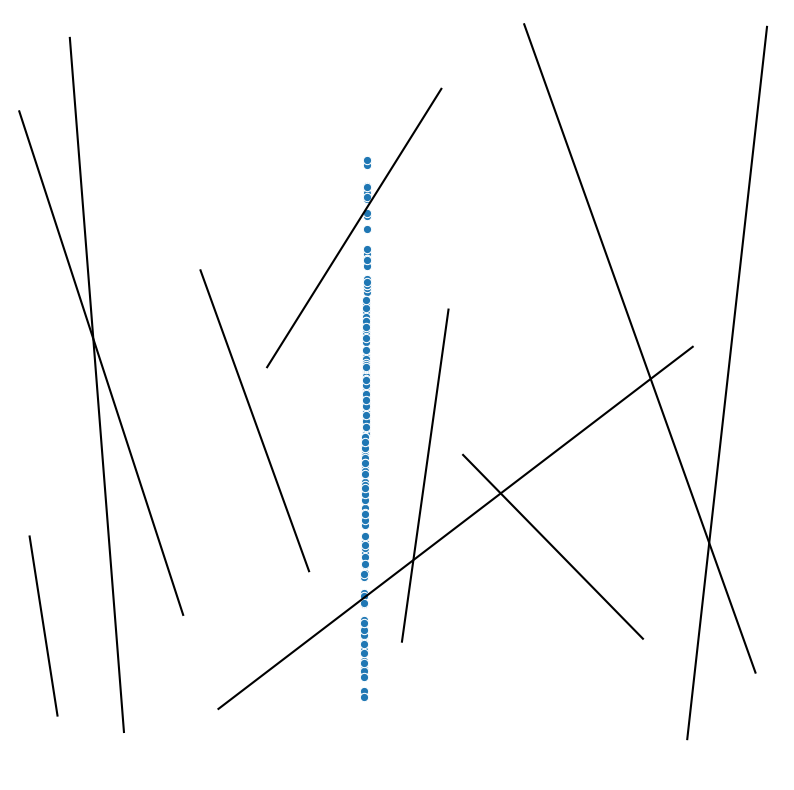}
    \caption{NWB $h \sharp \eta$}
  \end{subfigure}
  \begin{subfigure}{.16\textwidth}
    \centering
    \includegraphics[width=0.95\linewidth]{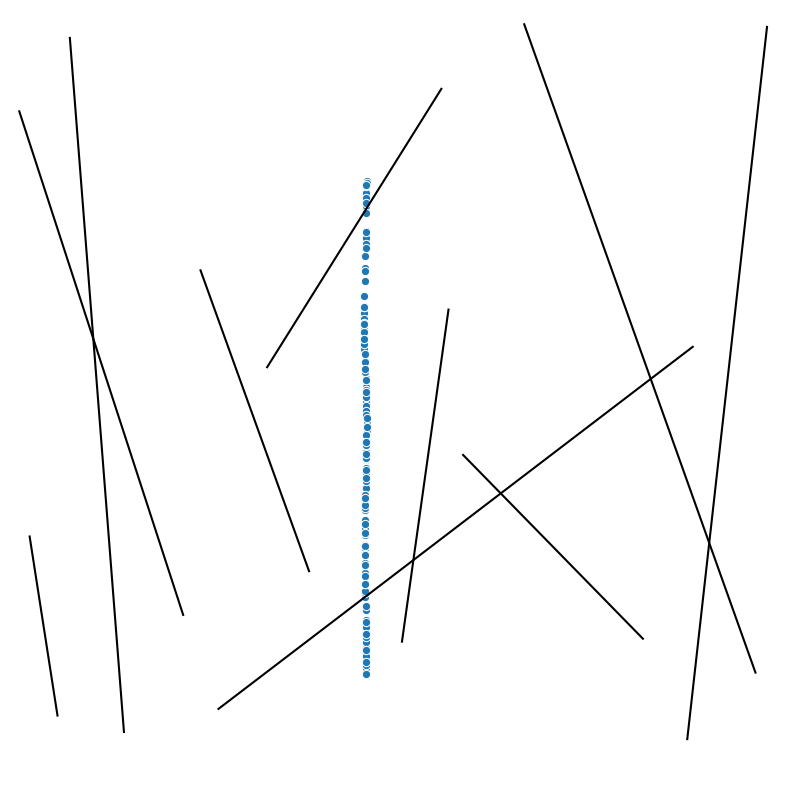}
    \caption{NWB $\nabla g_i \sharp \mu_i$}
  \end{subfigure}
  \begin{subfigure}{.16\textwidth}
    \centering
    \includegraphics[width=0.95\linewidth]{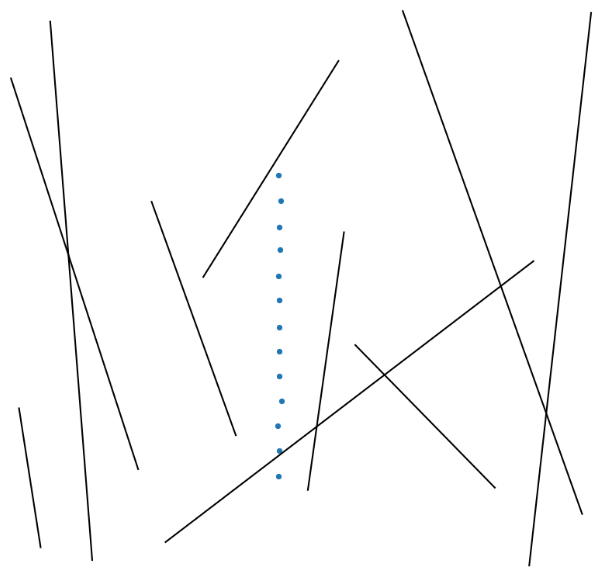}
    \caption{\citet{claici2018stochastic}}
  \end{subfigure}

  \begin{subfigure}{.14\textwidth}
    \centering
    \includegraphics[width=0.95\linewidth]{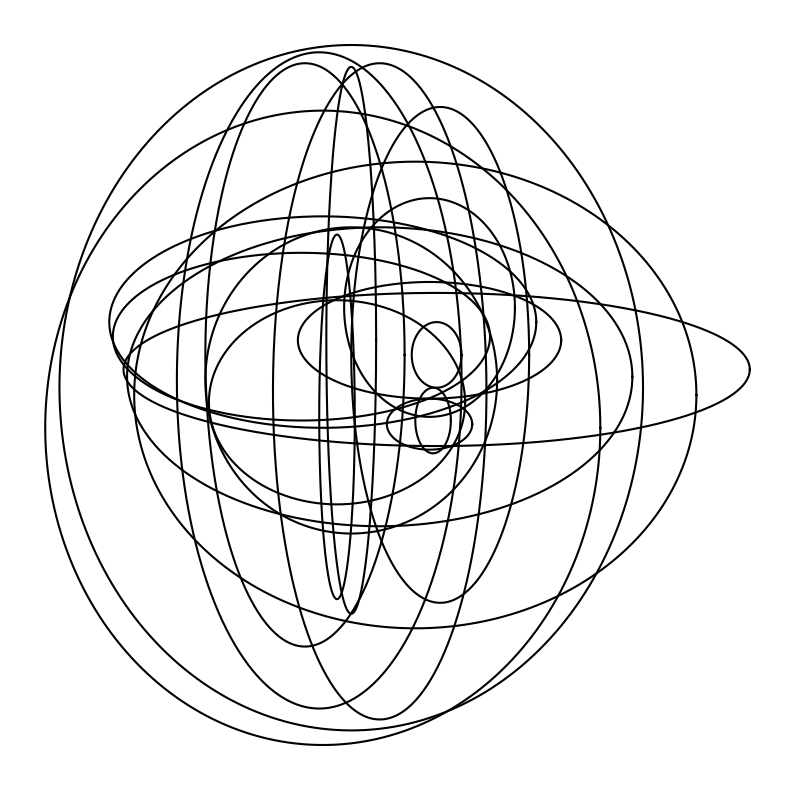}
    \caption{20 ellipses}
  \end{subfigure}
  \begin{subfigure}{.16\textwidth}
    \centering
    \includegraphics[width=0.95\linewidth]{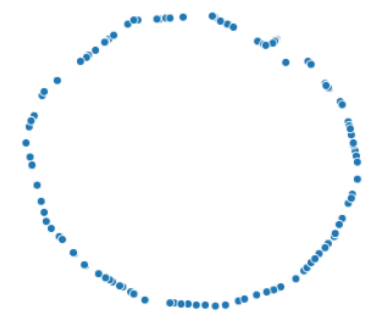}
    \caption{NWB $h \sharp \eta$}
  \end{subfigure}
  \begin{subfigure}{.16\textwidth}
    \centering
    \includegraphics[width=0.95\linewidth]{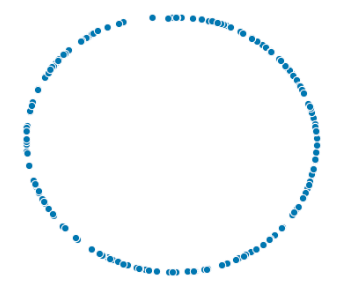}
    \caption{NWB $\nabla g_i \sharp \mu_i$}
  \end{subfigure}
  \begin{subfigure}{.16\textwidth}
    \centering
    \includegraphics[width=0.95\linewidth]{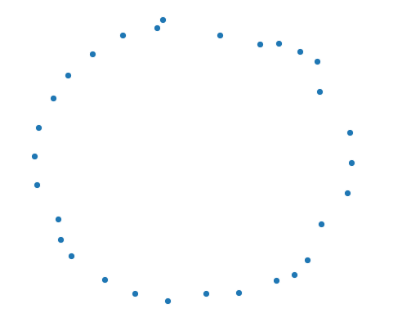}
    \caption{\citet{claici2018stochastic}}
  \end{subfigure}
  \caption{(a)-(c):Wasserstein barycenter of 10 distributions supported on random lines; (d)-(g):Wasserstein barycenter of 10 uniform marginal distributions supported on random ellipses shown in (d). 200 points are sampled from estimated barycenter. 13 points and 30 points are sampled from line and ellipse barycenter through the \citet{claici2018stochastic} because this is the maximum number of points allowed for the \citet{claici2018stochastic} to terminate in a reasonable amount of time.}
  \label{fig:line2d}
\end{figure}

We illustrate the performance of NWB in learning the Wasserstein barycenter when the marginal distributions are sharp. The common setup is all networks have 6 neurons for each hidden layer and the input Gaussian $\eta$ dimension is 1.
\paragraph{Line marginals}
We follow the examples reported in~\citet[Figure 4]{claici2018stochastic}, where the marginal distributions are uniform distributions on $10$ random two-dimensional lines as shown in Figure~\ref{fig:line2d}. It is observed that our algorithm is able to learn the sharp barycenter.
The network $f_i$ and $g_i$ each has 4 layers and $h$ has 4 layers.  $h$ network is linear. Learning rate is 0.0001. The inner loop iteration numbers are $K_1=6$ and $K_2=4$.
\paragraph{Ellipse marginals}
We also tested NWB on another example~\citep[Figure 6]{claici2018stochastic} to learn the barycenter of 10 uniform marginals supported on ellipses and obtained excellent results.
The network $f_i$ and $g_i$ each has 5 layers and $h$ has 4 layers. The initial learning rate is 0.001 and the learning rate drops $90$ percent every 15 epochs. The inner loop iteration numbers are $K_1=10$ and $K_2=6$.
\subsection{Learning the 2D and 3D Wasserstein Barycenter}
This is for the results in Figure \ref{fig:2or3d}.

The network $f_i$ and $g_i$ each has 4 layers and $h$ has 5 layers. In $h$ network, there is a batch normalization layer before each hidden layer.
All networks have 16 neurons for each hidden layer.
the input Gaussian $\eta$ dimension is equal to the marginal distribution dimension.

\paragraph{Circle-square example}  Learning rate is 0.001.

\paragraph{Block example} Learning rate is 0.001.

\paragraph{Digit 3 example} $f$ and $g$ learning rate is 0.0001, and $h$ is 0.001. Learning rate drops $90$ percent every outer cycle 12000 iterations. Our algorithm converges after 25000 outer cycle iterations.

\subsection{Scalability with the dimension}

\paragraph{Gaussian} The results are displayed in Figure \ref{fig:Gaussian 3 marginal highD results}.

The network $f_i$ and $g_i$ each has 4 layers and $h$ has 5 layers. In $h$ network, there is a batch normalization layer before each hidden layer.
All networks have $\max(10,2D)$ neurons for each hidden layer, where $D$ is the dimension of marginal distributions.
The input Gaussian $\eta$ dimension is equal to the marginal distribution dimension.
Learning rate is 0.001.
\paragraph{MNIST 0 and 1} The results are displayed in Figure \ref{fig:0-1} and Figure \ref{fig:backward}.

The network $f_i$ and $g_i$ each has 5 layers and $h$ has 5 layers. In $h$ network, we use batch normalization and dropout (probability 0.2 to be zeroed) operation before each hidden layer.
All networks have 1024 neurons for each hidden layer.
The input Gaussian $\eta$ dimension is 16.
Learning rate is initially 0.0001 for network $f_i$ and $g_i$; 0.001 for $h$, and drops $90$ percent every 1500 outer cycle iterations.
Our algorithm converges after 7500 outer cycle iterations.

\begin{figure}[ht!]
  \begin{subfigure}{0.24\textwidth}
    \centering
    \includegraphics[width=1\linewidth]{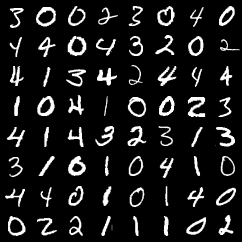}
    \caption{$\mu_1$: MNIST 0-4 digit}
  \end{subfigure}\hfill
  \begin{subfigure}{0.24\textwidth}
    \centering
    \includegraphics[width=1\linewidth]{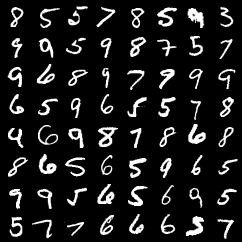}
    \caption{$\mu_2$: MNIST 5-9 digit}
  \end{subfigure}\hfill
  \begin{subfigure}{0.24\textwidth}
    \centering
    \includegraphics[width=1\linewidth]{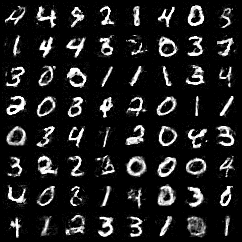}
    \caption{backward to $\mu_1$}
  \end{subfigure}\hfill
  \begin{subfigure}{0.24\textwidth}
    \centering
    \includegraphics[width=1\linewidth]{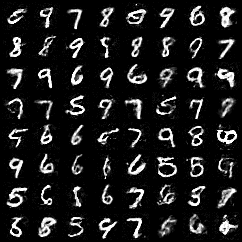}
    \caption{backward to $\mu_2$}
  \end{subfigure}
  \caption{MNIST 0-4 and 5-9 barycenter (784-dimensional problem): (a)-(b) Marginal distributions consisting of 0-4 and 5-9 digits; (c)-(d) Generating digit 0-4 and 5-9 from random input $Z \sim \mathcal{N} (\b0,I)$ using our architecture with the map $\nabla f_i( h(Z))$.}
  \label{fig:mnist-group}
  \vskip -0.2in
\end{figure}

\paragraph{MNIST 0-4 and 5-9}
To further evaluate our algorithm as a generative model for marginal distributions, we tested our algorithm on a upgraded task based on MNIST 0 and 1 experiment above.
The results are shown in Figure \ref{fig:mnist-group}.
The first marginal $\mu_1$ is an empirical distribution consisting of digit 0,1,2,3,4 samples and the second marginal $\mu_2$ is for digit 5,6,7,8,9.
We generate fresh samples from the barycenter using the generator $h(Z)$, where $Z \sim \mathcal{N} (\b0,I)$. We push-forward the samples $h(
  Z)$ through the maps $\nabla f_1(h(Z))$ and $\nabla f_2(h(Z))$ to generate new samples from the marginal distributions. It's expected that the panel (c) contains of only digits 0-4, and panel (d) only digits 5-9 and the results are consistent with the expectation.

The network $f_i$ and $g_i$ each has 5 layers and $h$ has 5 layers. In $h$ network, we use batch normalization before each hidden layer.
All networks have 1024 neurons for each hidden layer.
The input Gaussian $\eta$ dimension is 8.
Learning rate is initially 0.0001 for network $f_i$ and $g_i$; 0.001 for $h$, and drops $90$ percent every 25000 outer cycle iterations.
The number of outer cycle iterations is set to be $100,000$.

\begin{figure}[ht!]
  \begin{subfigure}{0.15\textwidth}
    \centering
    \includegraphics[width=1\linewidth]{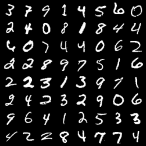}
    \caption{$\mu_1$: MNIST}
  \end{subfigure}\hfill
  \begin{subfigure}{0.15\textwidth}
    \centering
    \includegraphics[width=1\linewidth]{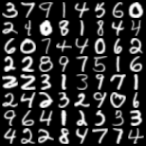}
    \caption{$\mu_2$: USPS}
  \end{subfigure}\hfill
  \begin{subfigure}{0.15\textwidth}
    \centering
    \includegraphics[width=1\linewidth]{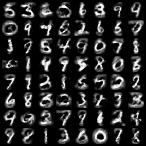}
    \caption{NWB $h \sharp \eta$}
  \end{subfigure}\hfill
  \begin{subfigure}{0.15\textwidth}
    \centering
    \includegraphics[width=1\linewidth]{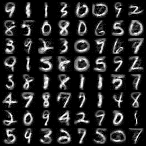}
    \caption{NWB $\nabla g_i \sharp (\mu_i)$}
  \end{subfigure}\hfill
  \begin{subfigure}{0.15\textwidth}
    \centering
    \includegraphics[width=1\linewidth]{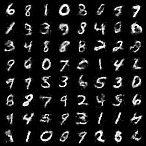}
    \caption{backward to $\mu_1$}
  \end{subfigure}\hfill
  \begin{subfigure}{0.15\textwidth}
    \centering
    \includegraphics[width=1\linewidth]{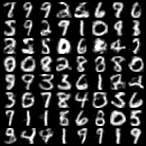}
    \caption{backward to $\mu_2$}
  \end{subfigure}
  \caption{USPS and MNIST barycenter (256-dimensional problem): (a)-(b) Marginal distributions consisting of MNIST and USPS digits;(c)-(d) NWB generates barycenter by generator $h$ and pushforward $\nabla g_i$;  (e)-(f) Generating MNIST and USPS digits from random input $Z \sim \mathcal{N} (\b0,I)$ using our architecture with the map $\nabla f_i( h(Z))$.}
  \label{fig:usps-mnist}
  \vskip -0.2in
\end{figure}

\paragraph{MNIST and USPS}  We tested our algorithm NWB on different datasets: MNIST and USPS. The results are displayed in Figure \ref{fig:usps-mnist}. We resize the MNIST samples to be $16 \times 16$ to be consistent with the USPS dataset. The dimension of this problem is thus 256. MNIST shows slimmer and smaller fonts compared to USPS digits. The barycenter fuses the two dataset styles, whereas $\nabla g_i \sharp (\mu_i)$ exhibits tidier results. Figure \ref{fig:usps-mnist} (e)-(f) show that our algorithm is able to generate new samples from both marginals (MNIST and USPS) with random Gaussian input using the same approach as in the previous example.

The network $f_i$ and $g_i$ each has 5 layers and $h$ has 6 layers. In $h$ network, we use batch normalization before each hidden layer.
All networks have 512 neurons for each hidden layer.
The input Gaussian $\eta$ dimension is 128.
Learning rate is initially 0.0001 and drops $90$ percent every 6000 outer cycle iterations.
The number of outer cycle iterations is set to be 75000.

\subsection{Subset posterior aggregation}
The results are displayed in Table \ref{tab:bayesian_inference}.
We preprocess the training data as follows \citep{korotin2021continuous}: i) apply the stochastic approximation trick to each ${\mu}_i$ \cite{minsker2014scalable}; ii) remove the mean of each marginal by shifting $\tilde{Y}_i=Y_i -m(\mu_i)$, where $m(\mu_i)$ is the mean of distribution $\mu_i$ \cite{alvarez2016fixed};
iii) scale each marginal distributions to be in a proper magnitude. Note that scaling won't affect $\text{BW}_{2}^{2} \textendash \text{UVP}$ value.
We use the same data preprocessing methods for other barycenter methods.
The network $f_i$ and $g_i$ each has 5 layers and $h$ has 5 layers. In $h$ network, there is a batch normalization layer before each hidden layer.
All networks have 10 neurons for each hidden layer.
The input Gaussian $\eta$ dimension is 8.
Learning rate is 0.01. Our algorithm converges after $8000$ outer cycle iterations.

\subsection{ Color palette averaging}
The results are displayed in Figure \ref{fig:pushforward images} and Figure \ref{fig:color palettes}.
The batch size is $M=1200$.
The network $f_i$ and $g_i$ each has 4 layers and $h$ has 5 layers. In $h$ network, there is a batch normalization layer before each hidden layer.
$f_i$ and $g_i$ networks have 16 neurons for each hidden layer, and $h$ has 32 neurons for each hidden layer.
The input Gaussian $\eta$ dimension is 3.
Learning rate is 0.001. Our algorithm converges after $100000$ outer cycle iterations.

\subsection{Serving as a Generative Adversarial Model in the one marginal setting}
WGAN and WGAN-GP results are generated using Pytorch-GAN library \cite{Erik2019github}.
W2GN results are generated using \citet{korotingithub} and adopt DenseICNN architecture proposed in the paper.
We refer the reader to the \citet[Section B.2, Section C.1]{korotin2021wasserstein} for DenseICNN and pretrain details.
The number of total training samples for both two experiments is 60000.

\paragraph{Gaussian mixture} The results are displayed in Figure \ref{fig:Gaussian mixture 1 marginal 2D results}.

For NWB, the networks $f_i$ and $g_i$ each has 5 layers and the generative network $h$ has 6 layers. All networks have 10 neurons for each hidden layer. The initial learning rate is 0.001.  The batch size is $M=60$.

For WGAN and WGAN-GP, they all use fully-connected linear layers and ReLU activation function. All discriminators and generators have 4 layers and 512 neurons for each hidden layer. Learning rate is 0.0001. The batch size is 256. The number of total iteration is 50000.

For W2GN, all netorks use DenseICNN $[3; 128, 128, 64]$ architecture. Here 3 is the rank of each input-quadratic skip-connection’s Hessian matrix. Each following number represents the size of a hidden dense layer in the sequential part of the network.
The batch size is 1024.
Learning rate is initially 0.001 and drops $90$ percent every 25000 iterations. The number of total iteration is 50000.

\paragraph{MNIST}
The results are displayed in Figure \ref{fig:mnist GAN}. We normalize MNIST pixel values to be in range $[-1,1]$ before training.

For NWB, the network $f_i$ and $g_i$ each has 5 layers and $h$ has 6 layers. In $h$ network, there is a batch normalization layer before each hidden layer.
All networks have 1024 neurons for each hidden layer.
The input Gaussian $\eta$ dimension is 64.
Learning rate is initially 0.0001 and drops $90$ percent every 100 epochs. The total epoch is set to be 500 epochs.

For WGAN and WGAN-GP, to be fair, they all use the same batch size, batch-normalization and fully-connected linear layers as NWB.
The activation function is LeakyReLU.
From input layer to output layer, the generator neuron for each layer is $[100,128,256,512,1024,784]$; and the discriminator is $[784,512,256,1]$.
The final layer for the generator is also tanh.
Learning rate is initially 0.0001 and drops $90$ percent every 300 epochs. The total epoch is set to be 1500 epochs.

For W2GN, all networks use DenseICNN $[2; 2048, 2048, 2048]$ architecture.
The batch size is also 100.
Learning rate is initially 0.0001 and drops $90$ percent every 300 epochs. The total epoch is set to be 1500 epochs.

\section{Experiment details for CDWB \citep[Section 4.4]{CutDou14}}
We use POT library~\cite{flamary2017pot} and adopt Earth Movers distance solver \cite{bonneel2011displacement} when solving OT programming in the inner loop.

\section{Experiment details for CRWB \cite{li2020continuous}}
We use the code given by \citet{li2020github}. We use quadratic regularization, which is empirically more stable than entropic regularization.
We set potential networks as fully connected neural networks.
The hidden layer sizes are given by
$$[\max(128, 2D), \max(128, 2D), \max(128, 2D)],$$ where $D$ is the marginal distribution dimension.
The activation functions are all ReLU.
The batch size as 1024.
We use Adam optimizer with fixed learning rate $10^{-4}$ for Bayesian inference and MNIST examples and $10^{-3}$ for Gaussian examples.
We use Monge map to recover barycenter samples \citep[Equation (13)]{li2020continuous}.
The total number of iterations is set to 50000.

\section{Experiment details for CWB \cite{korotin2021continuous}}
We use the code in \url{https://openreview.net/forum?id=3tFAs5E-Pe}. As mentioned in the \citet[Section A]{korotin2021continuous}, we also pretrain the potential networks as an initialization step.
We use Adam optimizer with fixed learning rate $10^{-4}$ for Bayesian inference and MNIST examples and $10^{-3}$ for Gaussian examples. Other setups are exactly the same as the \citep[Section C.4.1]{korotin2021continuous}.

\section{Optimization landscape for class of quadratic functions}\label{sec:landscape}
In order to understand and compare various optimization formulations to estimate Wasserstein barycenter, it is insightful to examine them for special cases where analysis is feasible. For this purpose, we study the optimization landscape of our formulation and~\cite{korotin2021continuous} in the special case  where the class of functions is restricted to quadratic functions. In particular, we show that our optimization formulation simplifies to a smooth concave-convex-concave optimization for this special case, while the formulation of~\cite{korotin2021continuous} is non-smooth and non-convex.

We consider the simplest case that the functions are parameterized as follows: $h(z)=z+\alpha$, $f_i(x) = \frac{1}{2}\|x\|^2 + \beta_i^T x$, and $g_i(y) = \frac{1}{2}\|y\|^2 + \gamma_i^T y$, where $\alpha,\beta_i,\gamma_i\in \mathbb{R}^n$ are the parameters that serve as optimization variables. Then, in this case, the optimization problem~\eqref{eq:FICNN final object} simplifies to
\begin{equation}
  \min_\alpha\max_{\{\beta_i\}_{i=1}^N}\min_{\{\gamma_i\}_{i=1}^N} \sum_{i=1}^N a_i\left[ \frac{1}{2}\|\gamma_i\|^2 +\beta_i^T(\gamma_i + m_i - \alpha)\right]
\end{equation}
where $m_i = \mathbb{E}_{\mu_i}[Y^i]$. The objective function is convex in $\gamma_i$, linear in $\beta_i$ and $\gamma_i$. Inserting the optimal value for $\gamma_i=-\beta_i$ yields
\begin{equation}
  \min_\alpha\max_{\{\beta_i\}_{i=1}^N} \sum_{i=1}^N a_i\left[ -\frac{1}{2}\|\beta_i\|^2 +\beta_i^T( m_i - \alpha)\right].
\end{equation}
This is concave in $\beta_i$. Inserting the optimal value $\beta_i=m_i-\alpha$ yields
\begin{equation}
  \min_\alpha \frac{1}{2}\|\alpha\|^2 - \alpha^T \sum_{i=1}^N a_im_i +  \sum_{i=1}^N a_i \|m_i\|^2
\end{equation}
which is convex in $\alpha$ with optimal value at $\alpha = \sum_{i=1}^N a_im_i$. This means that the optimal generator $h(z) = z +  \sum_{i=1}^N a_im_i$ learns average mean of the marginal distributions. This is the exact Wassserstein barycenter for the case that marginal distributions are  Gaussian distributions with the same covariance.

In contrast, consider the optimization formulation of~\citep[Eq. 14]{korotin2021continuous} with the following parameterization: $\psi_i^\dagger(x) = \frac{1}{2}\|x\|^2 + \alpha_i^T x$  and $\bar{\psi}_i^{\dagger\dagger}(x) = \frac{1}{2}\|x\|^2 + \beta_i^T x$. Then, the optimization problem simplifies to
\begin{equation}
  \min_{\{\alpha_i\}_{i=1}^N,\{\beta_i\}_{i=1}^N}  \sum_{i=1}^N a_i\left[-\frac{1}{2}\|\alpha_i\|^2 - \alpha_i^T\beta_i - m_i^T\beta_i\right] + \tau\mathbb{E}_{\hat{P}}\left[ \left(\sum_{i=1}^N a_i \beta_i^TY\right)_+ \right] + \lambda \sum_{i=1}^N a_i\|\alpha_i+\beta_i\|^2
\end{equation}
where $\hat{P}$ is a distribution that should be chosen such that $\tau \hat{P}$ is larger than the barycenter density (with $\tau>1$). Although the optimization problem involves single minimization compared to our min-max-min formulation, the optimization objective is much more complicated. Our first observation is that the optimization problem is not convex in $\alpha_i$ if $\lambda <\frac{1}{2}$ (the optimization algorithm diverges). Inserting the optimal value $\alpha_i=-\beta_i$, the optimization becomes
\begin{equation}
  \min_{\{\beta_i\}_{i=1}^N}  \sum_{i=1}^N a_i\left[\frac{1}{2}\|\beta_i\|^2 - m_i^T\beta_i\right] + \tau\mathbb{E}_{\hat{P}}\left[ \left(\sum_{i=1}^N a_i \beta_i^T Y\right)_+ \right].
\end{equation}
This is convex, but non-smooth optimization problem in $\beta_i$.  In order for the expected solution $\beta_i = m_i - \sum_{i=1}^N a_im_i$ be optimal for this problem, it must satisfy the first-order optimality condition
\begin{equation*}
  a_i(\beta_i - m_i) + \tau a_i \partial l(0) = 0
\end{equation*}
where $l(\xi) := \mathbb{E}_{\hat{P}}\left[ \left(\xi Y\right)_+ \right]$ and $\partial l(0)$ denotes an element in sub-differential of $l(\xi)$ at $\xi=0$. The function $l(\xi)=\mu_+ \xi$ for $\xi>0$ and $l(\xi) = -\mu_-\xi$ for $\xi<0$, where $\mu_+ = \mathbb{E}_{\hat{P}}\left[ \left( Y\right)_+ \right] $ and $\mu_- = \mathbb{E}_{\hat{P}}\left[ \left( -Y\right)_+ \right]$. As a result,  the sub-differential $\partial l(0) \in [-\mu_-,\mu_+]$.
Therefore, summing the first-order optimality condition for $i=1,\ldots,N$ implies
\begin{align*}
  \sum_{i=1}^N a_im_i \in [-\tau \mu_-,\tau \mu_+].
\end{align*}
Although, this condition holds when $\hat{P}$ is Gaussian centered at the barycenter location $\sum_{i=1}^N a_im_i$ (with $\tau>1$), it may not hold with other $\hat{P}$. For example, if $\hat{P}$ is $N(0,1)$, then $\mu_+=\mu_-=\frac{1}{\sqrt{2\pi}}$, and the condition does not hold if $ \sum_{i=1}^N a_im_i > \frac{\tau}{\sqrt{2\pi}}$.

\end{document}